\PassOptionsToPackage{dvipsnames}{xcolor}
\documentclass[onefignum,onetabnum]{siamart220329}
\usepackage{amssymb,amsbsy,amsmath,amsfonts,amssymb,amscd}
\usepackage{mathrsfs,hyperref}
\usepackage{cleveref}
\usepackage{macros}

\usepackage{epsfig}
\usepackage[margin = 1in]{geometry}
\graphicspath{{./figures/}}
\usepackage{tikz}
\usetikzlibrary{patterns}
\usetikzlibrary{quotes, angles}
\usetikzlibrary{decorations.markings}
\usetikzlibrary{svg.path} 

\usepackage{amsfonts}
\usepackage{graphicx}
\usepackage{subcaption}
\usepackage{amsmath}
\usepackage{amssymb}
\usepackage{amsmath}
\usepackage{bm}
\usepackage{cases}
\usepackage[shortlabels]{enumitem}
\usepackage{algorithmic}

\usepackage{xcolor}

\newcommand{\revise}[1]{\textcolor{blue}{#1}}
\usepackage{xcolor}

\newcommand{\revone}[1]{#1}
\newcommand{\revtwo}[1]{#1}

\DeclareMathOperator*{\argmin}{arg\,min}

\title{Cryo-EM as a Stochastic Inverse Problem\thanks{Submitted to the editors. \funding{This work is partially supported by the National Science Foundation through grant  DMS-2409855 and by the Office of Naval Research  through grant
N00014-24-1-2088.}}}

\author{Diego Sanchez Espinosa\thanks{Center for Applied Mathematics, Cornell University, Ithaca, NY (\email{ds973@cornell.edu}).}
\and Erik Henning Thiede\thanks{Department of Chemistry, Cornell University, Ithaca, NY (\email{eht45@cornell.edu}).}
\and Yunan Yang\thanks{Department of Mathematics, Cornell University, Ithaca, NY (\email{yunan.yang@cornell.edu})}}

\headers{}{Diego S.~Espinosa, Erik H.~Thiede, Yunan Yang}

\date{\today}

\begin{document}
\maketitle
\begin{abstract}
Cryo-electron microscopy (Cryo-EM) enables high-resolution imaging of biomolecules, but structural heterogeneity remains a central challenge in three-dimensional reconstruction. Standard pipelines often approximate heterogeneity by a finite set of conformational classes, which can be restrictive when the underlying variability is continuous. In this work, we formulate heterogeneous cryo-EM reconstruction as a stochastic inverse problem over probability measures. The observed image distribution is modeled as the random push-forward of an unknown conformational distribution through the cryo-EM imaging operator, including random viewing directions and image noise. We recover this latent distribution by minimizing a statistical discrepancy between observed and simulated image distributions, using losses such as the KL divergence, Maximum Mean Discrepancy, and energy distance. The resulting variational problem is optimized over probability measures through a Wasserstein gradient flow, discretized by an interacting particle method. We further analyze local inverse stability, sample complexity, and cryo-EM observability, identifying how recovery depends on image noise, particle number, Monte Carlo approximation, model mismatch, and the visibility of structural perturbations under the imaging operator. We also connect the formulation to maximum a posteriori approaches, interpreting them as discretize-then-optimize schemes, and give consistency conditions under which such discretizations converge to the underlying infinite-dimensional problem. Numerical experiments on synthetic systems and protein models demonstrate recovery of continuous conformational distributions and compare the particle method with Gaussian-mixture baselines. The proposed framework provides a general methodology for stochastic inverse problems involving random forward operators.
\end{abstract}

\begin{keywords}
cryo-EM, stochastic inverse problem, Wasserstein gradient flow,  particle method,  push-forward map
\end{keywords}

\begin{MSCcodes}
65M32,  49Q22,  65M75,  65K10
\end{MSCcodes}


\section{Introduction}
Understanding the molecular structure of biomolecules is fundamental for elucidating their function, dynamics, and interactions. Cryo-electron microscopy (Cryo-EM) has emerged as a transformative technology for characterizing the structures of biomolecules \cite{Frank2006, Henderson_1995, Callaway2015, Subramaniam2016}. Traditional data analysis pipelines then attempt to reconstruct the underlying three-dimensional (3D) structure from the ensemble of noisy two-dimensional (2D) projections \cite{Sigworth2015, Glaeser2016}. Due to advances in detector technology, electron optics, and sample preparation, experimentalists can then recover biomolecular structures to sub-nanometer, or even atomic, resolutions \cite{FARUQI2007549, Kuhlbrandt2014, Nakane2020, Carragher2019}. However, one of the major complications in cryo-EM is structural heterogeneity—the fact that biomolecules, such as proteins, often adopt multiple atomic arrangements in 3D-space, known as ``conformations''. This heterogeneity is not merely an experimental nuisance: conformational changes regulate processes such as enzyme catalysis and signal transduction \cite{Alberts2003,Chothia1986,Goodsell2000}, and their characterization can inform therapeutic design, mutagenesis experiments, and the study of disease mechanisms \cite{Shoichet2004,Dill2012}.

\revtwo{Traditional reconstruction pipelines, including RELION \cite{scheres2012relion,Scheres2012bayes} and cryoSPARC \cite{Punjani2021,Punjani2020uniform}, often represent heterogeneity through a finite number of conformational classes. While widely used and effective for well-separated states, these methods are often believed to be insufficient for capturing smooth or continuous structural variability.}

\revtwo{Another direction seeks low-dimensional descriptions of
continuous heterogeneity, including manifold embedding and diffusion maps~\cite{Frank2016,Moscovich2020}, linear or covariance-based models~\cite{Punjani2021,TAGARE2015}, flexible-motion models such as 3DFlex~\cite{Punjani2023F}, and neural-network latent-variable models such as cryoDRGN~\cite{Zhong2021}. These methods provide coordinates, modes, or latent variables for structural variability and can sometimes be connected to conformational landscapes or free-energy-like interpretations. Yet another  direction aims to estimate or reweight a probability distribution over conformational states. Tang et al.~\cite{Tang2023} propose
an ensemble-reweighting strategy that assigns weights to a prior ensemble of structures according to agreement with observed images, while RECOVAR~\cite{recovar} estimates low-order statistics of the conformational distribution through regularized covariance estimation, followed by low-dimensional embedding, density estimation, and kernel-regression-based
reconstruction. See \cite{tang2023review} for a broader review.}

Our approach is complementary to these methods: we formulate heterogeneous cryo-EM reconstruction as a stochastic inverse problem over probability measures. The unknown object is the conformational distribution, and the observed image distribution is modeled as its random push-forward through the imaging model. We pose this reconstruction task as a variational problem over probability measures. We then seek the structural distribution that minimizes a statistical discrepancy, such as  the Kullback--Leibler (KL) divergence or Maximum Mean Discrepancy (MMD), between the predicted image distribution and the observed empirical image distribution. We optimize this objective using a Wasserstein gradient flow, which gives a geometric descent dynamics on the space of probability measures. The resulting flow is discretized by an interacting particle method, so that the evolving particle ensemble represents the recovered conformational distribution.

The paper makes three main contributions. First, it introduces a stochastic inverse-problem formulation for heterogeneous cryo-EM with random forward operators and derives the corresponding Wasserstein gradient-flow particle method. Second, it analyzes local inverse stability, sample complexity, and cryo-EM observability, identifying how recovery depends on image noise, particle number, Monte Carlo approximation, model mismatch, and angular coverage. Third, it connects the proposed optimize-then-discretize formulation with maximum-likelihood and maximum-a-posteriori approaches, which can be interpreted as discretize-then-optimize schemes, and gives consistency conditions relating the two formulations.

Numerical experiments illustrate the behavior of the method on synthetic examples and protein models. We study a one-dimensional test problem, a nanocluster heterogeneity example, and a realistic protein-motion example. We also include a controlled one-mode protein study comparing the particle method with Gaussian-mixture baselines and examining sensitivity to particle number and image noise.

The paper is organized as follows. Section~\ref{sec:SIP} introduces the stochastic inverse-problem formulation and the Wasserstein gradient-flow method. Section~\ref{sec:OTD-DTO} discusses optimize-then-discretize and discretize-then-optimize approaches, including the connection with MAP-based cryo-EM methods. Section~\ref{sec:stability} studies local stability, sample complexity, and observability. Section~\ref{sec:numerics} presents numerical experiments, and Section~\ref{sec:conclusion} concludes.

\section{Cryo-EM as a Stochastic Inverse Problem}\label{sec:SIP}
In a cryo-EM experiment, a solution containing many copies of a  protein is snap-frozen into a thin layer of vitreous ice.  

Imaging the layer with an electron microscope yields many images, each displaying a protein viewed at a random rotation. A simplified model for the  imaging pipeline is given by

\begin{equation}\label{eq:cryo-EM-eqn}
    y = T(\theta,\omega) = H^{\omega_r}\,\theta  +  \omega_n,
\end{equation}

where $H^{\omega_r}$ is a linear operator on the space of 3D structures that incorporates a random rigid motion in $SO(3)$,  an orthographic projection onto the $z$-axis, and convolution with the microscope's point-spread function. The random variable $\omega$ has two independent components, $\omega=(\omega_n,\omega_r)$.  The noise term $\omega_n$ is additive Gaussian noise, $\omega_n\sim\mathcal{N}(0,\sigma^{2}I)$, while $\omega_r$ encodes the random imaging geometry and is distributed according to a measure $\rho_{\omega_r}$.  

In contrast with traditional approaches to cryo-EM, the object of interest is not a single structure $\theta$, but a conformational distribution $\rho_\theta\in\mathcal P(\Theta)$. Given the empirical distribution of observed images $\rho_{y}$, our objective is to infer the distribution of underlying molecular structures $\rho_{\theta}$ via the \textit{random} parameter-to-observable map~\eqref{eq:cryo-EM-eqn}. 
This inference task is challenging: depth information is lost in the projection, the viewing directions are latent, and the signal-to-noise ratio is often extremely low. These challenges motivate us to consider a new approach for cryo-EM.

Below, we present our primary theoretical contributions of this work: we formalize recovering the distribution $\rho_\theta$ as a stochastic inverse problem, relax it to a variational problem, and show how this variational problem can be solved using gradient flows.

\subsection{From Deterministic to Stochastic Inverse Problems}

In a deterministic inverse problem, a forward operator $T:\Theta\rightarrow Y$ maps a parameter $\theta\in\Theta$ to data $T(\theta)\in Y$. The observed data $y^\delta\in Y$ may contain noise and therefore may not lie in the range $T(\Theta)$. In stochastic inverse problems (SIPs)~\cite{breidt2011measure,li2024differential}, the unknown parameter is lifted from a single point in the set $\Theta$ to a probability distribution over $\Theta$ that lies in the set $\mathcal{P}(\Theta)$, which consists of all probability distributions defined over $\Theta$. Data also become distributions over $Y$ that live in $\mathcal{P}(Y)$.

In many SIPs, the operator that maps the parameter distribution to the data distribution is a push-forward operator $F = T_{\#}$  where the map $T: \Theta \rightarrow Y$ is a deterministic forward map. In these setups, the observed data distribution is the push-forward measure of the parameter distribution in $\mathcal{P}({\Theta})$. We give the definition of the push-forward measure below.

\begin{definition}\emph{(Push-Forward Measure).}
Let $(X,\mathcal{A})$ and $(Y, \mathcal{B})$ be measurable spaces, and let $f: X \to Y$ be a measurable function. The push-forward measure $f_{\#}\mu$ of $\mu$ under $f$ is a measure on $Y$ defined by
\[
f_{\#} \mu(B) = \mu(f^{-1}(B))\,,
\]
for all $ B \in \mathcal{B}$, the $\sigma$-algebra of set $Y$.
\end{definition}
We address two common scenarios where the push-forward measure can be easily understood. 

\begin{itemize}
\item If $\mu$ is absolutely continuous with respect to the Lebesgue measure, it is associated with a probability density function $\rho(x)$ such that $d\mu(x) = \rho(x) dx$. If $f$ is a differentiable bijection, then the push-forward measure $f_\#\mu$ is also associated with a probability density function $\Tilde{\rho}(x)$ where the relationship between $\rho(x)$ and $\Tilde{\rho}(x)$ can be understood through the change-of-variable formula \revise{$\rho(x) = \Tilde{\rho}(f(x)) \left|\text{det}(\nabla f)\right|$}.

\item Consider $\mu = \sum_{i=1}^N w_i \delta_{x_i}$,  a weighted sum of Dirac deltas, for $N\in \mathbb{N}$, $\sum_{i=1}^N w_i = 1$ and $w_i \geq 0$. Then the push-forward measure has the explicit form

\[
f_{\#} \mu = \sum_{i=1}^N w_i \delta_{f(x_i)}\,.
\]
That is, the push-forward map does not change the weight of the samples, but geometrically moves the support of the sample from $x_i \in X$ to $f(x_i) \in Y$.
\end{itemize}
\smallskip

\subsection{General Stochastic Inverse Problem}

While most SIPs use a deterministic push-forward map, cryo-EM and related applications~\cite{vadeboncoeur2025efficient} involve a random forward map $T_\omega$, where $\omega$ is a random variable with distribution $\mu_\omega$.

For each realization $\homega$ drawn from $\mu_\omega$,  the induced map $T_{\homega}: \Theta \rightarrow Y$,   pushes the parameter probability distribution $\rho_\theta$ to the data probability distribution over the $Y$ domain. One can also view $T_{\homega}$ as a map parameterized by  $\homega$, which motivates us to use the following notations $T_\omega$, $T_\theta$ and $T$ hereafter:
\[
    y = T_{\omega}(\theta) \,\,\Longleftrightarrow \,\, y = T (\omega, \theta) \,\,\Longleftrightarrow \,\,  y = T_\theta(\omega) \,.
\]

The forward operator that maps a given parameter distribution $\rho_\theta$ to the corresponding predicted data distribution $\rho_y$ can be expressed as
\begin{eqnarray}
\rho_y &=& F(\rho_\theta) \nonumber \\
&=& T_{\#}   \left( \mu_\omega  \otimes \rho_\theta \right)  = \iint \delta_{y = T(\homega, \hat\theta)}\,d\mu_\omega (\homega)  \,  d\rho_\theta(\hat\theta)  \label{eq:ip0}  \\
&=&\int  T_{\homega}\,_{\#} \rho_\theta \, \,  d\mu_\omega (\homega)\label{eq:ip} \\
&=& \int T_{\hat\theta} \,_{\#}\mu_\omega  \,  d\rho_\theta(\hat\theta)\,. \label{eq:ip_conv} 
\end{eqnarray}
We write our forward operator in these three equivalent ways with different interpretations:
\begin{itemize}
    \item In~\eqref{eq:ip0}, the data distribution $\rho_y$ is a push-forward measure from the joint distribution $\mu_\omega  \otimes \rho_\theta$ by the map $T$. This perspective reduces to those discussed in~\cite{breidt2011measure,li2024differential}. In particular, $\pi(y,\homega, \hat\theta)$ is the joint distribution of all three variables where
    \begin{equation}  \label{eq:joint}
    d\pi(y,\homega, \hat\theta) = \delta_{y = T(\homega, \hat\theta)} d\mu_\omega (\homega)  \,  d\rho_\theta(\hat\theta)\,
    \end{equation}
    while $\rho_y$ is regarded as the marginal distribution of $\pi$ after integrating over the $\homega$ and $\hat\theta$ variables.\smallskip
    
    \item In~\eqref{eq:ip}, the data distribution is a weighted ensemble of all realizations of the push-forward measure $T_{\homega}\,_{\#} \rho_\theta$ based on the law $\mu_\omega$. More precisely, $T_{\homega}\,_{\#} \rho_\theta$ is the marginal distribution of $\pi$ after integrating over the $\hat\theta$ variable.
    \smallskip
    
    \item In~\eqref{eq:ip_conv}, the data distribution is a weighted ensemble of all realizations of the push-forward measure $T_{\hat\theta}\,_{\#} \rho_\omega$ based on the law $\rho_\theta$. Since $\rho_\theta$ showed up  as the integration measure, \eqref{eq:ip_conv} is quite favorable when representing $\rho_\theta$ as an empirical measure. Similarly, $T_{\hat\theta}\,_{\#} \rho_\omega$ is the marginal distribution of the joint distribution $\pi$ in~\eqref{eq:joint} after integrating over the $\homega$ variable.
\end{itemize}
The goal of the SIP is to recover $\rho_\theta$, given a noisy observed distribution $\rho_y^\delta$, the law $\mu_\omega$ and the forward operator \eqref{eq:ip0}, \eqref{eq:ip} or \eqref{eq:ip_conv}.

If we assume that $T$ is invertible with an executable map $T^{-1}$ and the data distribution $\rho_y^\delta$ is in the range of the map $T_\#$, we can solve the inverse problem directly through 
\[
\mu_\omega^* \otimes \rho_\theta^* = \left(T^{-1}\right)_{\#} \rho_y^\delta\,.
\]
\revise{where $\rho_y^\delta$ is our observed data distribution with $\delta$ highlighting the presence of noise $\delta$.}  
However, in almost all interesting inverse problems, $T$ is not invertible. For example, in cryo-EM, $T$ is a projection operator whose output dimension is strictly lower than the input dimension; see Equation~\eqref{eq:cryo-EM-eqn}. On the other hand, the fact that $T^{-1}$ does not exist does not necessarily imply that $F$ is not invertible or that the SIP is ill-posed. We therefore turn to a variational framework to solve the inverse problem computationally. 

\subsection{The Variational Framework}
Due to the presence of noise and the nonlinearity of the forward map $F$, 
one can relax this infinite-dimensional SIP~\eqref{eq:ip} to a variational problem, which takes the form
\begin{equation}\label{eq:min}
\inf_{\rho_\theta \in \Omega \subset  \mathcal{P}(\Theta)} E(\rho_\theta), \quad E(\rho_\theta) =D\left(F(\rho_\theta),  \rho_y^\delta\right)\,,
\end{equation}
where the search domain $\Omega$ is a subset of $\mathcal{P}(\Theta)$, with the latter representing all probability measures over the domain $\Theta$, and $D$ quantifies the discrepancy between $\rho_y^\delta$ and $F(\rho_\theta)$.
\revtwo{Under suitable compactness assumptions on the admissible set $\Omega$ and the lower semicontinuity of the objective function $E(\rho_\theta)$~\cite{li2024stochastic}}, 
the minimizer of the variational problem exists and 
the ``$\inf$'' in~\eqref{eq:min} can be replaced by the ``$\min$''. We thus assume such conditions hold and all variational problems admit a minimizer hereafter.

The space of probability distributions is  not a linear space. The constraint of non-negativity and the conservation of mass, as well as its central role in fields such as statistics and probability, have given rise to a wide range of geometries. Below, we give a few examples of metrics and divergences \revtwo{that can be used for $D$.}
\begin{definition}\label{def:wp}\emph{($p$-Wasserstein metric).}
Let $(\mathcal{X}, d)$ be a metric space, and let $\mu$ and $\nu$ be two probability measures on $\mathcal{X}$ with finite $p$-th order moments. The \emph{$p$-Wasserstein distance} (also known as the $W_p$ metric) between $\mu$ and $\nu$ is defined by
\[
W_p(\mu, \nu) = \left( \inf_{\gamma \in \Gamma(\mu, \nu)} \int_{\mathcal{X} \times \mathcal{X}} d(x, y)^p \, d\gamma(x, y) \right)^{1/p},
\]
where $\Gamma(\mu, \nu)$ denotes the set of all \emph{couplings} of $\mu$ and $\nu$. A coupling $\gamma$ is a probability measure on $\mathcal{X} \times \mathcal{X}$ such that for all measurable subsets $A, B \subseteq \mathcal{X}$, $\gamma(A \times \mathcal{X}) = \mu(A)$ and $\gamma(\mathcal{X} \times B) = \nu(B)$.
\end{definition}
The $W_p$ metric measures the minimal ``cost'' of transporting mass from one distribution to another, where the cost is given by the distance to the $p$ power, $d(x, y)^p$, between $x$ and $y$.

\begin{definition}\label{def:f-divergence}\emph{($f$-divergence).}
Let $(\Omega, \mathcal{F})$ be a measurable space, and let $\mu$ and $\nu$ be two probability measures on this space such that $\mu$ is absolutely continuous with respect to $\nu$ (denoted $\mu \ll \nu$). Let $f: (0, \infty) \rightarrow \mathbb{R}$ be a convex function with $f(1) = 0$. The \emph{$f$-divergence} between $\mu$ and $\nu$ is defined by
\[
D_f(\mu \, \| \, \nu) = \int_{\Omega} f\left( \frac{d\mu}{d\nu}(x) \right) \, d\nu(x),
\]
where $\dfrac{d\mu}{d\nu}$ is the Radon--Nikodym derivative of $\mu$ with respect to $\nu$.
\end{definition}

Different choices of the convex function $f$ yield various well-known divergences. In particular, the Kullback--Leibler (KL) divergence corresponds to $f(t) = t \ln t - t + 1$:
\begin{equation}\label{eq:KL}
    D_{\mathrm{KL}}(\mu \, \| \, \nu) = \int_{\Omega} \ln\left( \frac{d\mu}{d\nu}(x) \right) \, d\mu(x).
\end{equation}

\begin{definition}\label{def:MMD}\emph{(Maximum Mean Discrepancy).}
Let $k: \mathcal{X} \times \mathcal{X} \rightarrow \mathbb{R}$ be a continuous, symmetric, positive-definite kernel function. Let $\mu$ and $\nu$ be two probability distributions defined on $\mathcal{X}$. The Maximum Mean Discrepancy (MMD) between $\mu$ and $\nu$ is defined by
\begin{equation}\label{eq:MMD}
\text{MMD}( \mu, \nu) = \sqrt{\mathbb{E}_{x, x' \sim \mu} [k(x, x')] + \mathbb{E}_{y, y' \sim \nu} [k(y, y')] - 2\, \mathbb{E}_{x \sim \mu, y \sim \nu} [k(x, y)]}\,.
\end{equation}
\end{definition}
It can be shown that $\text{MMD}( \mu, \nu) \geq 0$,
and  the choice of kernel $k$ affects the sensitivity of MMD to differences between two distributions. 
\revtwo{Given independent samples $\{x_i\}_{i=1}^m \overset{\mathrm{i.i.d.}}{\sim}\mu$
and $\{y_j\}_{j=1}^n \overset{\mathrm{i.i.d.}}{\sim}\nu$, with the two samples
independent of each other, an unbiased empirical estimator of the squared MMD is}

\[
\widehat{\text{MMD}}^2 = \frac{1}{m(m - 1)} \sum_{i \neq i'} k(x_i, x_{i'}) + \frac{1}{n(n - 1)} \sum_{j \neq j'} k(y_j, y_{j'}) - \frac{2}{mn} \sum_{i, j} k(x_i, y_j).
\]
The above formula makes the MMD loss both practical and effective for quantifying discrepancies between empirical measures, without requiring density estimation, a very appealing feature in high-dimensional settings. 
\revtwo{The estimator is unbiased because the diagonal terms are omitted in the
within-sample sums. Indeed, for $i\neq i'$, independence gives
\[
\mathbb{E}[k(x_i,x_{i'})]
=
\mathbb{E}_{x,x'\sim\mu}[k(x,x')]; \qquad \mathbb{E}[k(y_j,y_{j'})]
=
\mathbb{E}_{y,y'\sim\nu}[k(y,y')].
\]
For the cross term, independence of the two samples gives
\[
\mathbb{E}[k(x_i,y_j)]
=
\mathbb{E}_{x\sim\mu,y\sim\nu}[k(x,y)].
\]
}
Averaging over all off-diagonal and cross terms therefore recovers exactly
$\mathrm{MMD}^2(\mu,\nu)$.

We will use the so-called \emph{energy distance}~\cite{Sejdinovic_2013} in our numerical section.
\begin{equation}\label{eq:energyMMD}
\mathrm{D_{E}}(\mu,\nu) =
\sqrt{2\, \mathbb{E}_{x \sim \mu,\, y \sim \nu} \left[ \|x - y\| \right]
- \mathbb{E}_{x,\, \tilde{x} \sim \mu} \left[ \|x - \tilde{x}\| \right]
- \mathbb{E}_{y,\, \tilde{y} \sim \nu} \left[ \|y - \tilde{y}\| \right]|}.
\end{equation}
This corresponds to choosing the kernel $k(x, y) = -\|x - y\| + \|x\| + \|y\|$.

\subsection{Gradient Flow Equations}
We now derive a computational method for solving~\eqref{eq:min}. One of the most common approaches is gradient descent or, in the limit of infinitesimal step sizes, a gradient flow. \revtwo{A gradient is not merely a derivative; it depends on the metric structure chosen on the parameter space and on the analytical properties of the objective functional
$E$.}

Since our optimization problem is formulated over the space of probability distributions, a good choice of metric is important. For instance, taking the $L^2$ gradient flow  would yield the following partial differential equation (PDE)
\[
\partial_t \rho_\theta(\theta,t) = - \frac{\delta E(\rho_\theta)}{\delta \rho_\theta}(\theta,t)  \,,
\]
whose solution does not preserve positivity and mass conservation, which are essential structures of probability distributions. In this work, we focus on the quadratic Wasserstein distance ($W_2$) as the geometry to define gradients; see Definition~\ref{def:wp} with $p = 2$. \revtwo{Other choices may be possible, but $W_2$ provides a useful tangent-space structure on $\mathcal P_2(\mathcal X)$ through velocity fields and the continuity equation. We refer to Ambrosio, Gigli, and Savar'e~\cite{ambrosio2008gradient}, Villani~\cite{villani03}, and Santambrogio~\cite{santambrogio2015optimal} for background on Wasserstein gradient flows; related MMD gradient flows are studied in~\cite{arbel2019maximum}.}

We use $\frac{\delta E(\rho_\theta)}{\delta \rho_\theta} $ to denote the first-order variation of the objective function $E$ with respect to the parameter distribution $\rho_\theta$. \revtwo{We assume that, for $\mu_\omega$-a.e. $\omega$, the map $T_\omega$ is differentiable in $\theta$, that $\delta D/\delta\rho_y$ is differentiable in $y$, and that the resulting integrand and its $\theta$-derivative are dominated by a $\mu_\omega$-integrable function. These assumptions justify differentiating under the integral sign. The Wasserstein gradient flow is}
\begin{eqnarray} 
\partial_t \rho_\theta(\theta,t) &=& \nabla_\theta \cdot \left(\rho_\theta(\theta,t)  \nabla_\theta \frac{\delta E(\rho_\theta)}{\delta \rho_\theta} \right) \nonumber \\
&=& \nabla_\theta \cdot \left(\rho_\theta(\theta,t)  \nabla_\theta \left( \int \frac{\delta D}{\delta \rho_y} (T_{\homega}(\theta))  d\mu_\omega(\homega)  \right) \right) \nonumber \\
&=&   \nabla_\theta  \cdot \left(\rho_\theta(\theta,t) \left(  \int    \nabla T_{\homega}^\top  \,\,  \nabla_y \frac{\delta D}{\delta \rho_y} (T_{\homega}(\theta))   d \mu_\omega(\homega)   \right)   \right)  \,,\label{eq:W2_GF}
\end{eqnarray} 
where $\frac{\delta D}{\delta \rho_y}$ is the first-order variation of $D(\rho_y, \rho_y^\delta)$ with respect to $\rho_y$, which can be seen as a function of $y$, and $ \nabla T_{\homega}^\top $ is the transpose of the Jacobian of the forward map $T_{\homega}$ for a fixed realization of $\homega$.
Moreover, Equation~\eqref{eq:W2_GF} is the continuity equation with the velocity  
$$- \int    \nabla T_{\homega}^\top  \,\,  \nabla_y \frac{\delta D}{\delta \rho_y} (T_{\homega}(\theta))   d \mu_\omega(\homega) .
$$  Thanks to the equivalence between Lagrangian and Eulerian perspectives on fluid modeling:
\[
\frac{dx}{dt} = v(x,t)\quad \text{(Lagrangian ODE)}\qquad \text{and}\qquad \partial_t \rho + \nabla \cdot (v \rho) = 0 \quad \text{(Eulerian PDE)},
\]
Equation~\eqref{eq:W2_GF} naturally gives rise to a deterministic ensemble particle method for numerical simulation:
\begin{equation}\label{eq:particle}
\frac{d}{dt} \theta_i(t) = - \int \nabla T_{\homega}^\top \, \nabla_y \frac{\delta D}{\delta \rho_y}\left(T_{\homega}\left(\theta_i(t)\right)\right) \, d\mu_\omega(\homega), \quad 1 \leq i \leq N,
\end{equation}
where $\rho_\theta(\theta,t) \approx \frac{1}{N} \sum_{i=1}^N \delta_{\theta_i(t)}(\theta) =: \widehat{\rho}_\theta$. Equation~\eqref{eq:particle} is classified as an ensemble particle method because $\frac{\delta D}{\delta \rho_y}$ depends on $F(\widehat{\rho}_\theta)$, which incorporates contributions from all particles $\{\theta_i\}_{i=1}^N$.
We stress that we use ``particles'' to refer to specific structures used in our optimization algorithm, and NOT to refer to images of proteins (a common nomenclature in biological cryo-EM).
 
Here is a concrete numerical algorithm to solve the gradient flow equation~\eqref{eq:W2_GF} using the particle method~\eqref{eq:particle} based on a simple forward Euler time-stepping scheme:
\smallskip
\begin{itemize}
    \item[Step 1] Draw $N$ samples $\{\theta_i(0)\}_{i=1}^N$ from the initial parameter distribution $\rho_{\theta}(\theta,0)$. Set time domain step $\Delta t$ (also called learning rate).
    \item[Step 2] Solve Equation~\eqref{eq:particle} forward in time and obtain particles $\{\theta_i^{n}\}_{i=1}^N$ where $\theta_i^n\approx \theta_i(n \Delta t)$ at any fixed $n \in \mathbb{N}$. The obtained empirical measure $$\frac{1}{N} \sum_{i=1}^N \delta_{\theta_i^n} (\theta)
    $$ approximates the PDE solution $\rho_\theta(\theta, n \Delta t)$ in the distributional sense, subject to numerical error, random error and potential (density) estimation error.
\end{itemize}

\subsection{Implementation Details}

There are a few key terms in the forward operator evaluation~\eqref{eq:ip} and the particle formulation~\eqref{eq:particle} that require careful attention.

\subsubsection{The Fr\'echet Derivative}
The term $\frac{\delta D}{\delta \rho_y} = \frac{\delta
D(\rho_y, \rho_y^\delta)}{\delta \rho_y}$ is the Fr\'echet derivative of the discrepancy $D$ appearing in~\eqref{eq:min} with respect to $\rho_y$.  Different choices of $D$ to quantify the difference between probability distributions lead to different concrete forms of $\frac{\delta D}{\delta \rho_y}$:
\begin{itemize}
\item $D$ is the KL-divergence: 
\[
\frac{\delta D}{\delta \rho_y}(y)  = \log\frac{\rho_y}{\rho_y^\delta} (y)+ 1\,.
\]
  
\item 
$D$ is the square of the MMD given in Equation~\eqref{eq:MMD} for a positive definite kernel $k(x, y)$.
The Fr\'echet derivative then becomes
\[
\frac{\delta D}{\delta \rho_y} (y)  = \int k(y,y') d\rho_y(y') -  \int k(y,y') d\rho_y^\delta(y').
\]
One significant benefit of MMD over the KL divergence is the fact that \textit{one does not need density estimation} if $\rho_y^\delta$ and $\rho_y$ are supplied in terms of particles.   Indeed,  if $\rho_y = \frac{1}{N} \sum_{i=1}^N \delta_{y_i}(y)$, $\rho_y^\delta = \frac{1}{M} \sum_{j=1}^M \delta_{y_j^*}(y)$,  we may use the following approximation:
\begin{eqnarray*}
D(\rho_y,\rho_y^\delta) &\approx & \frac{1}{2N^2} \sum_{i\neq i'}^N k(y_i,y_{i'} ) +   \frac{1}{2M^2}\sum_{j\neq j'}^M k(y_j^*,y_{j'}^* ) - \frac{1}{NM} \sum_{i}^N\sum_{j}^M k(y_i,y_{j}^* ) \\
\frac{\delta D}{\delta \rho_y} (y)  &\approx & \frac{1}{N} \sum_{i=1}^N k(y,y_i ) -   \frac{1}{M} \sum_{j=1}^M k(y,y_j^* )\,.
\end{eqnarray*}
\end{itemize}

\subsubsection{The Adjoint Operator}
 The term $\nabla T_{\homega}^\top$ is the adjoint of the Jacobian of the forward map $T_{\homega}$. In practice, we do not need to form or store the full Jacobian. To evaluate the Wasserstein gradient flow in~\eqref{eq:W2_GF}, we only need the action of this adjoint Jacobian on the vector $\nabla_y\frac{\delta D}{\delta\rho_y}\left(T_{\homega}(\theta)\right).$ This is a vector-Jacobian product, which can be computed efficiently using reverse-mode automatic differentiation. In our implementation, we use \texttt{jax.vjp} from JAX~\cite{jax2018github} to evaluate
$\nabla T_{\homega}^\top
\nabla_y\frac{\delta D}{\delta\rho_y}\left(T_{\homega}(\theta)\right)$ directly, without explicitly constructing the Jacobian matrix.

\subsubsection{Integration With Respect To \texorpdfstring{$\mu_\omega$}{mu\_omega}}
Both the forward map~\eqref{eq:ip} and the particle dynamics~\eqref{eq:particle} require integration with respect to $\mu_\omega$. This is a key difference between our cryo-EM-based SIP and deterministic push-forward SIPs~\cite{breidt2011measure,li2024differential}. This additional integration operation could increase the sampling and computational costs by using the Monte Carlo method or other numerical quadrature rules. For example, 
\[
- \int    \nabla T_{\homega}^\top  \,\,  \nabla_y \frac{\delta D}{\delta \rho_y} (T_{\homega}(\theta_i))   d \mu_\omega(\homega)  \approx - \frac{1}{K}\sum_{k=1}^K \nabla T_{\homega_{ik}}^\top  \,\,  \nabla_y \frac{\delta D}{\delta \rho_y} (T_{\homega_{ik}}(\theta)) \,,  
\]
which implies that for each particle $\theta_i$, we additionally require $K$ samples $\{\homega_{ki}\}_{k=1}^K$, as well as $K$ evaluations of $\nabla T_{\homega}^\top$ and $T_{\homega}(\theta)$. To improve computational efficiency, we explore several sampling strategies to manage the integration with respect to $\mu_\omega$.

After applying the forward Euler scheme in time for~\eqref{eq:W2_GF}, multiplying both sides by an arbitrary test function $\phi(\theta):\Theta\rightarrow \mathbb{R}$, and \revtwo{integrating} over $\Theta$, we obtain
\begin{eqnarray*}
\int \phi({\hat{\theta}}) d\rho({\hat{\theta}}, t_{n+1} ) &=& \int \phi({\hat{\theta}}) d\rho({\hat{\theta}}, t_{n} ) - \Delta t    \int  \nabla_{\hat{\theta}} \phi \cdot \int R({\hat{\theta}},\homega) d\rho({\hat{\theta}}, t_{n}) d\mu_\omega(\homega)  ) \\
&=& \iint \left(\phi -  \Delta t\,  \nabla_{\hat{\theta}} \phi \cdot R({\hat{\theta}},\homega) \right) d\rho({\hat{\theta}}, t_{n}) d\mu_\omega(\homega)  \\
&=& \iint \varphi({\hat{\theta}},\homega)\, d\rho({\hat{\theta}}, t_{n}) d\mu_\omega(\homega)   \,,
\end{eqnarray*}
where 
\begin{equation}\label{eq:R}
    R({\hat{\theta}},\homega)  :=   \nabla_{\hat{\theta}} T_{\homega}^\top  \,\,  \nabla_y \frac{\delta D}{\delta \rho_y} (T_{\homega}({\hat{\theta}}))
\end{equation} 
and $ \varphi({\hat{\theta}},\homega) := \phi -  \Delta t\,  \nabla_{\hat{\theta}} \phi \cdot R({\hat{\theta}},\homega)$. Similarly, multiplying the forward operator~\eqref{eq:ip} both sides by an arbitrary  test function $\psi:Y \rightarrow \mathbb{R}$ and integrating over $Y$, we have
\begin{eqnarray*}
\int \psi(y) \rho_y dy &=& \iint \psi(y) d \left( T_{\homega}{\ }_\# \rho_\theta \right)(y) dy\,  d\mu_{\omega}(\homega)  \\
&=& \iint \psi(T_{\homega}({\hat{\theta}})) d\rho_\theta({\hat{\theta}}) \,  d\mu_{\omega}(\homega) \\
&=& \iint \tilde{\varphi}({\hat{\theta}},\homega) \, d\rho_\theta({\hat{\theta}}) d\mu_{\omega}(\homega)\,,
\end{eqnarray*}
where $\tilde{\varphi}({\hat{\theta}},\homega):=\psi(T_{\homega}({\hat{\theta}}))$.  Therefore, both evaluating the forward operator according~\eqref{eq:ip} and updating the gradient based on~\eqref{eq:particle}  can be seen as sampling from a joint distribution $\rho_\theta\otimes \mu_{\omega}$ of two independent random variables. (Recall we assume that the law of random variable $\omega$ is independent of the parameter distribution.)

There are multiple ways of approximating the expectation of a general $\varphi$ under the distribution $\rho_\theta\otimes \mu_{\omega}$  based on the Monte Carlo method.  To illustrate how the choice of sampling scheme can affect the error, we consider three possibilities.
\begin{eqnarray}
   I &=&  \iint \varphi({\hat{\theta}},\homega)\, d\mu_{\omega}(\homega) d\rho_\theta({\hat{\theta}}) \nonumber \\
   &\stackrel{\text{\#1}} 
   {\approx}& \frac{1}{N} \sum_{i=1}^N  \varphi({\hat{\theta}}_i,\homega_i)=: \hat{I}_1,\quad \{({\hat{\theta}}_i,\homega_i)\} \stackrel{\text{i.i.d.}}{\sim}\rho_{\hat{\theta}}\otimes \mu_{\omega}   \label{eq:method1}\\
    &\stackrel{\text{\#2}}{\approx}& \frac{1}{N}\frac{1}{K} \sum_{i=1}^N\sum_{k=1}^K \varphi({\hat{\theta}}_i,\homega_k)=: \hat{I}_2, \quad \{{\hat{\theta}}_i\}\stackrel{\text{i.i.d.}}{\sim}\rho_\theta \quad \{\homega_k \}\stackrel{\text{i.i.d.}}{\sim}\mu_{\omega}\label{eq:method2}\,,\\
    &\stackrel{\text{\#3}}{\approx}& \frac{1}{N}\frac{1}{K} \sum_{i=1}^N\sum_{k=1}^K \varphi({\hat{\theta}}_i,\homega_{ik})=: \hat{I}_3, \quad \{{\hat{\theta}}_i\}\stackrel{\text{i.i.d.}}{\sim}\rho_\theta \quad \{\homega_{ik} \}\stackrel{\text{i.i.d.}}{\sim}\mu_{\omega}\label{eq:method3}\,.
\end{eqnarray}
In Method $1$ given in~\eqref{eq:method1}, we sample from the joint law $ \rho_\theta\otimes \mu_{\omega}$, requiring $N$ samples of both the structure and noise parameters. In Method $2$ presented in~\eqref{eq:method2}, all ${\hat{\theta}}_i$'s share the same set of samples $\{\homega_k\}_{k=1}^K \sim \mu_\omega$, requiring $N$ sampled structure parameters and  $K$ sampled noise parameters. In Method $3$ detailed in~\eqref{eq:method3},  every parameter sample ${\hat{\theta}}_i$ has its own set of samples $\{\homega_{ik}\}_{k=1}^K$ from $\mu_\omega$, requiring $N$ sampled structure parameters and $NK$ sampled noise parameters.

\revtwo{Under the sampling assumptions stated in~\eqref{eq:method1}--\eqref{eq:method3}, the three estimators are unbiased. Indeed,
\[
\mathbb{E}[\hat I_1]=\mathbb{E}_{(\hat\theta,omega)\sim\rho_\theta\otimes\mu_\omega}[\varphi(\hat\theta,\omega)]= I.
\]
For Methods~$2$ and~$3$, conditioning on the samples $\{\hat\theta_i\}_{i=1}^N$ gives 
\[
\mathbb{E}[\hat I\mid \hat\theta_1,\ldots,\hat\theta_N]=\frac{1}{N}\sum_{i=1}^N\mathbb{E}_{\omega\sim\mu_\omega}
[\varphi(\hat\theta_i,\omega)],
\]
and therefore $\mathbb{E}[\hat I]=I$ by another application of the tower
property}.

The variances of the estimators $\hat{I}_1$, $\hat{I}_2$ and $\hat{I}_3$
are the following:
\begin{eqnarray*}
\text{Var}[\hat{I}_1] &=& \frac{1}{N} \text{Var}_{({\hat{\theta}},\homega)\sim \rho_\theta\otimes \mu_{\omega}}[\varphi({\hat{\theta}},\homega)], \\
\text{Var}[\hat{I}_2] &=&  \frac{1}{NK} \text{Var}_{({\hat{\theta}},\homega)\sim \rho_\theta\otimes \mu_{\omega}}[\varphi({\hat{\theta}},\homega)] + \left (\frac{1}{N}-\frac{1}{NK}\right) \text{Var}_{{\hat{\theta}}\sim \rho_\theta}\left[\mathbb{E}_{\homega\sim \mu_{\omega}}\left[\varphi({\hat{\theta}},\homega)\right]\right]  \\
   && + \left (\frac{1}{K}-\frac{1}{NK}\right) \text{Var}_{\homega\sim\mu_\omega}\left[\mathbb{E}_{{\hat{\theta}}\sim\rho_\theta}\left[\varphi({\hat{\theta}},\homega)\right]\right]\,,\\
\text{Var}[\hat{I}_3] &=& \frac{1}{NK} \text{E}_{{\hat{\theta}}\sim \rho_\theta}\left[\text{Var}_{\homega\sim \mu_{\omega}}\left[\varphi({\hat{\theta}},\homega)\right]\right]  + \frac{1}{N}\text{Var}_{{\hat{\theta}}\sim \rho_\theta}\left[\mathbb{E}_{\homega\sim \mu_{\omega}}\left[\varphi({\hat{\theta}},\homega)\right]\right].
\end{eqnarray*}

The most efficient estimator depends on the relative cost of sampling $\hat{\theta}$, sampling $\hat{\omega}$, and evaluating $\varphi$. In this work, we use Method~1 for simplicity, specifically using Equation~\eqref{eq:forward_approx} for forward evaluation and Equation~\eqref{eq:grad_approx} for approximating the gradient dynamics:
\begin{align}
    &\rho_y \approx \frac{1}{N}\sum_{i=1}^N\delta_{y_i}, \qquad\,\,\, y_i = T_{\homega_i}({\hat\theta}_i), &\,\,\{\homega_i\}_{i=1}^N  \stackrel{\text{i.i.d.}}{\sim} \mu_\omega,\,\, \{{\hat\theta}_i\}_{i=1}^N \stackrel{\text{i.i.d.}}{\sim} \rho_\theta \label{eq:forward_approx} \\
   &\rho_{\theta}^{n+1} \approx  \frac{1}{N}\sum_{i=1}^N\delta_{{\hat\theta}_i^{n+1}}, \,\,\, {\hat\theta}_i^{n+1} = {\hat\theta}_i^n - \Delta t\, R({\hat\theta}_i^n,\homega_i), &\,\,\{\homega_i\}_{i=1}^N  \stackrel{\text{i.i.d.}}{\sim} \mu_\omega,\,\, \{{\hat\theta}_i^n\}_{i=1}^N \stackrel{\text{i.i.d.}}{\sim} \rho_\theta^n\label{eq:grad_approx}
\end{align}
where $R({\hat\theta},\homega)$ is defined in~\eqref{eq:R}, and $\rho_\theta^{n}$ is the short-hand notation for $\rho_\theta({\theta}, n \Delta t)$.

\section{Optimize-Then-Discretize (OTD) and Discretize-Then-Optimize (DTO) Approaches}\label{sec:OTD-DTO}

Optimization problems over infinite-dimensional spaces, including PDE-constrained optimization, optimal control, inverse problems, and the stochastic inverse problem considered here, are commonly approached in two ways: optimize-then-discretize (OTD) and discretize-then-optimize (DTO).

In an OTD approach, one first derives the continuous optimality conditions and then discretizes the resulting equations to obtain a numerical method. This preserves the analytical properties of the original problem, including symmetries and suitable function spaces. 
It can also lead to solutions that are more accurate and amenable to mathematical analysis~\cite{hager2000runge,burkardt2002insensitive,
abraham2004effect}. In particular, the Wasserstein geometry provides a natural route from the measure-valued variational problem to an interacting particle method.

However, OTD methods still require careful discretization. The continuous gradient flow may involve quantities that are difficult to evaluate from samples. For example, if the discrepancy $D$ in~\eqref{eq:min} is chosen to be an $f$-divergence, then evaluating the gradient generally requires estimating the image densities $\rho_y$ and $\rho_y^\delta$. This difficulty can be avoided by using sample-based discrepancies such as MMD. Moreover, our gradient descent algorithm~\eqref{eq:particle} based on the gradient flow  equation~\eqref{eq:W2_GF} may not be numerically consistent with the forward map evaluation~\eqref{eq:ip} unless careful attention has been paid to the discretization of both~\eqref{eq:ip} and~\eqref{eq:W2_GF}.

In contrast, the DTO approach first starts by directly discretizing the continuous problem and then solving the resulting finite-dimensional optimization problem. DTO schemes may suffer from a lack of convergence theory when the dimension of the parameterized unknown approaches infinity and an inconsistency with the original infinite-dimensional optimization problem. Nevertheless, they are often easy to derive, allowing the use of various numerical optimization techniques. To help make the DTO approach concrete, we give an example below and use it to re-derive a popular data analysis algorithm for cryo-EM data.

\subsection{Connection with Maximum a Posteriori (MAP) Approaches}\label{ssec:relion_connection}

Maximum a posteriori (MAP) estimation is one of the standard approaches for cryo-EM reconstruction~\cite{Scheres2012bayes}. Algorithms such as RELION~\cite{kimanius2021new,scheres2012relion} and cryoSPARC~\cite{punjani2017cryosparc} are commonly motivated from likelihood-based or Bayesian models. Here we show that objectives of this type, including the form appearing in~\cite[Eqn.~(2)]{kimanius2021new}, can also be obtained by applying a discretize-then-optimize approximation to a variational problem over conformational distributions.

To begin, we will attempt to optimize $\rho_\theta$ by minimizing the KL divergence between the predicted and observed data distribution with an added regularization term.
\begin{equation}\label{eq:relion_cont}
    \mathcal{L}(\rho_\theta) =   D_{\mathrm{KL}} (\rho_y^\delta || \rho_y) -  \lambda 
 \mathbb{E}_{\rho_\theta }\left[\log \rho_{\mathrm{p}} \right] =   \int \log\frac{\rho_y^\delta}{\rho_y}(y) d\rho_y^\delta (y) -  \lambda \int \log \rho_{\mathrm{p}} (\hat\theta) d \rho_\theta (\hat\theta),
\end{equation}
where $\rho_{\mathrm{p}}$ is the prior guess for the probability density of $\theta$.
To derive an expression for $\rho(y)$, we assume a specific image formation model.
We parameterize each structure by the electrostatic potential it exerts in three-dimensional space. For concreteness, let  $\theta$ lie in a finite-dimensional subspace of $L^2(\RR^3)$, and for simplicity, we will refer to $\theta$ as a vector. 
Applying the cryo-EM forward operator~\eqref{eq:cryo-EM-eqn}, the law of $y$, denoted by $\rho_y$, \revise{is given by}
\begin{equation}\label{eq:relion_fwd} 
   \rho_y =  F(\rho_\theta) =  \int T_{\hat\theta} \,_{\#}\mu_\omega  \,  d\rho_\theta(\hat\theta)    = \frac{1}{(\sqrt{2\pi}\sigma)^{d_y}}\iint \exp\left(-\frac{|y - H^{\widehat{\omega_r}} \, \hat{\theta}|^2}{2\sigma^2}\right)d \rho_{\omega_r}(\widehat{\omega_r})  d \rho_\theta (\hat\theta)\,.
\end{equation}
Here, ${d_y}$ is the dimension of $y$. 
Plugging~\eqref{eq:relion_fwd} into~\eqref{eq:relion_cont} and dropping the constant independent of $\rho_\theta$, we have an equivalent objective functional
\begin{equation}\label{eq:relion_cont_2}
\mathcal{L}(\rho_\theta) =   - \int \log \left(\iint \exp\left(-\frac{|y - H^{\widehat{\omega_r}} \, \hat{\theta}|^2}{2\sigma^2}\right)d \rho_{\omega_r}(\widehat{\omega_r}) \, d \rho_\theta(\hat{\theta}) \right)  d\rho_y^\delta - \lambda \int \log \rho_{\mathrm{p}} (\hat\theta) d \rho_\theta (\hat\theta)\,.
\end{equation}
Often, we observe the reference data distribution $\rho_y^\delta$ through finitely many samples, i.e.,
$\rho_y^\delta  \approx \frac{1}{N}\sum_{i=1}^N \delta_{y_i}$. Then~\eqref{eq:relion_cont_2} reduces to
\begin{equation}\label{eq:relion_cont_3}
\mathcal{L}_N(\rho_\theta) =   - \frac{1}{N} \sum_{i=1}^N \log \left(\iint \exp\left(-\frac{|y_i - H^{\widehat{\omega_r}} \, \hat{\theta}|^2}{2\sigma^2}\right)d \rho_{\omega_r}(\widehat{\omega_r}) \, d \rho_\theta(\hat{\theta}) \right)  - \lambda  \int \log \rho_{\mathrm{p}} (\hat\theta) d \rho_\theta (\hat\theta)\,.
\end{equation}

\revtwo{Assume that the observed cryo-EM images $\{y_i\}_{i=1}^N$ are i.i.d. samples
from the observed image distribution $\rho_y^\delta$, corresponding in the
model to independently sampled conformations, viewing directions, and noise realizations. Then, for any fixed $\rho_\theta$,
$\mathcal{L}_N(\rho_\theta)\to \mathcal{L}(\rho_\theta)$ almost surely as
$N\to\infty$, provided that the corresponding logarithmic integrand is
$\rho_y^\delta$-integrable. This follows directly from the strong law of
large numbers.}

Furthermore, we can also approximate $\rho_\theta$ by an empirical measure, $\rho_\theta \approx \frac{1}{K} \sum_{k=1}^K\delta_{\theta_k}$.
Then the discretized $\mathcal{L}_N$ under this particular parameter
distribution approximation becomes, again after dropping the constant,
\begin{equation}\label{eq:relion_cont_4}
    \mathcal{L}_{N,K} (\boldsymbol{\theta}) =   -\frac{1}{N} \sum_{i=1}^N \log \left(\sum_{k=1}^K\int \exp\left(-\frac{|y_i - H^{\widehat{\omega_r}} \, \theta_k|^2}{2\sigma^2}\right)d \rho_{\omega_r}(\widehat{\omega_r}) \right) -  \frac{ \lambda}{K}\sum_{k=1}^K \log \rho_{\mathrm{p}} (\theta_k) \,, 
\end{equation}
where $\boldsymbol{\theta} = (\theta_1,\ldots,\theta_K)$. We remark that~\eqref{eq:relion_cont_4} is precisely~\cite[Eqn.~(2)]{kimanius2021new} with $\lambda = \frac
KN$ and up to a normalizing constant $\frac{1}{N}$.

\subsection{Large Data Limit for DTO Schemes}\label{ssec:large_data_limit}
A key requirement for both OTD and DTO frameworks is consistency: as the discretization is refined, minimizers of the discretized problems should converge, in an appropriate sense, to minimizers of the continuous variational problem. For OTD schemes, consistency is often easier to analyze because one first derives the continuous optimality conditions and then discretizes them. For DTO schemes, the situation is more delicate, since the discretized problem is posed over a finite-dimensional search space, as in~\eqref{eq:relion_cont_4}, while the original problem is posed over an infinite-dimensional space of probability measures, as in~\eqref{eq:relion_cont}. Since many cryo-EM methods follow a DTO philosophy, we state sufficient conditions under which such discretizations are consistent with the continuous variational problem.

Recall the objective functional $E(\rho_\theta) = D(  F(\rho_\theta),\rho_y^\delta)$ in~\eqref{eq:min} where $D$ is a certain divergence or metric over the space of probability distributions. Define $$
E_N = D\left( F(\rho_\theta),\frac{1}{N}\sum_{i=1}^N\delta_{y_i}\right)
$$ where $\{y_i\}_{i=1}^N$ are samples of $\rho_y^\delta$; see~\eqref{eq:relion_cont_3} for an example. One may work with $E_N$ rather than $E$ due to either a limited access to the reference data distribution $\rho_y^\delta$ or discretization of the functional. We then have two variational problems:
\begin{eqnarray}
    &&\inf_{\rho_\theta\in \Omega \subset \mathcal{P}(\Theta)} E(\rho_\theta)\,, \label{eq:V1}\\ 
    &&\inf_{\rho_\theta\in \Omega \subset \mathcal{P}(\Theta)} E_N(\rho_\theta)\,.\label{eq:V2}
\end{eqnarray}
The convergence of~\eqref{eq:V2} to~\eqref{eq:V1} can be established through $\Gamma$-convergence~\cite{braides2006handbook}.

\begin{proposition}\label{prop:DTO-consistency}
Consider \revtwo{the} Polish space $(\mathcal{P}(\Theta), W_2)$.
Assume  that
\begin{enumerate}
    \item[\textbf{A1}] $D(\mu,\nu)$ is  jointly lower semi-continuous in $(\mu,\nu)$ with respect to the $(\mathcal{P}(Y), W_2)$ topology where $F(\Theta)\subseteq Y$, and 
    \item[\textbf{A2}] $\forall\lambda \in \mathbb{R}$,  $\exists K_\lambda\subset \Omega$ compact such that $\{\rho_\theta\in \Omega: E_N (\rho_\theta) \leq \lambda\}\subset K_\lambda$ for all $N$, and
    \item[\textbf{A3}] $E_N(\rho_\theta)\xrightarrow{N\rightarrow \infty} E(\rho_\theta)$ almost surely for any $\rho_\theta \in \Omega$.
\end{enumerate}
Then the following equation holds almost surely
\begin{equation}\label{eq:N_conv}
    \min_{\rho_\theta\in \Omega \subset \mathcal{P}(\Theta)} E(\rho_\theta) = \lim_{N\rightarrow \infty }\min_{\rho_\theta\in \Omega \subset \mathcal{P}(\Theta)} E_N(\rho_\theta).
\end{equation}
\end{proposition}
\begin{proof}
    First, note that the forward map~\eqref{eq:relion_fwd} based on~\eqref{eq:cryo-EM-eqn} is a linear push-forward map with a bounded Lipschitz constant. We have
\begin{equation}\label{eq:Lips}
    W_2(F(\rho_\theta), F(\rho_\theta')) \leq C W_2(\rho_\theta, \rho_\theta').
      \end{equation}
    Consequently, given any converging sequence $\{\rho_\theta^j\}_{j=1}^\infty \subset \Omega$ to $\rho_\theta$, we obtain a converging sequence $\{\rho_y^j\}_{j=1}^\infty \subset Y$ to $\rho_y = F(\rho_\theta)$. By Assumption~\textbf{A1},
    $$ 
    D(F(\rho_\theta), \nu)  = D(\rho_y, \nu) \leq  \liminf_{j} D(\rho_y^j, \nu) =  \liminf_{j} D(F(\rho_\theta^j), \nu)\,,
    $$
     for any $\nu$. When $\nu = \rho_y^\delta$ and $\nu=\frac{1}{N}\sum_{i=1}^N \delta_{y_i}$, 
    we obtain that $E$ and $E_N$ are lower semi-continuous, respectively. Since Assumption~\textbf{A2} implies the coercivity of $E_N$, together with the lower semi-continuity of $E_N$, $\forall N$, we know minimizers exist for the sequence of variational problems given in~\eqref{eq:V2}. 

    Next, we verify the $\Gamma$-convergence of $E_N$ to $E$. First, for every sequence $\{\rho_\theta^N\}_{N=1}^\infty$ converging to $\rho_\theta$, we have also $\{\rho_y^N\}_{N=1}^\infty$ converging to $\rho_y$ as a result of~\eqref{eq:Lips} where $\rho_y^N = F(\rho_\theta^N)$ and $\rho_y = F(\rho_\theta)$. Since $
    E_N(\rho_\theta^N) = D(\rho_y^N, \hat{\rho}^N_y)$ where $\hat{\rho}^N_y :=  \frac{1}{N}\sum_{i=1}^N \delta_{y_i}$, 
$\{\rho_y^N\}_N \xrightarrow{N\rightarrow\infty} \rho_y$ and  the empirical measure $\{\hat{\rho}^N_y\}_N \xrightarrow{N\rightarrow\infty}\rho_y^\delta$ almost surely~\cite[Thm.~11.4]{dudley2018real}, we have
 \[
  E(\rho_\theta) =  D(\rho_y, \rho^\delta_y)  \leq  \liminf_N D(\rho_y^N, \hat{\rho}^N_y) = \liminf_N E_N(\rho_\theta^N)   
 \]
as a result of the joint lower semi-continuity by~\textbf{A1}. So far, we have verified the ``liminf'' condition of the $\Gamma$ convergence. The ``existence of a recovery sequence'' directly holds because of Assumption~\textbf{A3}, which is equivalent to SLLN. Hence, $\{E_N\}_N$ $\Gamma$-converges to $E$.
 
Finally, by the Fundamental Theorem of $\Gamma$-convergence~\cite[Thm.~2.10]{braides2006handbook}, we get~\eqref{eq:N_conv}.
\end{proof}

\revtwo{We briefly comment on Assumptions \textbf{A1}--\textbf{A3} for the discrepancies used in this paper. For the MMD objective
\[
    E(\rho_\theta)=\mathrm{MMD}_k(F(\rho_\theta),\rho_y^\delta),
    \qquad
    E_N(\rho_\theta)=\mathrm{MMD}_k(F(\rho_\theta),\widehat\rho_y^N),
\]
Assumption~\textbf{A1} holds whenever $k$ is bounded and continuous, since MMD is continuous under weak convergence, and hence lower semicontinuous under $W_2$-convergence. Moreover,
\[
\left|
\mathrm{MMD}_k(F(\rho_\theta),\widehat\rho_y^N)
-
\mathrm{MMD}_k(F(\rho_\theta),\rho_y^\delta)
\right|
\leq
\mathrm{MMD}_k(\widehat\rho_y^N,\rho_y^\delta),
\]
so \textbf{A3} follows from the law of large numbers for kernel mean embeddings. The same argument applies to the energy distance, viewed as a distance-induced MMD, provided the image laws have finite first moments.
}
\revtwo{
For the KL-based formulation, the finite-data objective is interpreted as the empirical negative log-likelihood
\[
    E_N^{\rm KL}(\rho_\theta)
    =
    -\frac1N\sum_{i=1}^N \log q_{\rho_\theta}(y_i)
    +
    R(\rho_\theta),
\]
where $q_{\rho_\theta}$ is the density of $F(\rho_\theta)$. Under Gaussian image noise, compact computational domains, continuity of the forward map, and a uniform positive lower bound on $q_{\rho_\theta}$, the corresponding population cross-entropy is lower semicontinuous and $E_N^{\rm KL}\to E^{\rm KL}$ by a uniform law of large numbers on compact sublevel sets.
}

\revtwo{Finally, \textbf{A2} is a compactness/coercivity assumption. It is automatic if $\Theta$ is compact and $\Omega\subset\mathcal P_2(\Theta)$ is closed. Otherwise it can be enforced by restricting to measures with bounded second moments or by adding a coercive prior/regularization term.
}

Next, we study the convergence in terms of the particle discretization of the parameter distribution $\rho_\theta$. Recall $\mathcal{P}(\Theta)$ in~\eqref{eq:min} represents the space of probability measures over $\Theta$. Without loss of generality, we assume that the search domain $\Omega = \mathcal{P}(\Theta)$. We denote the set of all empirical measures in $\mathcal{P}(\Theta)$ by
\[
\mathcal{P}_\text{em}(\Theta) = \left\{\rho_\theta  := \frac
1K\sum_{k=1}^K \delta_{\theta_k}: \, \theta_1,\ldots,\theta_K \in \Theta,\,\forall K\in\mathbb{N} \right\}\,,
\]
and the set of empirical measures consisting of $K$ particles by
\[
\mathcal{P}_\text{em}^K(\Theta) = \left\{\rho_\theta  := \frac
1K\sum_{k=1}^K \delta_{\theta_k}: \, \theta_1,\ldots,\theta_K \in \Theta \right\}\,.
\]

\begin{lemma}\label{lem:K_conv}
    Consider an objective functional $E:\mathcal{P}(\Theta)\rightarrow \mathbb{R}$ that maps a probability distribution $\rho_\theta$ to a real number. Assume $E$ is continuous in $\rho_\theta$ with respect to the weak topology, and both variational problems 
    \[
    \inf_{\rho_\theta \in \mathcal{P}(\Theta)} E(\rho_\theta),\quad \inf_{\rho_\theta \in \mathcal{P}_\text{em}^K(\Theta)} E(\rho_\theta)
    \]
    admit minimizers, $\forall K\in\mathbb{N}$. We then have
    \begin{equation*}
        \lim_{K\rightarrow \infty} E(\rho_\theta^K) = E(\rho_\theta^*)\,,
    \end{equation*}
   where $\rho_\theta^K \in \argmin_{\rho_\theta \in \mathcal{P}_\text{em}^K(\Theta)}  E(\rho_\theta)$, and $\rho_\theta^* \in \argmin_{\rho_\theta \in \mathcal{P}(\Theta)}  E(\rho_\theta)$.
\end{lemma}
\begin{proof}
Since $\mathcal{P}_\text{em}(\Theta)$
is dense in $\mathcal{P}(\Theta)$ with respect to the weak topology, we have the following due to the continuity of $E$:
\[
  \inf_{\rho_\theta  \in \mathcal{P}_\text{em}(\Theta)}E(\rho_\theta) = \inf_{\rho_\theta \in \mathcal{P}(\Theta)} E(\rho_\theta)\,.
\]

Next, we prove that   
\begin{equation}\label{eq:goal}
\lim_{K\rightarrow \infty } \min_{ \rho_\theta \in \mathcal{P}_\text{em}^K(\Theta)}E  (\rho_\theta)  = \inf_{\rho_\theta  \in \mathcal{P}_\text{em}(\Theta)}E(\rho_\theta).
\end{equation}
For convenience, we use the short-hand notations: $A_K = \min_{ \rho_\theta \in \mathcal{P}_\text{em}^K(\Theta)}E  (\rho_\theta)$ and $B = \inf_{\rho_\theta  \in \mathcal{P}_\text{em}(\Theta)}E(\rho_\theta)$. Since $\mathcal{P}^K_\text{em}(\Theta) \subset\mathcal{P}_\text{em} $ for any $K$, we have $A_K \geq B$ and thus
\begin{equation}\label{eq:liminf_geq}
    \liminf_{K\rightarrow \infty} A_K \geq B. 
\end{equation}
On the other hand, based on the property of infimum, for any $\epsilon>0$, there exists $K(\epsilon) \in \mathbb{N}$ such that 
\[
A_{K(\epsilon)} \leq E  \left(\rho_\theta^{K(\epsilon)}\right) < B + \frac{\epsilon}{2}.
\]
For the given $\epsilon$ and this probability measure $\rho_\theta^{K(\epsilon)} \in \mathcal{P}_\text{em}^{K(\epsilon)}(\Theta)$, there exists $M(\epsilon)\in \mathbb{N}$ such that whenever $N\geq M(\epsilon)$, the best $N$-particle empirical measure approximation of $\rho_\theta^{K(\epsilon)}$ in the set $\mathcal{P}_\text{em}^N$, denoted by $\rho^N$, is close to $\rho_\theta^{K(\epsilon)}$ such that
\[
\Big| E  \left(\rho_\theta^{K(\epsilon)}\right) - E  \left(\rho^N\right) \Big| \leq \frac{\epsilon}{2}\,.
\]
Here, we use the assumption that $E$ is continuous in the weak topology. This leads to 
\[
A_N = \min_{ \rho_\theta \in \mathcal{P}_\text{em}^N(\Theta)}E  (\rho_\theta) \leq E  \left(\rho^N\right)\leq E  \left(\rho_\theta^{K(\epsilon)}\right)  + \frac{\epsilon}{2} \leq B + \epsilon,
\]
for any $N> M (\epsilon)$. Based on the arbitrariness of $\epsilon$, we then have 
$
\limsup\limits_{N\rightarrow\infty} A_N \leq B$.
Combining this inequality with~\eqref{eq:liminf_geq}, we obtain~\eqref{eq:goal}. Putting everything together, we have
\begin{equation*}
 \lim_{K\rightarrow \infty} E(\rho_\theta^K) = \lim_{K\rightarrow \infty } \min_{ \rho_\theta \in \mathcal{P}_\text{em}^K(\Theta)}E  (\rho_\theta) = \inf_{\rho_\theta  \in \mathcal{P}_\text{em}(\Theta)}E(\rho_\theta)  =  \min_{\rho_\theta \in \mathcal{P}(\Theta)} E (\rho_\theta) = E  (\rho_\theta^*) \,,
\end{equation*}
which finishes the proof.
\end{proof}

Note that \Cref{lem:K_conv} only states the convergence of the objective functional. To obtain the convergence of the minimizing distributions, we need extra assumptions. The following proposition provides a sufficient condition.

\begin{proposition}
    Let all conditions in~\Cref{lem:K_conv} hold and the objective functional $E$ have a unique minimizer $\rho_\theta^*$. Suppose $E$ satisfies an \L ojasiewicz-type inequality at $\rho_\theta^*$:
\begin{equation}\label{eq:PL_ineq}
d(\nu, \rho_\theta^*) \leq  C|E(\nu) -E (\rho_\theta^*)|^\alpha\,, \quad \forall\nu\in\mathcal{P}(\Theta)\,,
\end{equation}
for some constants $C,\alpha >0$, where $d(\nu, \rho_\theta^*)$ is some divergence or metric over $\mathcal{P}(\Theta)$. Then
\begin{equation*}
    \lim_{K\rightarrow \infty} d(\rho_\theta^K, \rho_\theta^*) = 0\,.
\end{equation*}
\end{proposition}
\begin{proof}
    The result is a direct consequence of~\eqref{eq:PL_ineq} and~\Cref{lem:K_conv}.
\end{proof}

\section{Local Stability, Sample Complexity, and Cryo-EM Observability}
\label{sec:stability}

\revtwo{
The cryo-EM inverse problem is inherently ill-conditioned: the forward model in~\eqref{eq:cryo-EM-eqn} includes projection, marginalization over unknown viewing directions and other nuisance variables, Contrast Transfer Function (CTF) modulation, possible symmetries, and additive noise. Consequently, the variational objective in~\eqref{eq:min} is not expected to be globally geodesically convex, even when the image-space discrepancy $D$ has favorable convexity properties. Thus, rather than proving global convergence of the implemented nonconvex particle algorithm, our goal in this section is to identify the local quantities that govern recovery.
}

\revone{
We focus on three related questions. First, when does the closeness of image distributions imply the closeness of the corresponding conformational distributions? Second, how do angular coverage, CTF effects, and noise affect the observability of perturbations to the structure? Third, how do finite-sample error, Monte Carlo approximation, particle approximation, optimization error, and model mismatch enter the final recovery error? The answers are summarized through a local inverse-stability constant, a physical observability operator, and an oracle-type recovery bound.
}

\subsection{Local inverse stability}
\label{sec:local-inverse-stability}

\revone{
We quantify local identifiability through the following inverse-stability estimate: 
we say that \(F_\sigma\) is locally stable around \(\rho^\star\), with constant
\(\gamma>0\), if for all \(\rho\) in a neighborhood of \(\rho^\star\),
\begin{equation}
    W_2(\rho,\rho^\star)
    \le
    \frac{1}{\sqrt{\gamma}}
    D(F(\rho),F(\rho^\star)),
    \label{eq:local_identifability}
\end{equation}
Here \(D\) denotes the distance-scale version of the chosen image-space discrepancy; for example, for an MMD objective one uses the MMD distance rather
than its square. The constant $\gamma>0$ measures how strongly perturbations to the structure density are visible after the cryo-EM image
formation process. Small $\gamma$ corresponds to an ill-conditioned inverse problem: distinct conformational distributions may produce very similar image distributions.

At the infinitesimal level, this constant is governed by the linearized forward map. Let
\begin{equation}
    \rho_t=(\mathrm{Id}+tv)_\#\rho^\star
    \label{eq:structure_density_perturbation}
\end{equation}
be a Wasserstein perturbation of the structure distribution. 
Here $v$ is a velocity field that perturbs $\rho^\star$ by transporting probability mass.  We give a complete definition in the Supplementary Material.

Perturbation of the structure density induces a signed perturbation of the image density.  We have
the following expression for the perturbed density.
\begin{equation}
    F(\rho_t)=q^\star+tL_{\rho^\star}v+o(t),
    \label{eq:F_perturbation}
\end{equation}
where
\begin{equation*}
    L_{\rho^\star}v
    :=
    \left.
    \frac{d}{dt}
    \right|_{t=0}
    F(\rho_t),
\end{equation*}
is a signed perturbation of the image density
$q^\star=F(\rho^\star)$.

The discrepancy $D$ determines the tangent scale in which image perturbations are measured. 
We assume that $D$ admits the local quadratic expansion. Letting $h$ be a signed perturbations with zero total mass,
\begin{equation}
    D(q^\star+\epsilon h,q^\star)^2
    =
    \epsilon^2 \|h\|_{D,q^\star}^2+o(\epsilon^2),
    \qquad
    \epsilon\to 0.
\end{equation}
Here $\|\cdot \|_{D, q^\star}$ is a $D$-dependent norm measuring the magnitude of perturbations from $q^\star$; we give expressions for this norm for KL and MMD in the Supplement.
Applying~\eqref{eq:F_perturbation}, we have
\begin{equation*}
    D(q^\star+t L_{\rho^\star}v,q^\star)^2
    =
    t^2\|L_{\rho^\star}v\|_{D,q^\star}^{2}
    +
    o(\epsilon^2),
    \qquad
    \epsilon\to0,
\end{equation*}
Moreover, \eqref{eq:structure_density_perturbation} implies $W_2 (\rho_t, \rho^\star) = t\|v\| + o(t)$.
Substituting both expressions into~\eqref{eq:F_perturbation} motivates the following linearized stability constant,
\begin{equation}
    \gamma_{\mathrm{lin}}^{(D)}
    =
    \inf_{v\neq0}
    \frac{
        \|L_{\rho^\star}v\|_{D,q^\star}^{2}
    }{
        \|v\|_{L^{2}(\rho^\star)}^{2}
    }.
\end{equation}
We note that $\gamma$ can also be interpreted as the smallest positive generalize eigenvalue of a discrepancy-dependent stability matrix; we refer the reader to Section~S1.2 of the Supplementary Material (\textbf{SM}) for more details.
}

\subsection{The Effect of the Imaging Model on Stability}
\label{sec:physical-observability}

\revone{
We now relate the observed linearized perturbation $L_{\rho^\star}v(y)$ to 
the cryo-EM acquisition model.
Recall from the image model in \eqref{eq:cryo-EM-eqn} that
\[
    y=T(\theta,\omega)=H_{\omega_r}\theta+\omega_n,
    \qquad
    \omega_n\sim N(0,\sigma^2I).
\]
To do this, we adopt a commonly used model for the physics of electron scattering in the projection and weak-phase approximations\cite{vulovic2014use}.
We assume that $\theta$ is the Fourier transform of the biomolecule's electronic scattering potential.
In this case, the imaging model is commonly expressed as
\begin{equation*}
    H_{\omega_r}=C P R_{\omega_r}
\end{equation*}
where $R_{\omega_r}$ is the rotation operator, $P$ is the projection operator along the $z$-axis
onto the image plane, and $C$ is the CTF, which we assume to be constant for simplicity. We write $\Omega_\omega$ for the space of acquisition
variables $\omega_r$, with distribution $\pi$. We also regard $\Theta$ as
the Hilbert space, or finite-dimensional structural subspace, in which
conformations and structural perturbations are represented.

As derived in Section~S1.1 of the \textbf{SM}, the perturbation to the image density can be written as
\begin{equation}
    L_{\rho^\star}v(y)
    =
    q^\star(y)\,
    \mathbb{E}_{\theta,\omega_r\mid Y=y}[A_v], \qquad A_v
    :=
    \frac{1}{\sigma^{2}}
    \left\langle C P R_{\omega_r}v(\theta), \omega_n \right\rangle_{L_{2,\mathcal Y}}.
    \label{eq:directiona_derivative}
\end{equation}
Here $\theta\sim\rho^\star$, $\omega_r\sim\pi$, and the $L_2$ inner product is
taken in the image space $\mathcal Y$. 
Observing that $\omega_n = y - H_{\omega_r} \theta$,
we can interpret $A_v$ as the response function of the image density to a perturbation of the structure density at a fixed noise realization.
Identity~\eqref{eq:directiona_derivative} shows that the observed perturbation \(L_{\rho^\star}v\) is obtained by conditioning this response on the observed image. 

The second moment of the response function is
\begin{equation}
\begin{aligned}
    \mathbb E_{\theta,\omega_r,\omega_{n}}[A_v^2]
    &=
    \frac{1}{\sigma^2}
    \int_\Theta
    \int_{\Omega_\omega}
    \left\|C P R_{\omega_r} v(\theta)\right\|_{L_2,\mathcal Y}^2
    \,d\pi(\omega_r)\,d\rho^\star(\theta) \\
    &=
    \frac{1}{\sigma^2}
    \int_\Theta
    \left\langle v(\theta),A_\pi v(\theta)\right\rangle_{L_{2,\Theta}}
    \,d\rho^\star(\theta),
\end{aligned}
\end{equation}
where
\begin{equation}
    A_\pi
    :=
    \int_{\Omega_\omega}
    R_{\omega_r}^{\ast}P^{\ast}C^{\ast}C P R_{\omega_r}
    \,d\pi(\omega_r).
\end{equation}
The map $A_\pi$ relates the imaging model to stability. Under a domination
condition on the tangent form, Section~S1.4 of the \textbf{SM} provides the upper
bound
\begin{equation}
    \gamma_{\mathrm{lin}}^{(D)}
    \le
    \frac{C_D}{\sigma^{2}}
    \lambda_{\min}(A_{\pi})
\end{equation}
on the relevant structural perturbation subspace. }
\revtwo{The acquisition law is part of the forward model and enters directly into the observability operator.
If an acquisition law causes the smallest eigenvalue of $A_\pi$ is small, then
$\gamma_{\mathrm{lin}}^{(D)}$ cannot be large. 
}

\subsection{Local error decomposition}
\label{sec:oracle-recovery-bound}
\revone{
We now combine the local inverse-stability condition with the numerical and statistical approximations used by the particle method. Let $\widehat{\rho}_{n,K,M}$ denote the distribution returned by the algorithm using $n$ observed images, $K$ particles, and $M$ independent Monte Carlo samples of the imaging variables used to approximate the random forward operator \(F\) and its gradient.
Assume that a local inverse-stability estimate holds on the relevant $K$-particle model class:
\begin{equation}
    W_2(\rho,\rho_K^\star)
    \le
    \frac{1}{\sqrt{\gamma_K^{(D)}}}
    D(F(\rho),F(\rho_K^\star)),
\end{equation}
where $\rho_K^\star$ is a best $K$-particles approximation of $\rho^\star$, and $\gamma_K^{(D)}$ is the observed stability constant restricted to the finite-dimensional perturbation class.
As derived in Section~S1.4 of the \textbf{SM}, we have the bound
\begin{equation}    W_2(\widehat{\rho}_{n,K,M},\rho^\star)
    \lesssim
    \frac{1}{\sqrt{\gamma_K^{(D)}}}
    \left[
        \epsilon_{\mathrm{stat}}(n)
        +
        \epsilon_{\mathrm{MC}}(M,K)
        +
        b_K^{\mathrm{img}}
        +
        \epsilon_{\mathrm{opt}}
        +
        \epsilon_{\mathrm{model}}
    \right]
    +
    b_K^{W_2},
\end{equation}
where the individual error terms are defined as follows.
\begin{itemize}
    \item $\epsilon_{\mathrm{stat}}(n)$ is the finite-sample error from observing only $n$ images.
    \item $\epsilon_{\mathrm{MC}}(M,K)$ is the Monte Carlo error from approximating the random forward map
    \item $b_K^{\mathrm{img}}$ is the image-space bias from restricting the structure distribution to $K$ particles
    \item $\epsilon_{\mathrm{opt}}$ is the optimization error of the implemented descent
    \item $\epsilon_{\mathrm{model}}$ is the model-mismatch error, and $b_K^{W_2}$ the best $K$-structure approximation error.
\end{itemize}
The bracketed terms are image-space errors and are therefore amplified by the inverse-stability factor $1/\sqrt{\gamma_K^{(D)}}$. In contrast, $b_K^{W_2}$ is already a latent-space approximation error and is not
multiplied by this factor.

To gain intuition about the scaling of these terms, we discuss their scaling when $D$ is chosen to be the MMD.
Empirical kernel mean concentration gives 
\begin{equation}
    \epsilon_{\mathrm{stat}}(n)=O_P(n^{-1/2}).
\end{equation}
Here \(X_n=O_P(a_n)\) means that \(X_n/a_n\) is bounded in probability: for every \(\varepsilon>0\), there exists \(C_\varepsilon<\infty\) such that
\[
    \mathbb P(|X_n|>C_\varepsilon a_n)<\varepsilon
\]
for all sufficiently large \(n\).

A uniform Monte Carlo estimate over a $K$-particle class gives the  scaling
\begin{equation}
    \epsilon_{\mathrm{MC}}(M,K)
    =
    O_P\left(
        \sqrt{\frac{dK\log M}{M}}
    \right),
\end{equation}
This estimate uses that the kernel is bounded, that the random feature map
\[
    (\theta,\omega)\mapsto k(T_\omega(\theta),\cdot)
\]
is uniformly Lipschitz in \(\theta\) on the domain of admissible particles, and that the \(K\)-particle class has effective dimension \(dK\). These assumptions allow a concentration bound for each fixed \(K\)-particle distribution, followed by a covering-number argument over the \(K\)-particle class.

If the target latent distribution has effective dimension $d$, then the best $K$-particle Wasserstein approximation has the quantization scale
\begin{equation}
    b_K^{W_2}=O(K^{-1/d}),
\end{equation}
and if $F$ is locally Lipschitz, the corresponding image-space bias satisfies
\begin{equation}
    b_K^{\mathrm{img}}=O(K^{-1/d}).
\end{equation}
The derivation and assumptions behind these estimates are given in Sections~S1.5 and~S1.6  of the \textbf{SM}. Combining the representative rates gives the recovery estimate
\begin{equation}
    W_2(\widehat{\rho}_{n,K,M},\rho^\star)
    \lesssim
    \frac{1}{\sqrt{\gamma_K^{(D)}}}
    \left[
        n^{-1/2}
        +
        \sqrt{\frac{dK\log M}{M}}
        +
        K^{-1/d}
        +
        \epsilon_{\mathrm{opt}}
        +
        \epsilon_{\mathrm{model}}
    \right]
    +
    K^{-1/d}.
\end{equation}

The remaining quantity is the local stability constant $\gamma_K^{(D)}$, which depends on both the data discrepancy and the physical projection operator. }

\revtwo{The present analysis does not prove a mean-field limit for the particle
dynamics. Such a result would require Dobrushin-type stability estimates for
the velocity field $V[\rho]$ induced by the image discrepancy and the random
cryo-EM forward \cite{dobrushin1979vlasov}, which we leave for future work.}

\section{Numerical Examples}\label{sec:numerics}

We present numerical experiments illustrating the proposed Wasserstein-gradient-flow particle method for stochastic inverse problems arising in cryo-EM. The examples increase in complexity: a one-dimensional inverse problem with additive noise, a two-dimensional nanocluster model with high-dimensional images, and a simulated cryo-EM experiment based on normal-mode conformational variability of HSP90. We also compare the particle method with Gaussian-mixture baselines and study the effect of particle number and image noise in a controlled one-mode protein setting.

Unless otherwise stated, the conformational distribution is represented by particles,
\[
\rho_\theta
\approx
\frac1K\sum_{k=1}^K\delta_{\theta_k},
\]
and the image discrepancy is evaluated using sample-based losses such as the energy distance~\eqref{eq:energyMMD}. \revtwo{Gradients are computed by automatic differentiation through the forward imaging model, and the particle updates follow the discretized form of~\eqref{eq:grad_approx}. In practice, we use ADAM to stabilize the stochastic first-order updates, since the objective is evaluated through Monte Carlo samples of the forward map and the resulting gradients can be noisy and differently scaled across latent coordinates~\cite{kingma2015adam}. Similar stochastic first-order optimization strategies are common in distributional learning and cryo-EM heterogeneity methods~\cite{genevay2018learning,kolouri2018sliced,arbel2019maximum,Zhong2021}. Sample code for the one-dimensional, nanocluster, and protein normal-mode
experiments is available at
\url{https://github.com/thiedelab/Wasserstein_cryo.git}.}

\begin{algorithm}
\caption{Gradient descent with image matching and Jacobian-vector products\label{alg:alg}}
\begin{algorithmic}
\STATE \textbf{Input:} Observed images $y^{\delta}$, initial modes $\theta^0 = \{\theta_j^0\}_{j=1}^M$, noise level $\sigma$, imaging parameters, protein parameters, learning rate vector $\eta$
\STATE \textbf{Output:} Optimized latent variables $\theta^{\text{final}}$
\FOR{iteration $l = 0, 1, 2, \ldots, N_{\max}$}
    \STATE Generate random noise $\{\hat{\omega}_{n_{j}}\}_{j=1}^N$ from $\mathcal{N}(0, \sigma^2 I)$
    \STATE Generate random orientations $\{\hat{\omega}_{r_j}\}_{j=1}^N$ from uniform quaternion distribution
    \STATE Set $y_j^l = T_{\hat{\omega}_j}(\theta^{l}_j)$ with $\hat{\omega}_j = (\hat{\omega}_{n_{j}}, \hat{\omega}_{r_{j}})$ for $j = 1,\dots,N$
    \STATE Compute $\frac{d}{dt} \theta_j(t)$ according to Equation~\eqref{eq:particle}
    \STATE Update $\theta_j$ using Adam: $\theta_j^{l+1} \leftarrow \text{Adam}\left(\frac{d}{dt} \theta_j(t), \theta_j^l,\eta\right)$
\ENDFOR
\STATE \textbf{return} $\theta^{\text{final}} = \theta^{\text{maxiter}}$
\end{algorithmic}
\end{algorithm}

\subsection{One-dimensional Test System}

We begin with a one-dimensional test problem designed to illustrate the core ideas of the framework and to compare two image-distribution discrepancies: KL divergence and energy distance. The forward model is the identity map with additive Gaussian noise, 
\[
T_{\omega}(\theta) = \theta + \omega, \quad \theta \in \mathbb{R} \quad \omega \sim \mathcal{N}(0, \sigma^2),
\]
where $\omega$ represents additive noise and $\sigma=1.5$. The true parameter distribution is defined as a mixture of two Gaussians:
\[
\theta \sim \frac{1}{2} \cdot \mathcal{N}(-2, 0.75^2) + \frac{1}{2} \cdot \mathcal{N}(2, 0.3^2).
\]
The observed data are generated by adding independent noise to the samples:
\[
y^{\delta} = y + \delta, \quad \delta \sim \mathcal{N}(0, \sigma^2).
\]
We generate $n=10^4$ observations from this model and run Algorithm~\ref{alg:alg} for $25000$ optimization iterations.

For the energy distance, the distributions are represented directly by empirical measures supported on the particle locations. For the KL divergence, the densities $\rho_{\theta}$, $\rho_y$, and $\rho_y^\delta$ are approximated by kernel density estimation with a Gaussian kernel~\cite{chen2017tutorialkerneldensityestimation},
\[
\rho_{y}^{N}(y) \approx \frac{1}{N}\sum_{j=1}^{N} \phi^{\epsilon}(y - y_j), \quad \phi^{\epsilon}(y) = \frac{1}{(2\pi\epsilon)^{1/2}} \exp\left( -\frac{y^2}{2\epsilon} \right).
\]

The KDE bandwidth for the KL objective is chosen using Silverman's rule of thumb, which gives $\epsilon=0.3496$ for this dataset. The initial latent distribution is $\rho_{\theta_0}=\mathcal N(0,1)$, represented by $10^4$ i.i.d. particles. We then apply the particle method in~\eqref{eq:particle} to approximate the Wasserstein gradient flow~\eqref{eq:W2_GF}.

\begin{figure}
    \centering
   \begin{subfigure}[b]{0.40\textwidth}
        \centering
        \includegraphics[width=\textwidth]{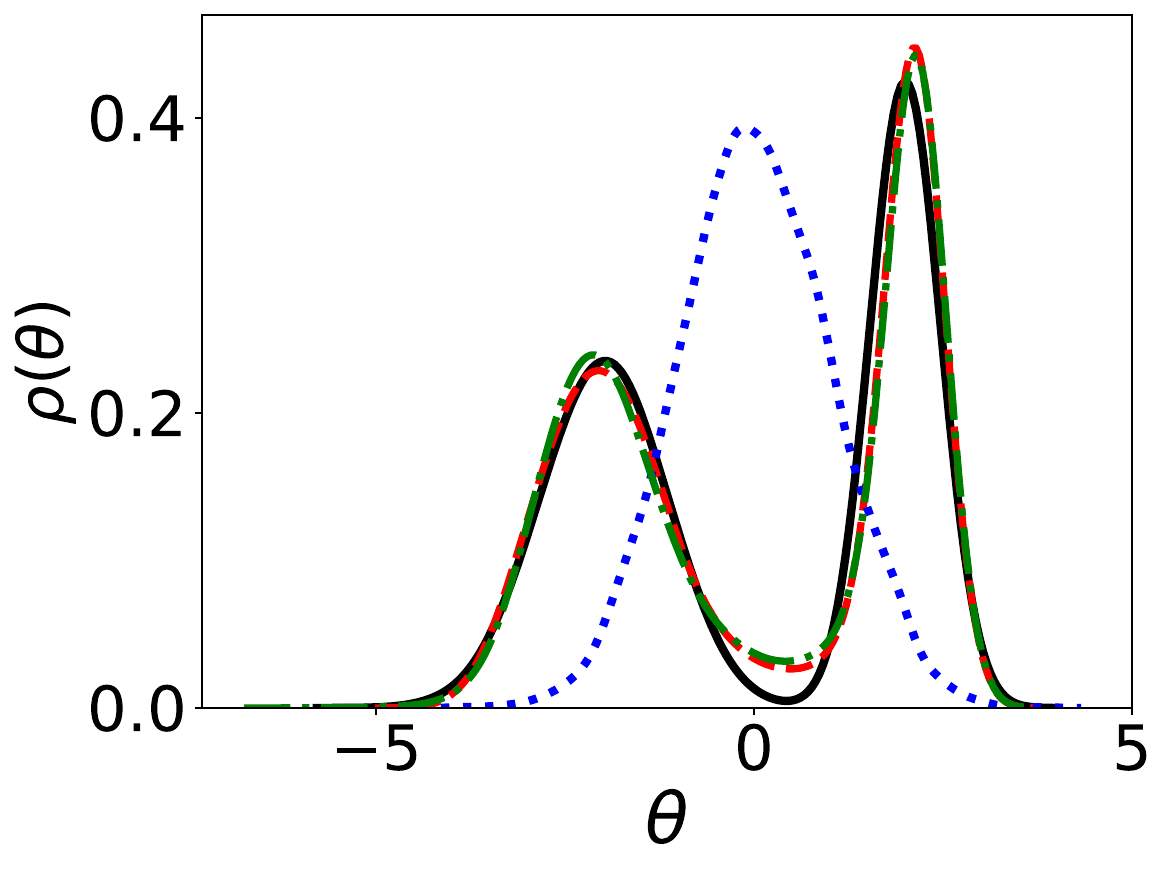} 
        \caption{Parameter distributions (initial, estimated, ground truth) using Energy distance and KL divergence..}
        \label{fig:struct_dist}
    \end{subfigure}
    \hfill
   \begin{subfigure}[b]{0.40\textwidth}
        \centering
        \includegraphics[width=\textwidth]{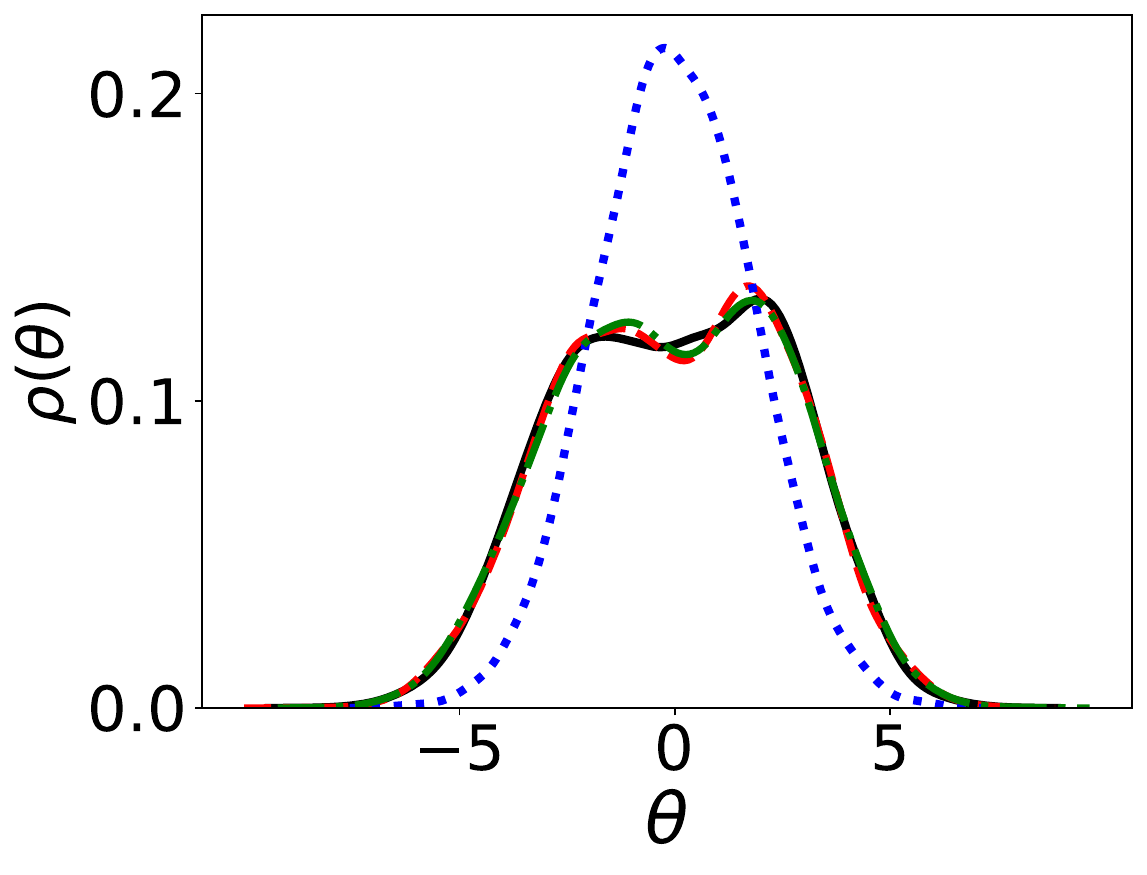} 
        \caption{Data distributions (initial, estimated, ground truth) using Energy distance and KL divergence.}
        \label{fig:data_dist}

    \end{subfigure}

   \begin{tikzpicture}

    \draw[line width=1pt, black] (0,0) -- (1.2,0);
    \node[anchor=west] at (1.4,0) {True distribution};

    \draw[line width=1pt, red, dashed] (4.5,0) -- (5.7,0);
    \node[anchor=west] at (5.9,0) {Final distribution KL};

    \draw[line width=1pt, blue, dotted] (0,-0.8) -- (1.2,-0.8);
    \node[anchor=west] at (1.4,-0.8) {Initial distribution};

    \draw[line width=1pt, dash pattern=on 4pt off 2pt on 1pt off 2pt, color={rgb,255:red,0; green,158; blue,115}] (4.5,-0.8) -- (5.7,-0.8);
    \node[anchor=west] at (5.9,-0.8) {Final distribution Energy};
\end{tikzpicture}

    \caption{One-dimensional Test System: Comparison among true, estimated, and initial distributions under two different loss functions: energy distance and KL divergence. Columns show parameter space (left) and data space (right). The legend indicates line styles for each distribution.}
    \label{fig:comparison_Kl_energy}
\end{figure}

Fig.~\ref{fig:comparison_Kl_energy} compares the reconstructions obtained with KL divergence and energy distance. Both objectives recover the ground-truth latent and data distributions accurately. Quantitatively, the final Wasserstein distance between the estimated and true latent distributions is $0.1911$ for energy distance and $0.1870$ for KL divergence, so KL gives a slightly smaller final error in this example.

\begin{figure}
    \centering
    \begin{subfigure}[b]{0.40\textwidth}
        \centering
        \includegraphics[width=1\textwidth]{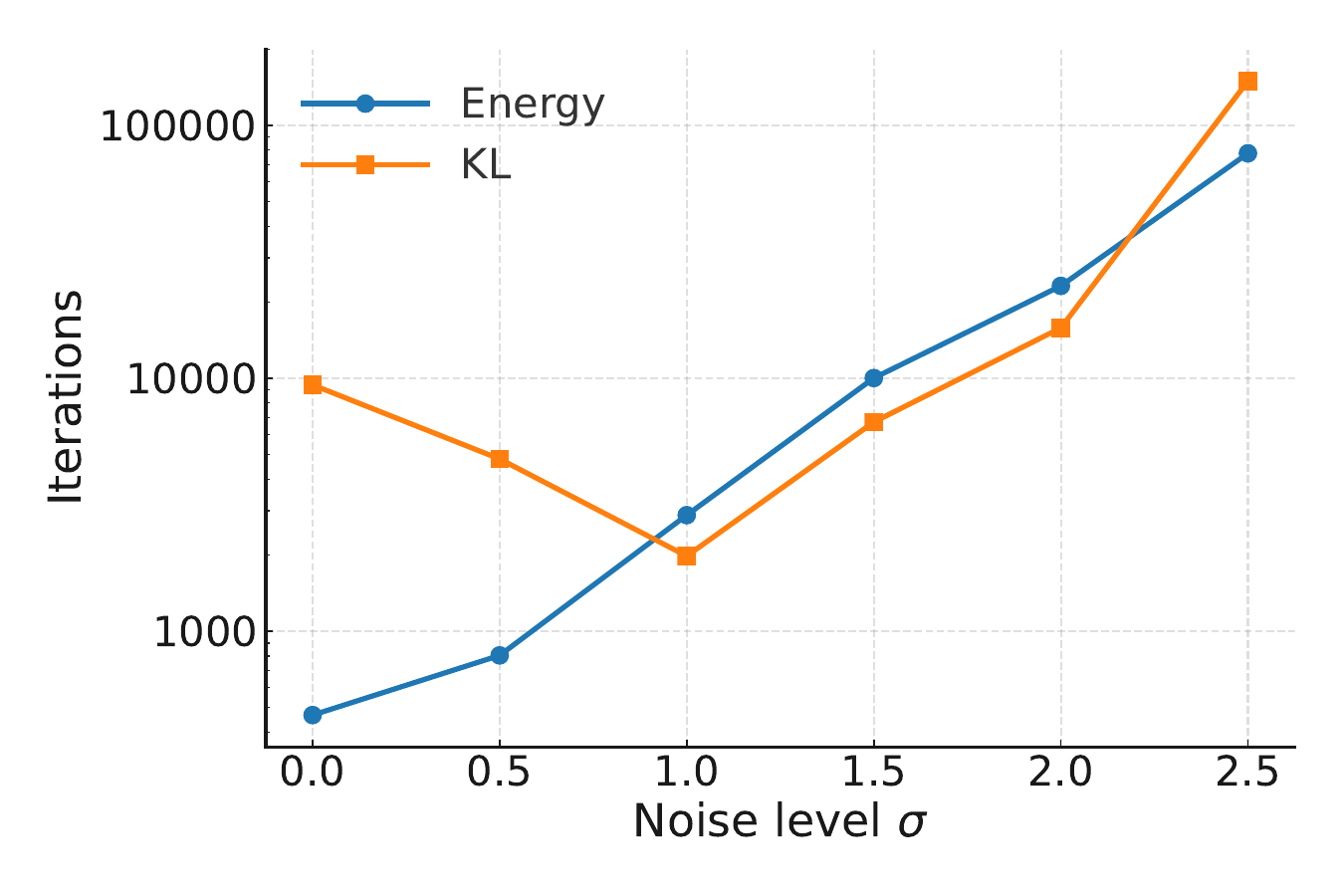} 
        \caption{Iterations needed to reach $W_2<0.2$.}
        \label{fig:Iterations}
    \end{subfigure}
    \begin{subfigure}[b]{0.40\textwidth}
        \centering
        \includegraphics[width=1\textwidth]{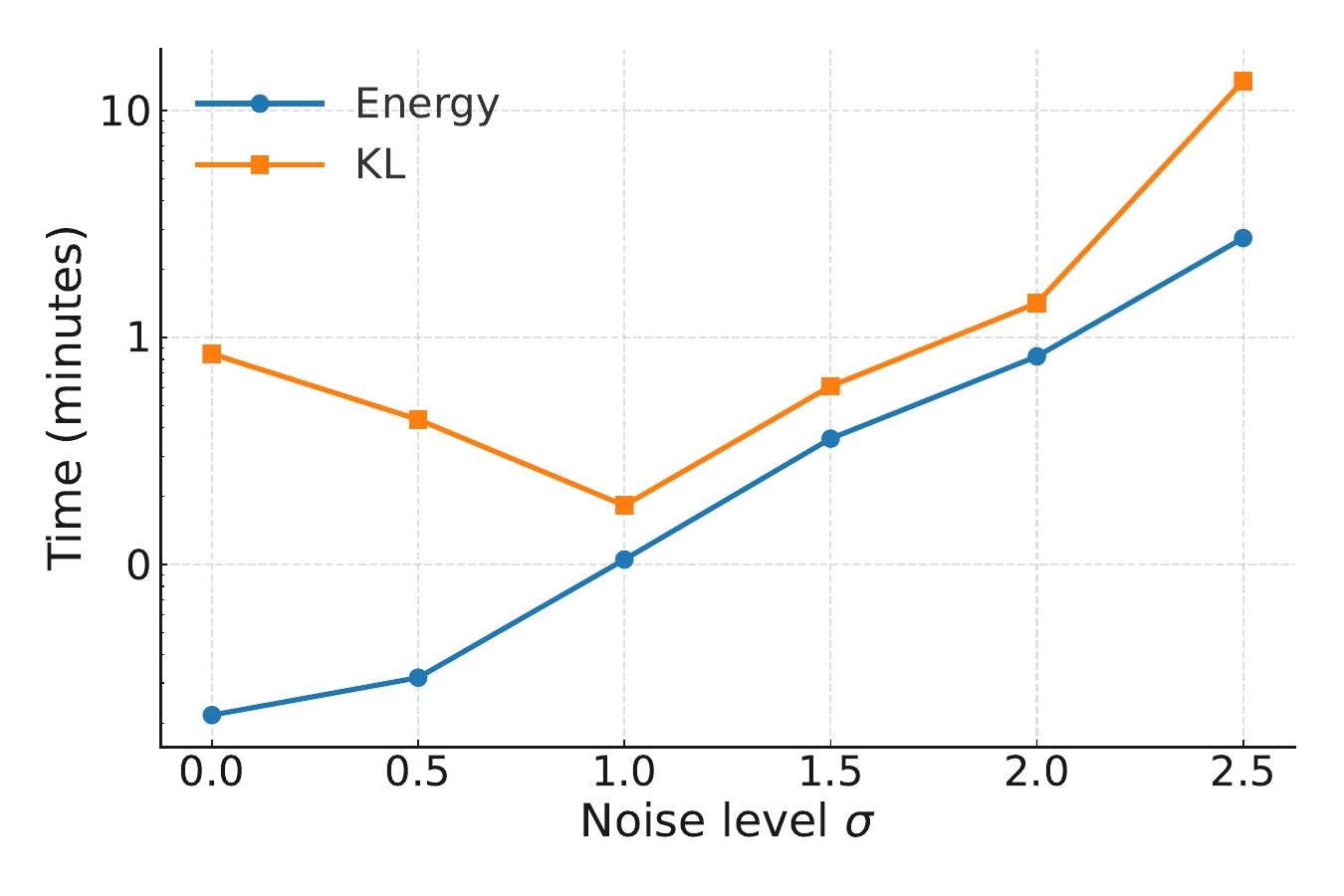} 
        \caption{Time needed to reach $W_2<0.2$.}
        \label{fig:Time}
    \end{subfigure}

    \caption{Comparison of the number of iterations (left) and computation time in minutes (right) required to reach a Wasserstein distance below 0.2 for different noise levels $\sigma$, using the Energy Distance (blue circles) and KL divergence (orange squares) as objective functions.}
    \label{fig:Time_analysis}
\end{figure}

Fig.~\ref{fig:Time_analysis} compares the computational effort required to reach $W_2<0.2$ across noise levels. Both objectives become more expensive as the noise level increases, but KL is more variable and generally slower, especially at higher noise levels. This is mainly because each KL iteration requires KDE-based density estimation, whereas energy distance can be evaluated directly from empirical samples. Although KL achieves a slightly smaller final Wasserstein error in Fig.~\ref{fig:comparison_Kl_energy}, energy distance is more stable with respect to noise and more computationally efficient in this experiment.

\subsection{Heterogeneity in Nanoclusters}

We next consider a two-dimensional latent example with high-dimensional image observations. The goal is to bridge the one-dimensional test problem and the protein experiments: the latent distribution remains fully visualizable, while the observations are images of size $128\times128$.

The latent variable $\theta=(w,h)\in\mathbb R^2$ parameterizes a four-particle nanocluster arranged at the vertices of a rectangle. Specifically, the particle locations are

\[
a_1=\frac{\theta}{2},\qquad
a_2=a_1\odot(1,-1),\qquad
a_3=a_1\odot(-1,1),\qquad
a_4=a_1\odot(-1,-1),
\]

where $\odot$ denotes componentwise multiplication. The deterministic forward map $T:\mathbb R^2\to\mathbb R^{128\times128}$ renders these four locations on a pixel grid by placing a Gaussian kernel at each particle location and summing the contributions. Details of the rendering procedure are given in Section~S1.7 of the \textbf{SM}.

The random forward model is
\[
T_\omega(\theta)=T(\theta)+\omega,
\qquad
\omega\sim\mathcal N(0,\sigma^2 I),
\]
with $\sigma=1.5$. The true latent distribution is the Gaussian mixture
\[
\theta \sim \alpha \, \mathcal{N} \left( \begin{bmatrix} 3 \\ 3 \end{bmatrix}, \begin{bmatrix} 0.5 & 0 \\ 0 & 0.5 \end{bmatrix} \right) + (1 - \alpha) \, \mathcal{N} \left( \begin{bmatrix} 5 \\ 5 \end{bmatrix}, \begin{bmatrix} 0.7 & 0.5 \\ 0.5 & 1 \end{bmatrix} \right),
\]
where $\alpha = 0.2 $. The initial guess $\theta_{0}\sim \mathcal{N}\left( \begin{bmatrix} 4 \\ 4 \end{bmatrix}, \begin{bmatrix} 0.8 & 0.3 \\ 0.3 & 0.8 \end{bmatrix} \right).$

We generate $n=1000$ noisy observations and run Algorithm~\ref{alg:alg} for $3000$ iterations. The objective is the energy distance~\eqref{eq:energyMMD}, with distributions represented by empirical measures supported on the particle locations. The particle-based gradient flow~\eqref{eq:W2_GF} is implemented using the scheme in~\eqref{eq:particle}.

\begin{figure}
    \centering
    \begin{subfigure}[b]{0.40\textwidth}
        \centering
        \includegraphics[width=\textwidth]{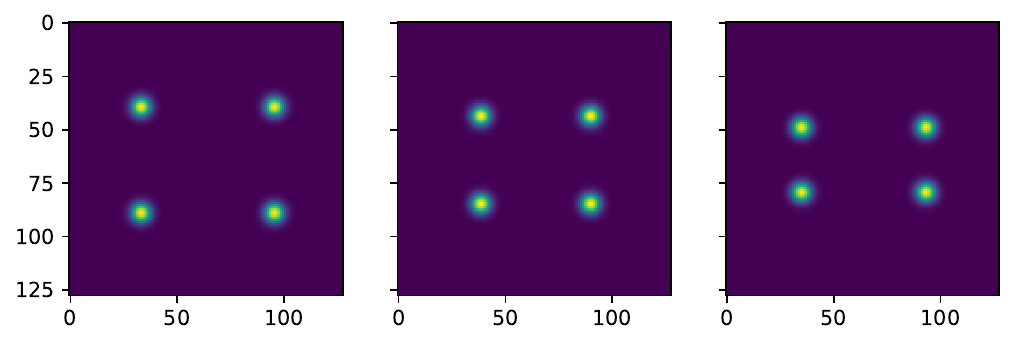} 
        \caption{True data (noiseless images).}
        \label{fig:symmetry_true_images}
    \end{subfigure}
    \begin{subfigure}[b]{0.40\textwidth}
        \centering
        \includegraphics[width=\textwidth]{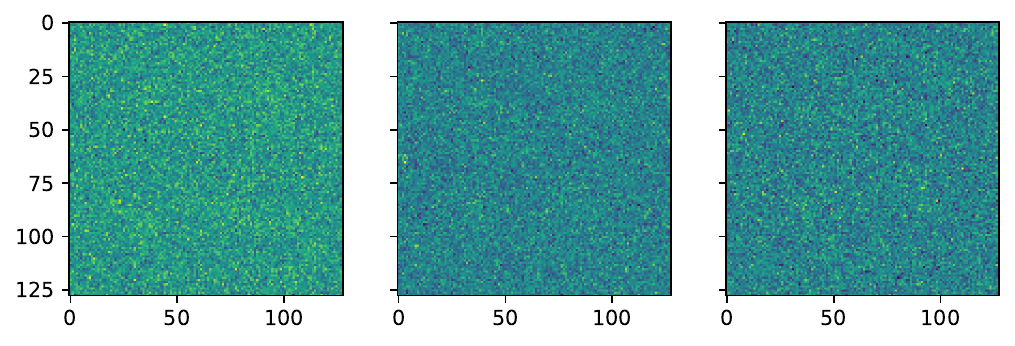} 
        \caption{Observed data with noise.}
        \label{fig:summetry_noisy_images}
    \end{subfigure}

    \caption{Heterogeneity in Nanoclusters: Visual comparison between the true images generated from the clean data (a) and the corresponding noisy observations (b).}
    \label{fig:image_noise_comparison}
\end{figure}

\begin{figure}
    \centering
    \includegraphics[width=0.9\textwidth]{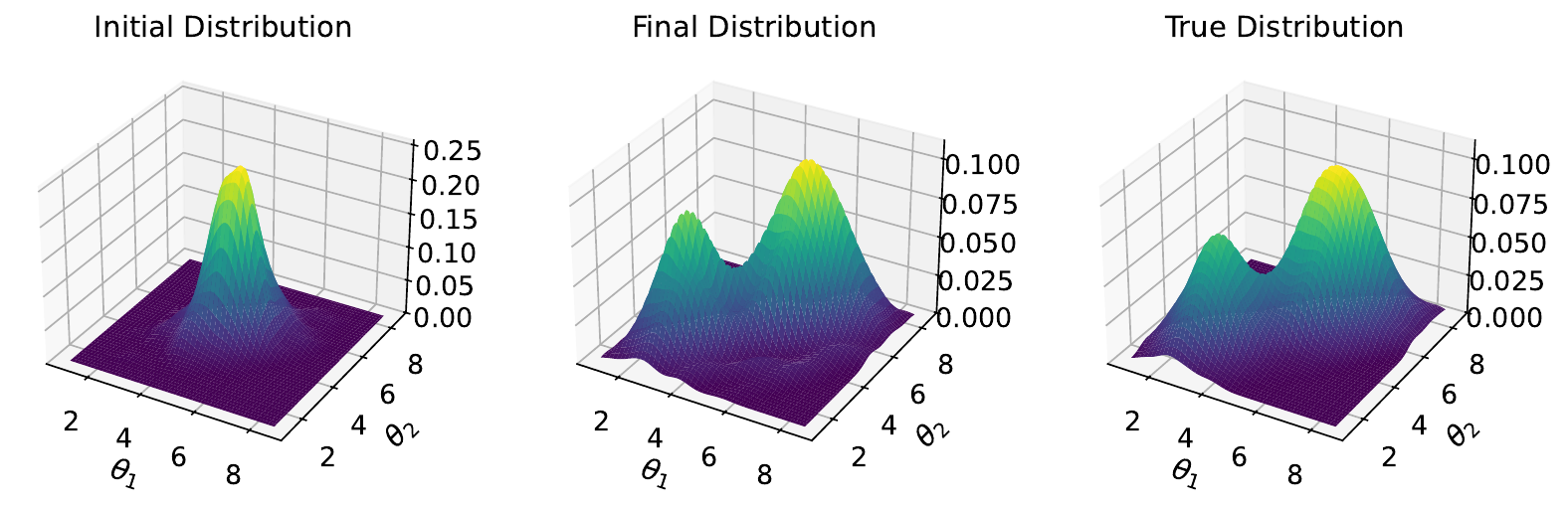}
    \caption{Heterogeneity in Nanoclusters: The initial parameter distribution (left), the estimated distribution (middle) after optimization and the ground truth distribution (right) are shown.}
    \label{fig:distribution_comparison}
\end{figure}

Fig.~\ref{fig:image_noise_comparison} shows examples of noiseless rendered images and the corresponding noisy observations. Fig.~\ref{fig:distribution_comparison} shows the initial, estimated, and ground-truth latent distributions. The recovered distribution improves substantially over the initial distribution: the energy distance in parameter space decreases from $0.429$ to $0.0163$, while the image-space discrepancy decreases from $0.0947$ to $0.0277$. This example demonstrates that the method can recover a low-dimensional latent distribution from high-dimensional image observations while retaining a fully visualizable ground truth.

\subsection{Simulated Protein Cryo-EM}\label{ssec:cryoem_example}

\begin{figure}
    \centering
    \raisebox{0.4\height}{%
        \begin{subfigure}[b]{0.35\textwidth}
            \centering
            \rotatebox{180}{%
                \includegraphics[width=0.8\textwidth]{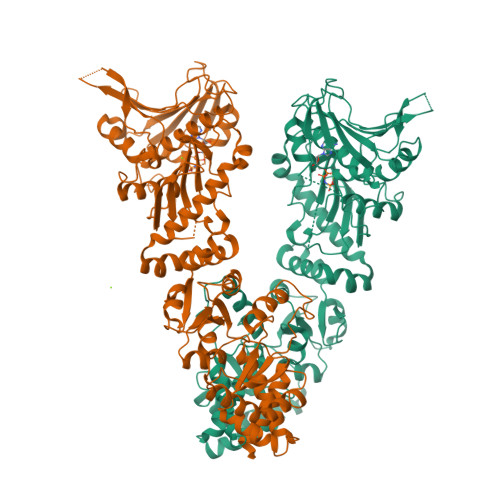}
            }
            \caption{Protein HSP90 structure.}
            \label{fig:hsp90_structure}
        \end{subfigure}
    }
    \hfill
    \begin{subfigure}[b]{0.60\textwidth}
        \centering
        \begin{subfigure}[b]{\textwidth}
            \centering
            \includegraphics[width=0.8\textwidth]{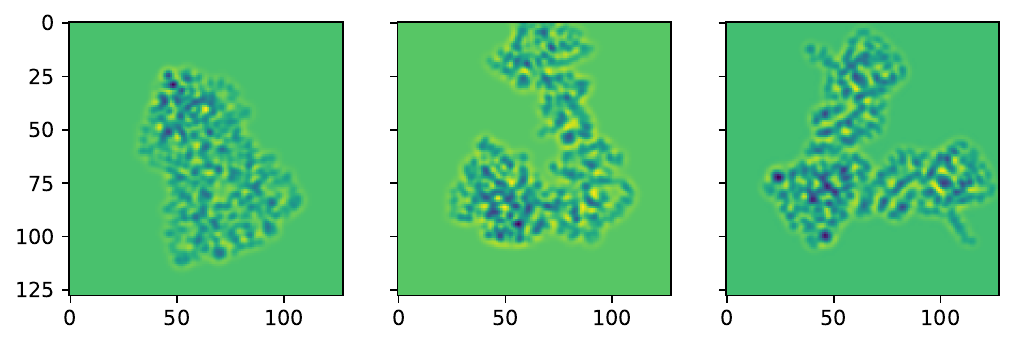}
            \caption{Simulated cryo-EM images without noise.}
            \label{fig:hsp90_clean}
        \end{subfigure}
        
        \vspace{0.5cm}
        
        \begin{subfigure}[b]{\textwidth}
            \centering
            \includegraphics[width=0.8\textwidth]{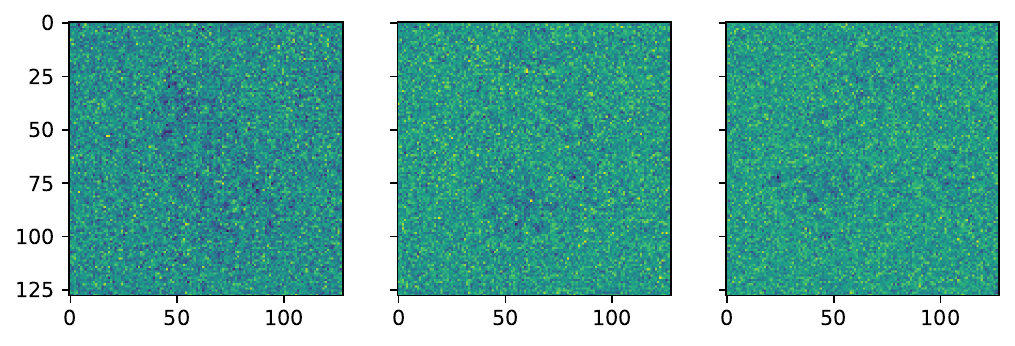}
            \caption{Simulated cryo-EM images with added noise.}
            \label{fig:hsp90_noisy}
        \end{subfigure}
    \end{subfigure}

    \caption{Illustrations of the 3D structure of protein HSP90 (a) and 2D simulated cryo-EM projections (b) and (c). In particular, (b) shows noiseless projections while (c) includes Gaussian noise to simulate experimental conditions.}
    \label{fig:hsp90_cryoem_simulation}
\end{figure}

The final example aims to emulate the recovery of conformational heterogeneity on a physically motivated protein-motion example. Using a standard model of cryo-EM image formation, we simulate cryo-EM-like images of the HSP90 protein undergoing a conformational change, taken from random rotations.  We then attempt to recover this conformational change from the observed images.

We represent protein conformational variability using a low-dimensional normal-mode basis. Normal modes capture the collective, low-frequency vibrations of the protein and are derived by solving the eigenvalue problem of the Hessian matrix associated with the potential energy landscape. The resulting eigenvectors represent directions of motion, while the eigenvalues quantify the stiffness or frequency of these motions. Biologically relevant conformational changes—such as hinge bending or domain opening and closing—are typically encoded in the low-frequency modes associated with the smallest eigenvalues~\cite{PhysRevLett.77.1905, PhysRevLett.79.3090,Bahar2010}. In this example, we focus on the four principal modes of the protein HSP90, with the initial structure adapted from Protein Data Bank entry 2IOP~\cite{shiau2006structural}.
These four modes, calculated using the \texttt{ProDy} python package~\cite{zhang2021prody}, give a compact yet physically meaningful basis for the optimization process. 

The selected normal modes define a finite-dimensional latent coordinate system:
a conformation is parameterized by coefficients $\theta=(\theta_1,\ldots,\theta_4)$ along these modes, and let $G(\theta)$ be the corresponding three-dimensional electrostatic potential.

\revtwo{Combining the cryo-EM image formation model~\eqref{eq:cryo-EM-eqn} with a low-dimensional representation of protein motion via structural modes, the images generated are given by:
\[
    y=T(\theta,\omega) = H^{\omega_r} G(\theta) + \omega_n\,,
\]

where $\omega_r$ is a random viewing direction and $\omega_n$ is additive Gaussian image noise. The protein-to-image map is implemented using \texttt{cryoJAX}~\cite{cryojax_library}. The image formation model includes the CTF through a weak-phase scattering model with image size $128\times128$, pixel size $1.5$~\AA{}, accelerating voltage $300$~kV, defocus $1000$~\AA{}, and astigmatism $0$~\AA{}.

We choose a synthetic target distribution so that the ground truth conformational distribution is known. The first two normal-mode coefficients are sampled from a bimodal Gaussian mixture,

\[
\theta_i=\frac{z_i}{\sqrt{\lambda_i}},
\qquad
z_i\sim
\frac12\mathcal N(9,1)
+
\frac12\mathcal N(-7,1),
\qquad
i=1,2,
\]
while the remaining coefficients are sampled as
\[
\theta_i=\frac{z_i}{\sqrt{\lambda_i}},
\qquad
z_i\sim\mathcal N(0,1),
\qquad
i=3,4.
\]
The initial particles are sampled from
\[
\theta_{0i}=\frac{v_i}{\sqrt{\lambda_i}},
\qquad
v_i\sim\mathcal U(-7,7),
\qquad
i=1,\ldots,4.
\]
The Gaussian-mixture form is used only to generate synthetic data. The reconstruction method itself does not assume a Gaussian mixture structure or independence across normal-mode coefficients; it represents the unknown conformational distribution nonparametrically by particles.}

We generate $3000$ noisy cryo-EM images using random rotations drawn uniformly from $SO(3)$ and optimize the energy distance between observed and simulated image distributions. Figs.~\ref{fig:hsp90_structure},~\ref{fig:hsp90_clean}, and~\ref{fig:hsp90_noisy} respectively display the 3D molecular structure, the noiseless projections generated from atomic positions, and the noisy observations used as input to the optimization framework.

\begin{figure}
    \centering
    \begin{subfigure}[b]{0.24\textwidth}
        \centering
        \includegraphics[width=\textwidth]{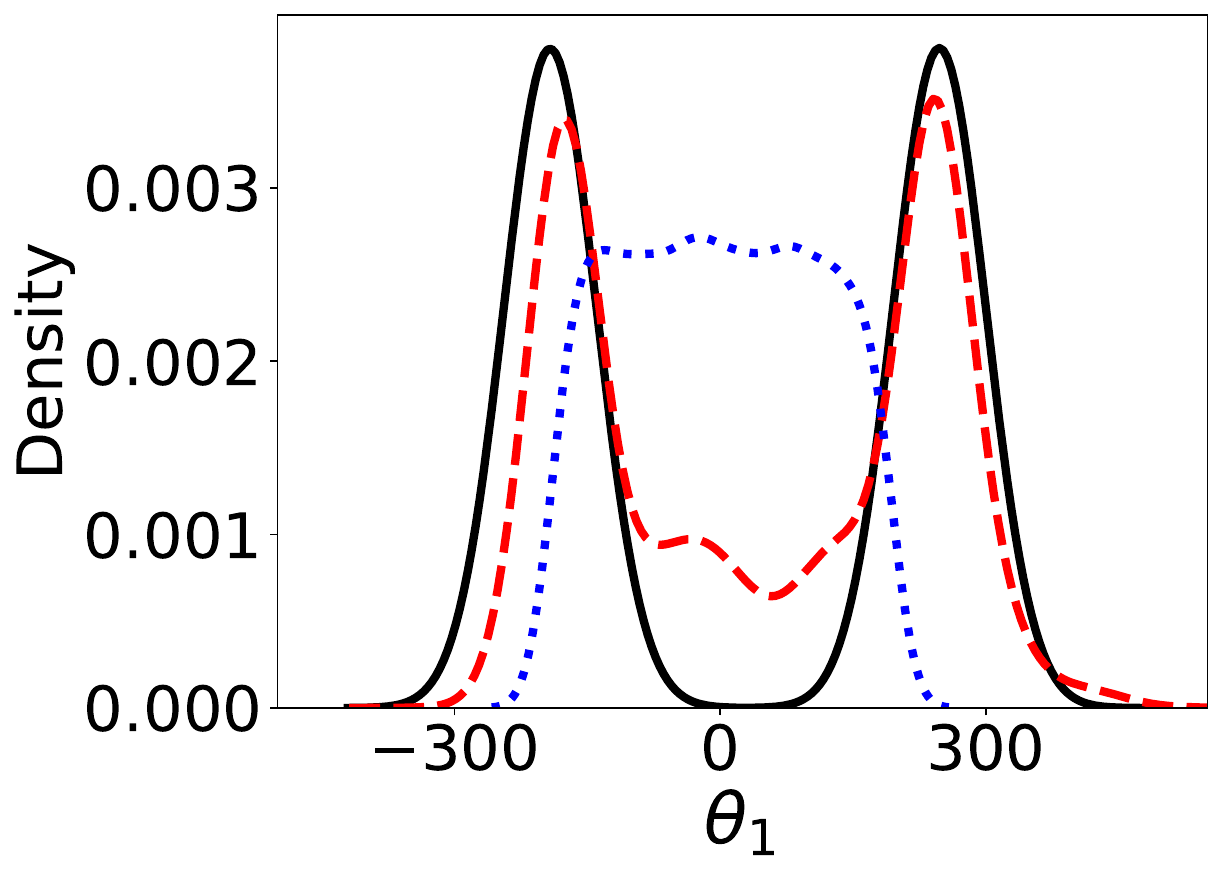} 
        \caption{Mode 1}
        \label{fig:mode1_dist}
    \end{subfigure}
    \begin{subfigure}[b]{0.24\textwidth}
        \centering
        \includegraphics[width=\textwidth]{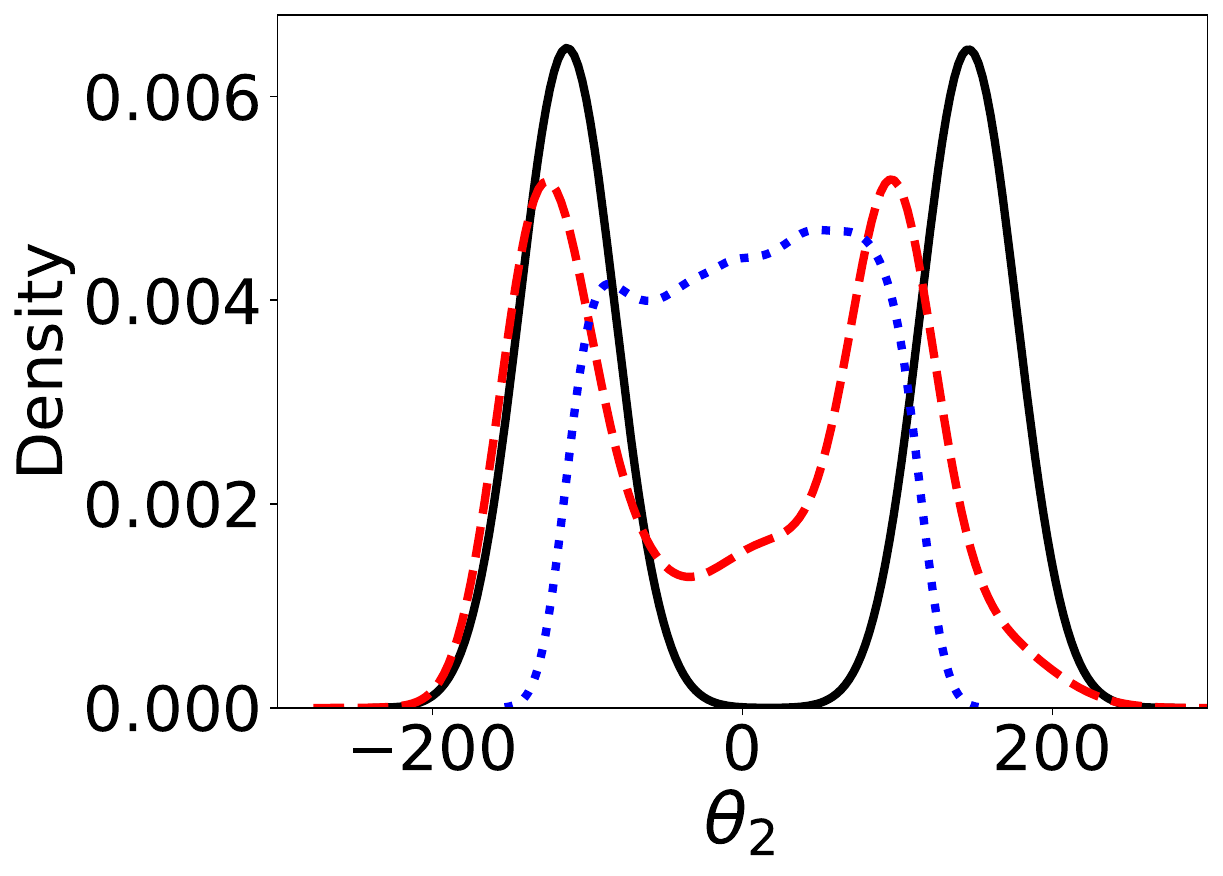} 
        \caption{Mode 2}
        \label{fig:mode2_dist}
    \end{subfigure}
    \begin{subfigure}[b]{0.24\textwidth}
        \centering
        \includegraphics[width=\textwidth]{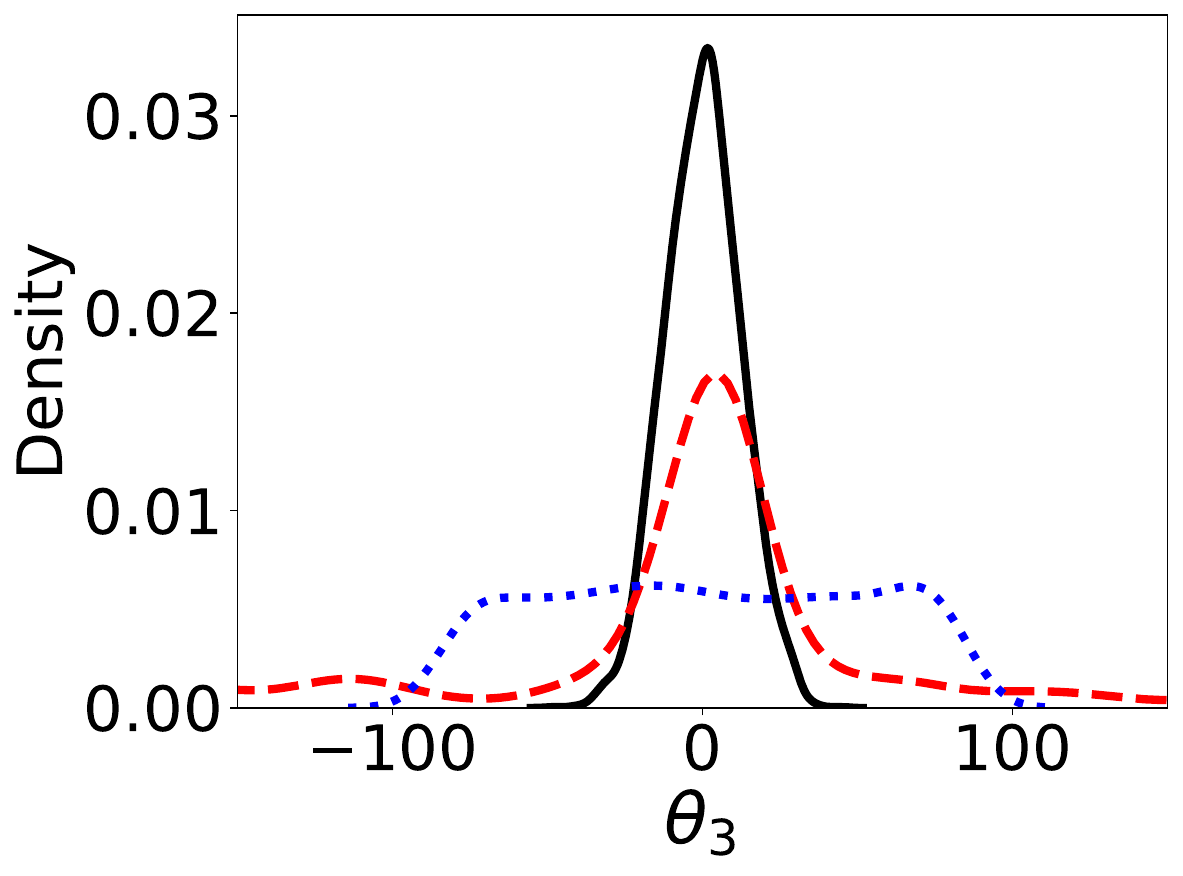} 
        \caption{Mode 3}
        \label{fig:mode3_dist}
    \end{subfigure}
    \begin{subfigure}[b]{0.24\textwidth}
        \centering
        \includegraphics[width=\textwidth]{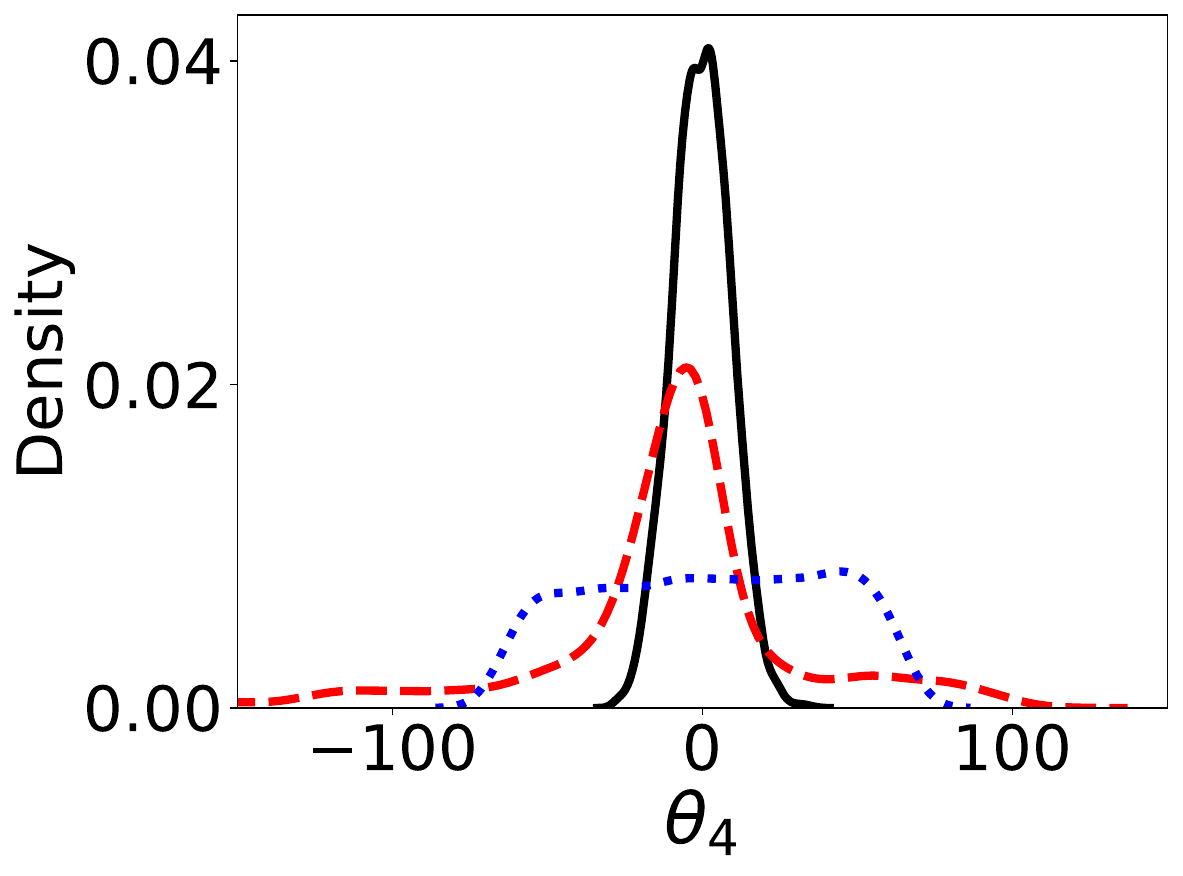} 
        \caption{Mode 4}
        \label{fig:mode4_dist}
    \end{subfigure}

    \vspace{1em}

    \begin{tikzpicture}
        \draw[line width=1pt, black] (0,0) -- (1.2,0);
        \node[anchor=west] at (1.4,0) {True distribution};

        \draw[line width=1pt, red, dashed] (4.5,0) -- (5.7,0);
        \node[anchor=west] at (5.9,0) {Final distribution};

        \draw[line width=1pt, blue, dotted] (9.0,0) -- (10.2,0);
        \node[anchor=west] at (10.4,0) {Initial distribution};
    \end{tikzpicture}

    \caption{Cryo-EM synthetic example: Comparison between the true parameter distribution (black), the estimated distribution (red), and the initial guess (blue). From left to right: Mode 1 to Mode 4.}
    \label{fig:modes}
\end{figure}

\begin{figure}
  \centering

  \begin{minipage}[b]{0.7\textwidth}
    \includegraphics[width=\textwidth]{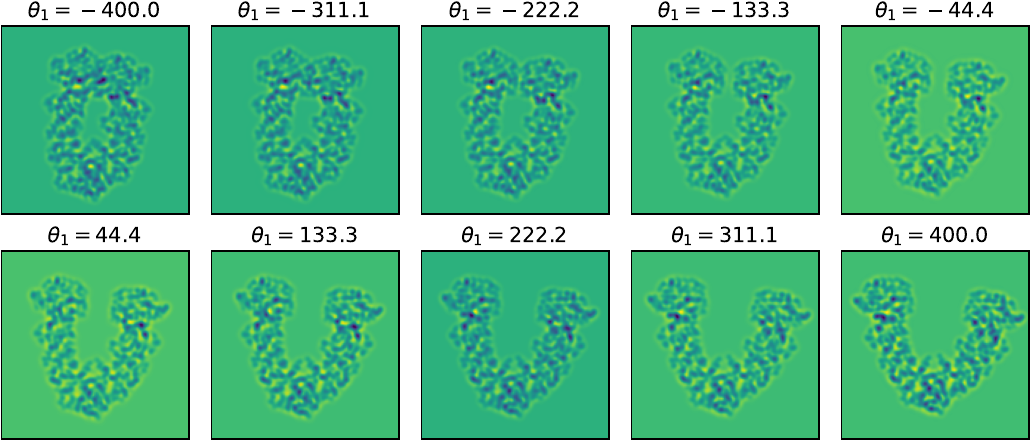}
  \end{minipage}

  \caption{
    Visualization of conformational variability along the first mode.
  }
  \label{fig:mode_move}
\end{figure}

Figure~\ref{fig:modes} compares the marginal distributions of the true, initial, and recovered normal-mode coefficients. The first two modes are recovered most accurately, while the third and fourth modes show larger deviations. To illustrate the structural meaning of these latent variables, Figure~\ref{fig:mode_move} shows representative conformations along the first normal-mode direction; visualizations for the remaining modes are provided in Supplementary Figure~S1. The mode-dependent recovery is consistent with these structural visualizations: leading modes induce larger and more visible conformational changes, whereas later modes produce subtler deformations that are more weakly observed after projection and noise.

The improvement is also visible quantitatively. The latent energy distance decreases from $89.21$ for the initial distribution to $19.44$ after optimization. In image space, the discrepancy to the observed distribution decreases from $1.13$ to $0.21$. Thus, the method improves both the simulated image distribution and the recovered latent conformational distribution, although recovery accuracy varies across normal modes. This is consistent with the stability discussion in Section~\ref{sec:stability}, where weakly observable structural directions lead to poorer latent recovery even when the image-space discrepancy is reduced.

\begin{figure}
    \centering
    
    \begin{subfigure}[b]{0.40\textwidth}
        \centering
        \includegraphics[width=\textwidth]{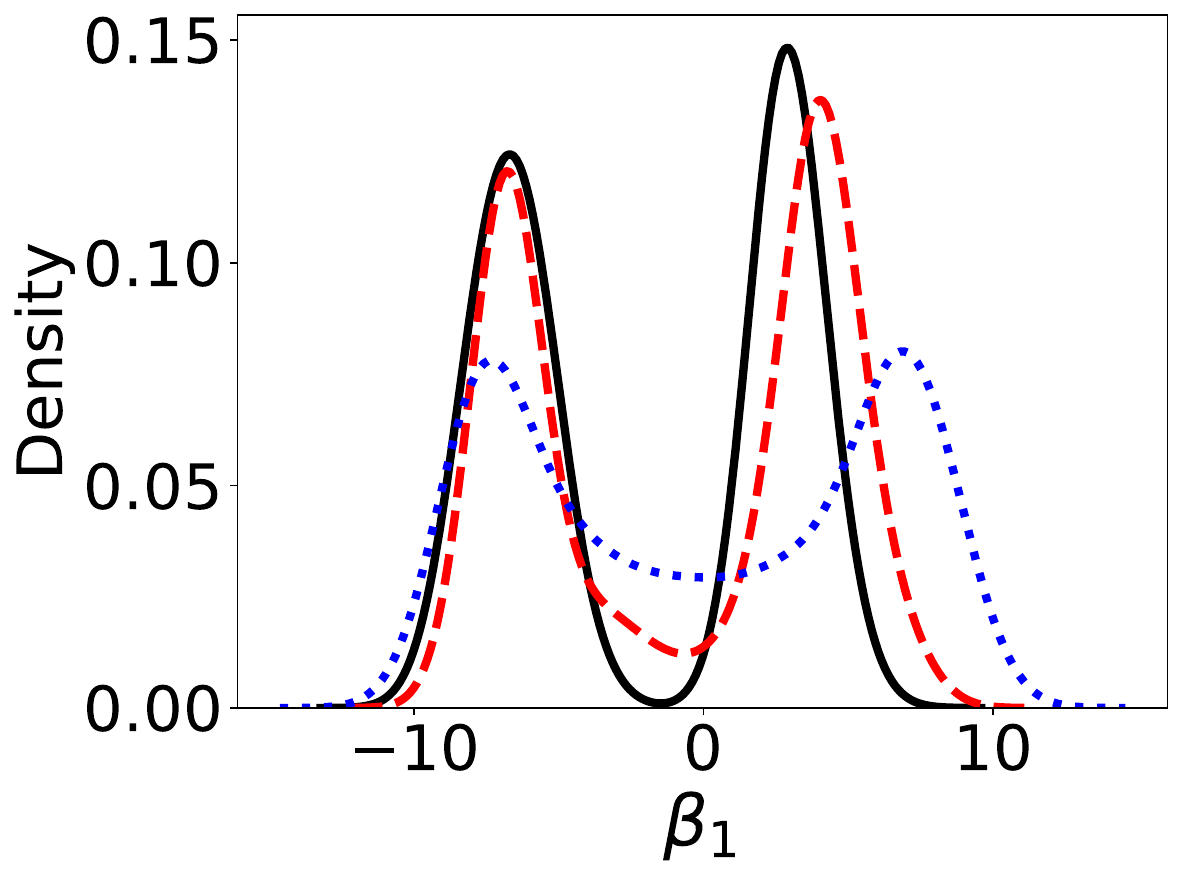} 
        \caption{First Principal Component}
        \label{fig:pca1_noisy}
    \end{subfigure}
    \begin{subfigure}[b]{0.40\textwidth}
        \centering
        \includegraphics[width=\textwidth]{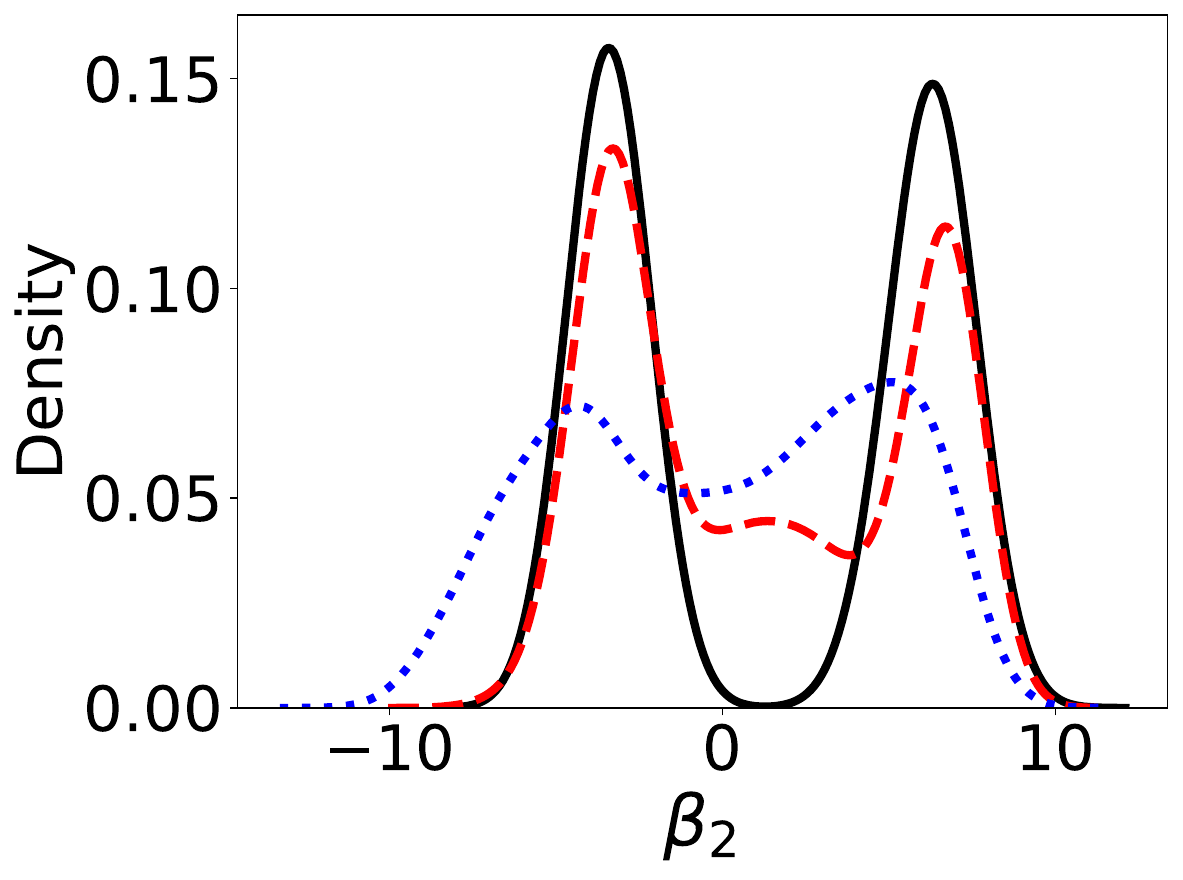} 
        \caption{Second Principal Component}
        \label{fig:pca2_noisy}
    \end{subfigure}

    \vspace{1em}

    \begin{tikzpicture}
        \draw[line width=1pt, black] (0,0) -- (1.2,0);
        \node[anchor=west] at (1.4,0) {True distribution};

        \draw[line width=1pt, red, dashed] (4.5,0) -- (5.7,0);
        \node[anchor=west] at (5.9,0) {Final distribution};

        \draw[line width=1pt, blue, dotted] (9.0,0) -- (10.2,0);
        \node[anchor=west] at (10.4,0) {Initial distribution};
    \end{tikzpicture}

    \caption{Cryo-EM Synthetic Example: Comparison among PCA for the true PCA distribution (black), initial PCA distribution (blue) and final PCA distribution (red) of noisy images.}
    \label{fig:pca}
\end{figure}

Finally, we compare generated and observed images by projecting them onto the first two principal components of the image dataset. Figure~\ref{fig:pca} shows that the recovered image distribution aligns well with the target distribution in the dominant PCA directions. This provides an image-space diagnostic consistent with the decrease in energy distance and confirms that the particle optimization improves the distribution of simulated cryo-EM images, not only the latent marginals.

\subsection{Comparison with Gaussian-mixture baselines} \label{sec:gmm_baseline}
\revone{To demonstrate the utility of a particle-based model for conformational heterogeneity, we compare our method with Gaussian-mixture baselines in a one-dimensional protein normal-mode experiment. We first consider a setting intentionally favorable to the Gaussian-mixture model, where the true latent distribution is itself a two-component Gaussian mixture and the correct number of components is included among the tested choices.}

\begin{figure}
    \centering

    \begin{subfigure}[t]{0.38\textwidth}
        \centering
        \includegraphics[width=\textwidth]{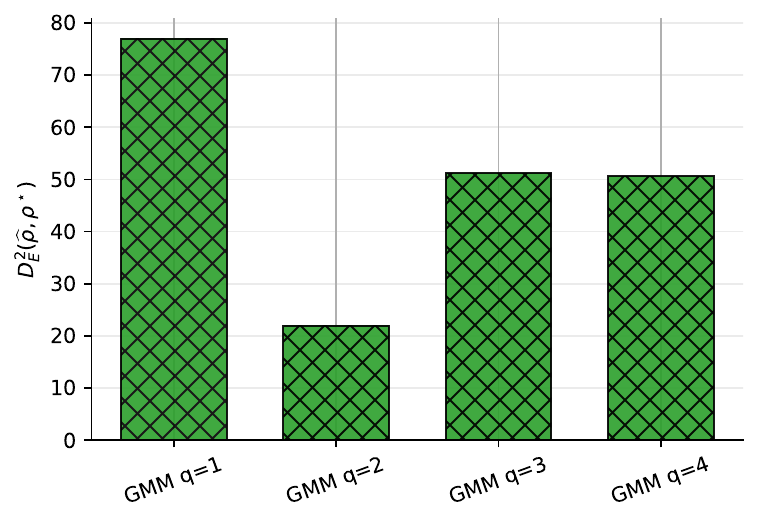}
        \caption{
        Latent-space energy distance for Gaussian-mixture baselines with
        $q=1,2,3,4$ components.
        }
        \label{fig:gmm_components_bar}
    \end{subfigure}
    \hfill
    \begin{subfigure}[t]{0.45\textwidth}
        \centering
        \includegraphics[width=\textwidth]{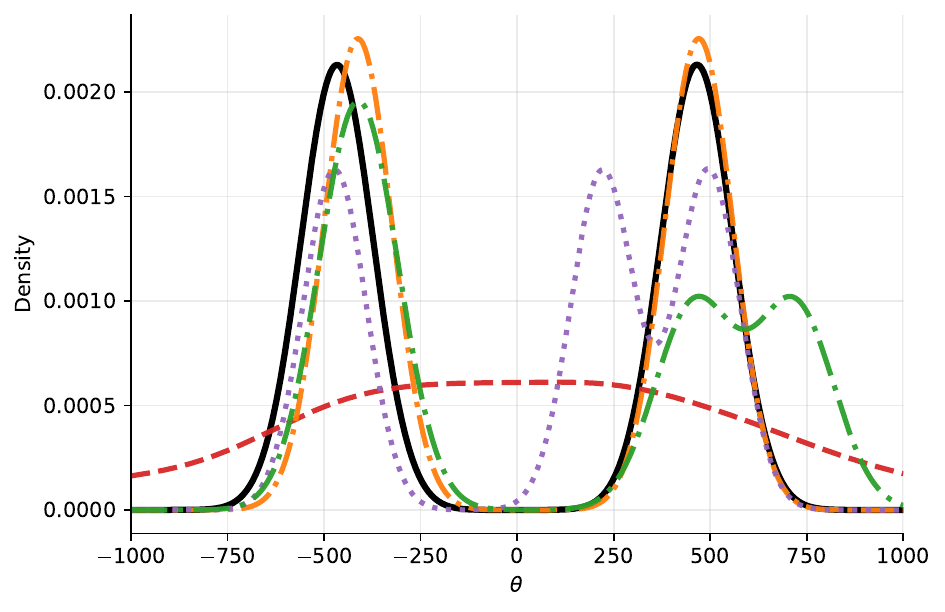}
        \caption{
        True latent density and recovered Gaussian-mixture densities for
        different choices of $q$.
        }
        \label{fig:gmm_components_density}
    \end{subfigure}

    \vspace{0.8em}

    \begin{tikzpicture}
        \draw[line width=1pt, black] (0,0) -- (1.0,0);
        \node[anchor=west] at (1.15,0) {True distribution};

        \draw[line width=1pt, red, dashed] (4.0,0) -- (5.0,0);
        \node[anchor=west] at (5.15,0) {GMM, $q=1$};

        \draw[line width=1pt, orange, dash dot] (7.2,0) -- (8.2,0);
        \node[anchor=west] at (8.35,0) {GMM, $q=2$};

        \draw[line width=1pt, purple, dotted] (0,-0.45) -- (1.0,-0.45);
        \node[anchor=west] at (1.15,-0.45) {GMM, $q=3$};

        \draw[line width=1pt, ForestGreen, dash pattern=on 7pt off 2pt on 1pt off 2pt] (4.0,-0.45) -- (5.0,-0.45);
        \node[anchor=west] at (5.15,-0.45) {GMM, $q=4$};
    \end{tikzpicture}

    \caption{
    Gaussian-mixture baseline in the one-dimensional protein normal-mode
    experiment with a two-Gaussian target distribution. 
    The model is misspecified for $q \neq 2$.
    }
    \label{fig:gmm_components}
\end{figure}

\revone{Figure~\ref{fig:gmm_components} compares Gaussian-mixture baselines with $q=1,2,3,4$ components. As expected, the correctly specified $q=2$ model achieves the smallest latent-space energy distance. The one-component model underfits the bimodal target, while using more than two components does not improve the result and can introduce spurious splitting of the modes. At the density level, the $q=2$ model accurately captures the two dominant modes, whereas the misspecified choices either oversmooth the distribution or introduce artificial structure. This confirms that Gaussian mixtures can be highly effective when the true distribution belongs to the assumed parametric family and the number of components is chosen correctly.
}

\begin{figure}
    \centering

    \begin{subfigure}[t]{0.40\textwidth}
        \centering
        \includegraphics[width=\textwidth]{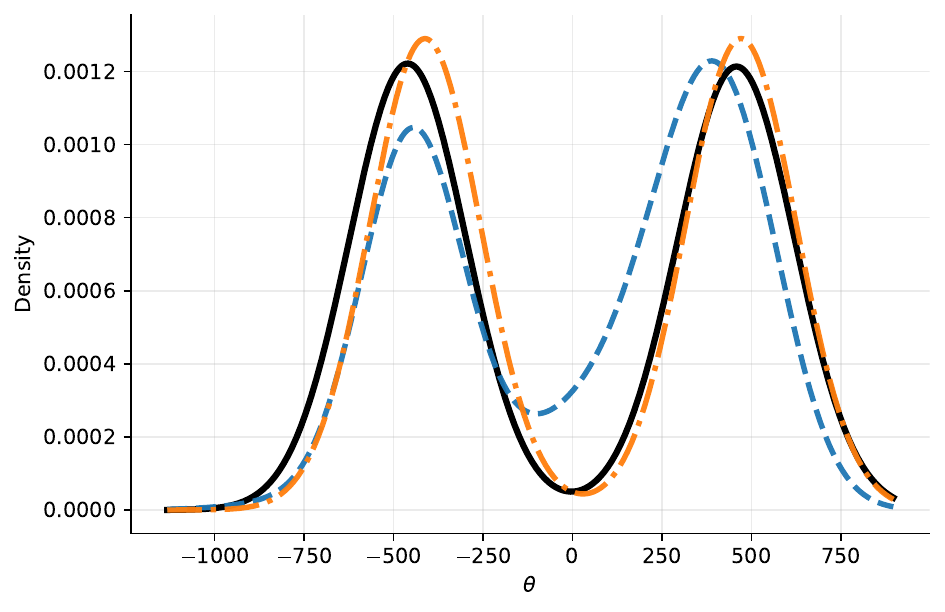}
        \caption{
        Latent density comparison.
        }
        \label{fig:wgf_vs_gmm_density_panel}
    \end{subfigure}
    \hfill
    \begin{subfigure}[t]{0.40\textwidth}
        \centering
        \includegraphics[width=\textwidth]{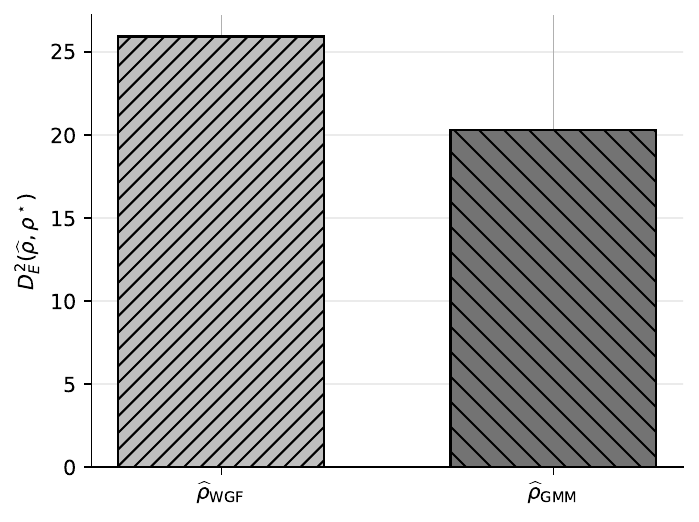}
        \caption{
        Latent-space energy distance.
        }
        \label{fig:wgf_vs_gmm_energy_panel}
    \end{subfigure}

    \vspace{0.8em}

    \begin{tikzpicture}
        \draw[line width=1pt, black] (0,0) -- (1.0,0);
        \node[anchor=west] at (1.15,0) {True distribution};

        \draw[line width=1pt, blue, dashed] (4.1,0) -- (5.1,0);
        \node[anchor=west] at (5.25,0) {WGF particles};

        \draw[line width=1pt, orange, dash dot] (7.7,0) -- (8.7,0);
        \node[anchor=west] at (8.85,0) {GMM, $q=2$};
    \end{tikzpicture}

    \caption{
    Comparison between the Wasserstein-gradient-flow particle method and the
    correctly specified $q=2$ Gaussian-mixture baseline in the one-dimensional
    protein normal-mode experiment with a two-Gaussian target distribution.
    }
    \label{fig:wgf_vs_gmm_comparison}
\end{figure}

\begin{figure}
    \centering

    \begin{subfigure}[t]{0.53\textwidth}
        \centering
        \includegraphics[width=\textwidth]{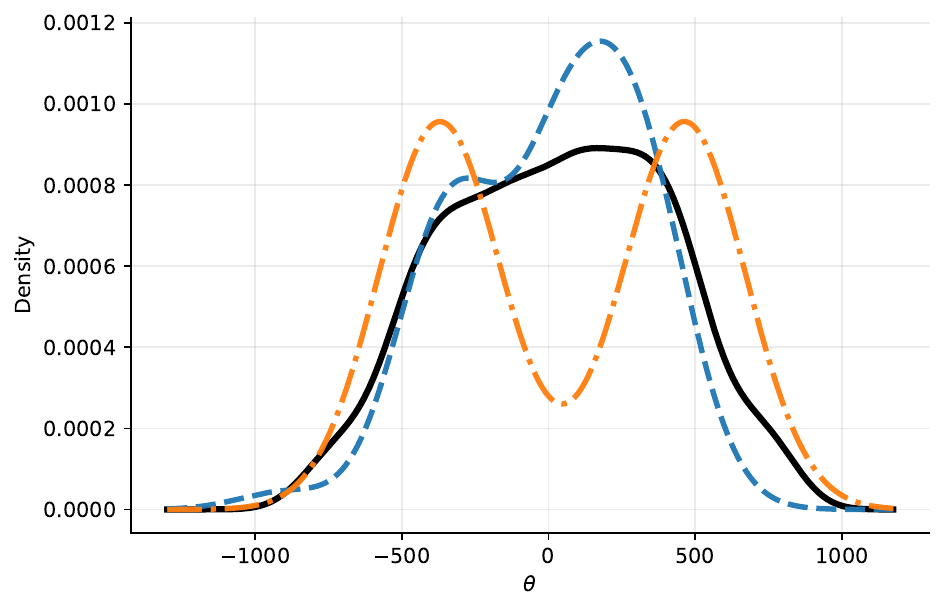}
        \caption{
        Latent density comparison.
        }
        \label{fig:uniform_wgf_gmm_density_panel}
    \end{subfigure}
    \hfill
    \begin{subfigure}[t]{0.39\textwidth}
        \centering
        \includegraphics[width=\textwidth]{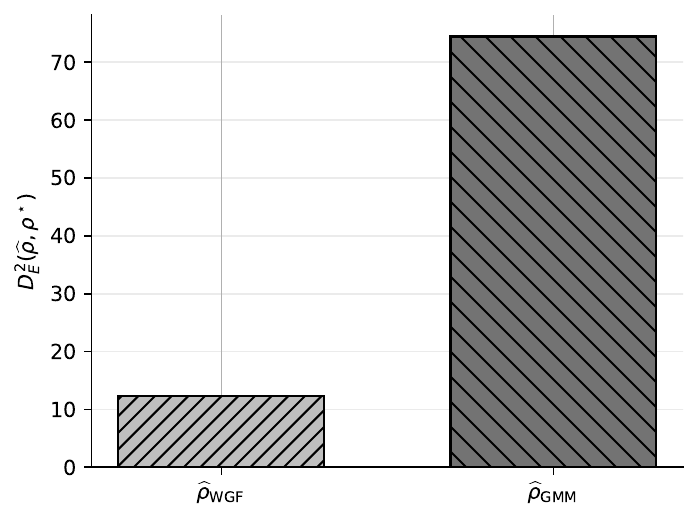}
        \caption{
        Latent-space energy distance.
        }
        \label{fig:uniform_wgf_gmm_energy_panel}
    \end{subfigure}

    \vspace{0.8em}

    \begin{tikzpicture}
        \draw[line width=1pt, black] (0,0) -- (1.0,0);
        \node[anchor=west] at (1.15,0) {True distribution};

        \draw[line width=1pt, blue, dashed] (4.1,0) -- (5.1,0);
        \node[anchor=west] at (5.25,0) {WGF particles};

        \draw[line width=1pt, orange, dash dot] (7.7,0) -- (8.7,0);
        \node[anchor=west] at (8.85,0) {GMM, $q=2$};
    \end{tikzpicture}

    \caption{
    Comparison between the Wasserstein-gradient-flow particle method and a two-component Gaussian-mixture baseline in the one-dimensional protein
    normal-mode experiment with an approximately uniform target distribution.  Here, the GMM baseline is misspecified for all values of $q$.
    }
    \label{fig:uniform_wgf_gmm_comparison}
\end{figure}

\revone{
We then compare the proposed particle method with the correctly specified $q=2$ Gaussian-mixture baseline. This is a strong baseline because the true latent distribution matches the assumed parametric family. As shown in Figures~\ref{fig:wgf_vs_gmm_density_panel} and~\ref{fig:wgf_vs_gmm_energy_panel}, the GMM baseline achieves a smaller latent-space energy distance, as expected in this well-specified setting. The particle method nevertheless recovers the qualitative bimodal structure without being given the number of modes or a Gaussian-mixture parametrization. This illustrates the main tradeoff: the WGF particle method is more flexible and less tied to a prescribed parametric family, but in this experiment it requires more iterations and longer runtime than the correctly specified GMM baseline. 

We next repeat the comparison in a deliberately misspecified setting for the Gaussian-mixture baseline. Here the ground-truth latent distribution is approximately uniform over an interval of the one-dimensional normal-mode coordinate, while the baseline is still restricted to a two-component Gaussian mixture. Thus the GMM has the correct latent dimension but the wrong distributional family.

Figure~\ref{fig:uniform_wgf_gmm_comparison} shows the resulting latent densities and latent-space energy distances. The Gaussian-mixture baseline concentrates mass near two separated modes and therefore fails to represent the flat target distribution. In contrast, the WGF particle method produces a substantially closer approximation of the uniform latent law. Importantly, the WGF particles are initialized from a Gaussian distribution rather than a uniform distribution, so the recovered flat profile is not inherited from the initialization but arises from matching the observed image distribution.

Together, the two comparisons illustrate the intended tradeoff. Parametric models can be highly efficient when the correct family is known, as in the two-Gaussian experiment. However, they can suffer significant bias under model misspecification, as in the approximately uniform example. The WGF particle method is less specialized and computationally more expensive, but it can adapt to latent distributions that are not well represented by a fixed Gaussian-mixture family.
}

\subsection{Effect of the number of particles and noise}

\begin{figure}
    \centering

    \begin{subfigure}[t]{0.43\textwidth}
        \centering
        \includegraphics[width=\textwidth]{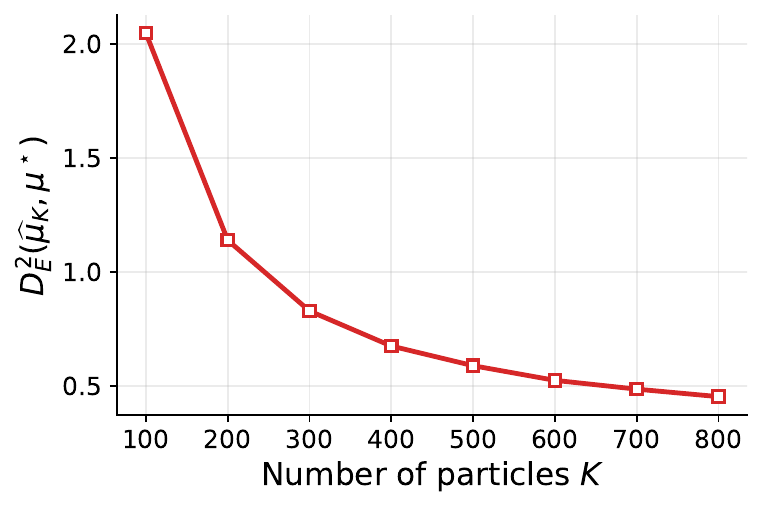}
        \caption{
        Image-space energy distance.
        }
        \label{fig:image_loss_vs_K_panel}
    \end{subfigure}
    \hfill
    \begin{subfigure}[t]{0.43\textwidth}
        \centering
        \includegraphics[width=\textwidth]{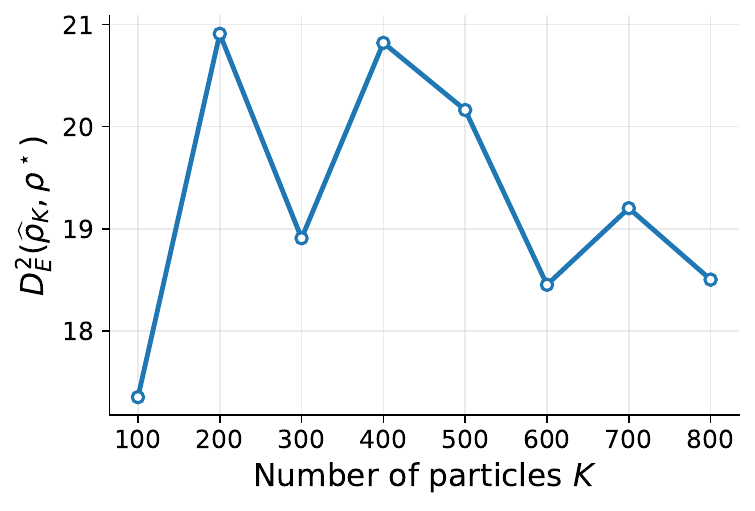}
        \caption{
        Latent-space energy distance.
        }
        \label{fig:latent_loss_vs_K_panel}
    \end{subfigure}

    \caption{
    Effect of the number of WGF particles $K$ in the one-mode protein experiment. The target image distribution is generated using $800$ images.  We compare recovery of both the image-space and structure-space distributions using the energy distance.
    }
    \label{fig:loss_vs_K}
\end{figure}

\revone{
We finally study two factors highlighted by the oracle bound in Section~\ref{sec:oracle-recovery-bound}: the number of particles $K$ and the image noise level. These experiments are performed in the one-mode protein setting.

Figure~\ref{fig:loss_vs_K} shows the dependence on $K$. As the number of particles increases, the image-space energy distance decreases, indicating that larger particle ensembles better approximate the predicted image distribution. The latent-space error, however, is nonmonotone. This is consistent with the theory: the objective is optimized in image space, and the cryo-EM forward map is not injective. Therefore, improved image-space matching does not necessarily imply monotone improvement in the recovered latent distribution.
}

\begin{figure}
    \centering
    \includegraphics[width=0.55\textwidth]{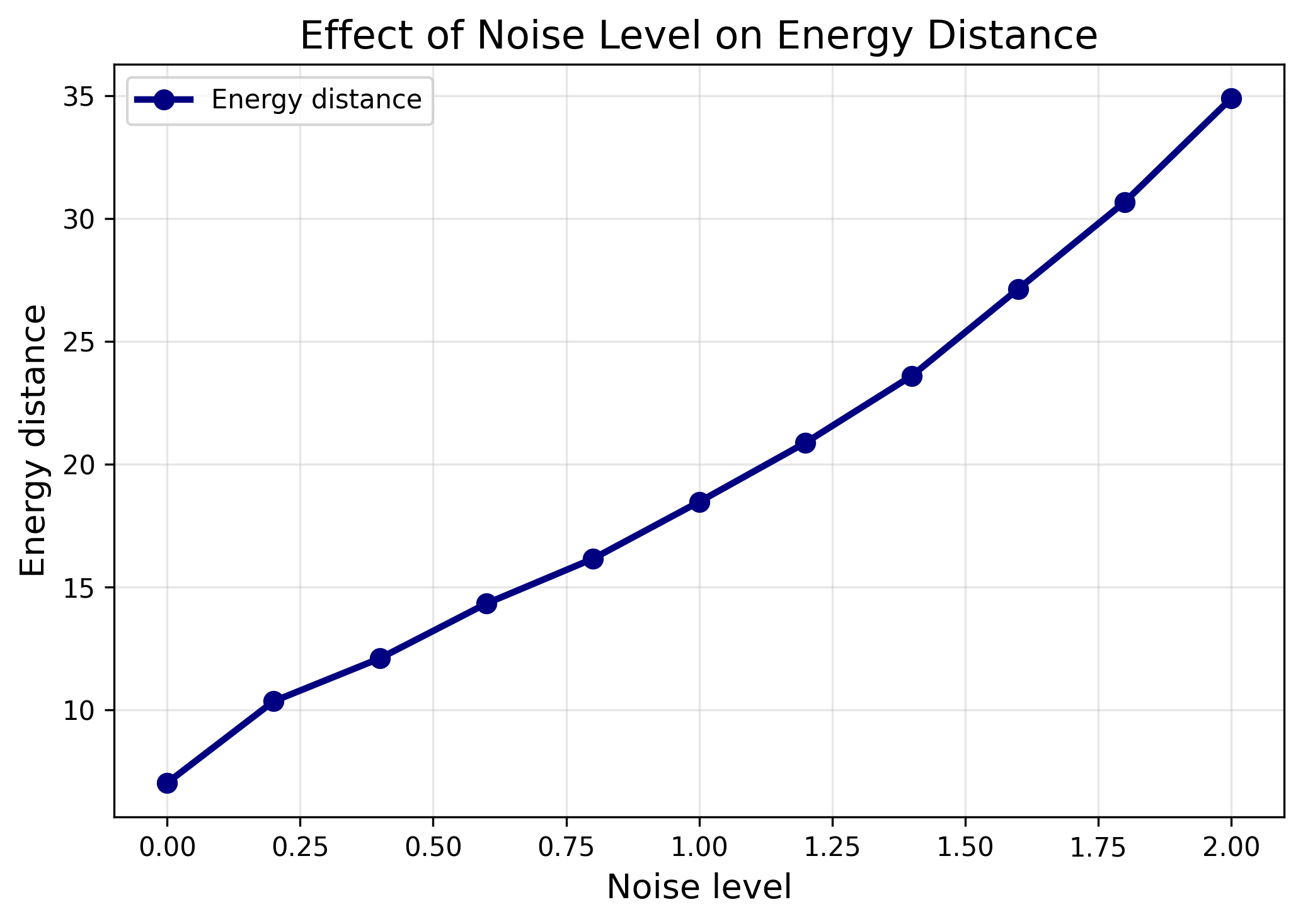}
    \caption{
    Effect of additive image noise on latent recovery in the one-mode protein
    experiment. The plot shows the energy distance between the recovered and
    true distributions in the normal-mode coordinate as a function of the noise
    level. 
    }
    \label{fig:latent_error_vs_noise}
\end{figure}

\revone{
Figure~\ref{fig:latent_error_vs_noise} shows the dependence on additive image noise. The latent-space error increases with noise, consistent with the local stability analysis: higher noise makes nearby conformations harder to distinguish and amplifies recovery error.
}
\section{Conclusion and Discussions}\label{sec:conclusion}

This work introduced a nonparametric framework for heterogeneous cryo-EM by formulating reconstruction as a stochastic inverse problem over probability measures. In this formulation, the unknown is a conformational distribution, and the observed image distribution is modeled as its random push-forward through the cryo-EM image formation process. This leads to a variational problem over probability measures, which we solve formally using a Wasserstein gradient flow and numerically using an interacting particle approximation.

The proposed formulation provides two complementary perspectives on heterogeneous reconstruction. First, it gives a flexible particle-based representation of structural heterogeneity that does not require prescribing a fixed number of conformational states or a parametric family for the latent distribution. Second, it clarifies the relationship between optimize-then-discretize and discretize-then-optimize approaches. In particular, MAP-type cryo-EM objectives can be interpreted as DTO approximations of a measure-valued variational problem, and we identify consistency conditions under which such discretized objectives converge to the corresponding continuous formulation.

The local stability and sample-complexity analysis identifies the main quantities that govern recovery: finite-sample error, Monte Carlo approximation, particle approximation, optimization error, model mismatch, image noise, and the observability of structural perturbations under the cryo-EM forward operator. This analysis also explains why accurate image-space matching does not automatically imply accurate latent recovery. Weakly observable structural directions, poor angular coverage, or high image noise can make distinct conformational distributions difficult to distinguish from their induced image distributions.

The numerical experiments support this picture in controlled settings. The one-dimensional and nanocluster examples demonstrate that the particle method can recover latent distributions from noisy observations generated by random forward maps, including high-dimensional image observations. The HSP90 normal-mode experiment provides a more physically motivated test case and shows that the method can recover meaningful low-dimensional conformational distributions from simulated cryo-EM-like images. The Gaussian-mixture comparisons illustrate the expected tradeoff between parametric efficiency and nonparametric flexibility: a correctly specified Gaussian-mixture model can outperform the particle method in a favorable low-dimensional setting, while the particle method is more robust when the true latent law does not belong to the prescribed parametric family. The sensitivity experiments further show that increasing the particle number improves image-space matching, whereas latent recovery remains governed by inverse stability, and that increasing image noise degrades conformational recovery.

\revtwo{These experiments should be interpreted as proof-of-concept studies rather than as a full high-resolution cryo-EM reconstruction pipeline. In the protein examples, the normal-mode parameterization provides a physically motivated low-dimensional model and allows evaluation against a known ground-truth distribution. However, it also restricts the recoverable conformational variability. Experimental datasets may contain anharmonic motions, local rearrangements, side-chain variability, or other effects outside the chosen latent subspace. Under such model discrepancy, the method can only recover the component of the conformational distribution expressible within the prescribed structural model.}

\revtwo{Several theoretical and computational questions remain open. The composed objective
$$
E(\rho_\theta)=D(F(\rho_\theta),\rho_y^\delta)
$$
is not expected to be globally geodesically convex because the cryo-EM forward map includes projection, rotations, CTF modulation, marginalization over nuisance variables, and noise. Thus, the Wasserstein gradient flow should be interpreted as a principled descent dynamics rather than as a globally convergent convex optimization method. Establishing sharper local regularity or Lojasiewicz-type inequalities for the composed objective near relevant minimizers remains an important direction for future work. A rigorous mean-field or Dobrushin-type stability theory for the finite-particle dynamics, together with a more detailed analysis of ADAM, minibatching, and Monte Carlo resampling, would further clarify the behavior of the implemented algorithm. Other geometries on probability measures, such as Wasserstein--Fisher--Rao or MMD-induced flows, may also be useful when reweighting, birth--death dynamics, or alternative regularization mechanisms are desired.}

Future work should benchmark the proposed formulation against production heterogeneity-analysis pipelines, including RELION classification, cryoSPARC heterogeneity tools, cryoDRGN, 3DFlex, and RECOVAR. Such comparisons will require incorporating experimental preprocessing, orientation estimation, CTF correction, and model-selection choices. More broadly, the framework developed here is not specific to cryo-EM and may be useful for other stochastic inverse problems in which the unknown object is a distribution over latent variables and the data are generated through a random forward operator.

\bibliography{references}
\bibliographystyle{abbrv}


\end{document}


\maketitle
\section{Supplementary Material}
\label{sec:supplement}

This document provides the derivations supporting Section~4 of the main text. The goal is not to prove a complete global convergence theory for the nonconvex particle method, but to justify the local stability quantities, the physical observability diagnostic, and the local error decomposition used in the paper.

The supplement is organized as follows. Section~S1.1 derives the Wasserstein tangent perturbation and the linearized image response. Section~S1.2 explains the origin of the linearized stability constant. Section~S1.3 gives the KL/Fisher and MMD tangent stability forms. Section~S1.4 relates the observed stability to the physical observability operator \(A_\pi\). Sections~S1.5 and~S1.6 derive the local error decomposition and the representative rates. Sections~S1.7 and~S1.8 provide numerical details for the nanocluster and protein normal-mode experiments.

\subsection{Wasserstein tangent perturbations and linearized image response}
\label{sec:sm-linearized-response}

Recall the noisy cryo-EM model from the main text,
\begin{equation}
    Y=T(\theta,\omega)=H_{\omega_r}\theta+\omega_n,
    \qquad
    \omega_n\sim \mathcal N(0,\sigma^2 I),
\end{equation}
where \(\omega_r\) denotes the acquisition variables, such as viewing direction, CTF parameters, shifts, defocus, and other imaging variables. In this supplement, we make explicit the decomposition
\begin{equation}
    H_{\omega_r}=C P R_{\omega_r},
\end{equation}
where \(R_{\omega_r}\) rotates the three-dimensional conformation, \(P\) is the projection operator, and \(C\) is the CTF or another linear image-domain filter. For notational simplicity, we write \(C\) instead of \(C_{\omega_r}\); the same formulas hold when the CTF also depends on \(\omega_r\). We write \(\Omega_\omega\) for the space of acquisition variables \(\omega_r\), with distribution \(\pi\). Throughout the supplement, \(F\) denotes the noisy forward map.

For a conformational distribution \(\rho\in\mathcal P(\Theta)\), the induced image density is
\begin{equation}
    q_\rho(y)
    =
    F(\rho)(y)
    =
    \int_\Theta
    \int_{\Omega_\omega}
    G_\sigma(y-H_{\omega_r}(\theta))
    \,d\pi(\omega_r)\,d\rho(\theta),
\end{equation}
where \(G_\sigma\) is the Gaussian density with covariance \(\sigma^2 I\). Let
\begin{equation}
    q^\star := F(\rho^\star)
\end{equation}
be the image density induced by the true conformational distribution.

We first recall the Wasserstein tangent perturbation used in the main text. Let
\begin{equation}
    \rho_t=(\operatorname{Id}+t v)_\#\rho^\star
\end{equation}
for an admissible velocity field \(v\in L^2(\rho^\star)\). Thus each latent point is displaced according to
\begin{equation}
    \theta \mapsto \theta+t v(\theta).
\end{equation}
For any smooth test function \(\zeta\), the definition of push-forward gives
\begin{equation}
    \int_\Theta \zeta(\theta)\,d\rho_t(\theta)
    =
    \int_\Theta
    \zeta(\theta+t v(\theta))
    \,d\rho^\star(\theta).
\end{equation}
Differentiating at \(t=0\), we obtain
\begin{equation}
    \left.\frac{d}{dt}\right|_{t=0}
    \int_\Theta \zeta(\theta)\,d\rho_t(\theta)
    =
    \int_\Theta
    \nabla \zeta(\theta)\cdot v(\theta)
    \,d\rho^\star(\theta).
\end{equation}
Equivalently,
\begin{equation}
    \left.\frac{d}{dt}\right|_{t=0}\rho_t
    =
    -\nabla_\theta\cdot(\rho^\star v)
\end{equation}
in the weak sense. This is the continuity-equation representation of a Wasserstein tangent vector.

The Wasserstein tangent norm of this perturbation is the minimal kinetic energy among all velocity fields generating the same signed perturbation. In particular, when \(v\) is the minimal-energy representative,
\begin{equation}
    \|v\|_{L^2(\rho^\star)}^2
    =
    \int_\Theta
    \|v(\theta)\|^2
    \,d\rho^\star(\theta)
\end{equation}
is the squared infinitesimal Wasserstein speed. Since the map
\begin{equation}
    T_t(\theta)=\theta+t v(\theta)
\end{equation}
is an admissible transport map from \(\rho^\star\) to \(\rho_t\), we have
\begin{equation}
    W_2(\rho_t,\rho^\star)
    \le
    t\|v\|_{L^2(\rho^\star)}.
\end{equation}
Under the standard regularity assumptions of Wasserstein tangent calculus, this upper bound is sharp to first order:
\begin{equation}
    W_2(\rho_t,\rho^\star)
    =
    t\|v\|_{L^2(\rho^\star)}
    +
    o(t).
\end{equation}
Thus \(\|v\|_{L^2(\rho^\star)}\) is the infinitesimal latent displacement measured in Wasserstein geometry. For more details, see Ambrosio, Gigli, and Savar\'e~\cite{ambrosio2008gradient} and Santambrogio~\cite{santambrogio2015optimal}.

We now compute the induced first-order perturbation of the image density. Define
\begin{equation}
    L_{\rho^\star}v(y)
    :=
    \left.\frac{d}{dt}\right|_{t=0}
    F(\rho_t)(y).
\end{equation}
Using the weak form of the perturbation and integrating by parts gives
\begin{equation}
\label{eq:sm-linearized-response-weak}
    L_{\rho^\star}v(y)
    =
    \int_{\Omega_\omega}
    \int_\Theta
    \rho^\star(\theta)
    v(\theta)\cdot
    \nabla_\theta G_\sigma(y-H_{\omega_r}(\theta))
    \,d\theta\,d\pi(\omega_r).
\end{equation}
If \(H_{\omega_r}\) is differentiable in \(\theta\), with
\begin{equation}
    J_{\omega_r}(\theta):=D_\theta H_{\omega_r}(\theta),
\end{equation}
then
\begin{equation}
    \nabla_\theta G_\sigma(y-H_{\omega_r}(\theta))
    =
    -J_{\omega_r}(\theta)^\top
    \nabla_yG_\sigma(y-H_{\omega_r}(\theta)).
\end{equation}
Since
\begin{equation}
    \nabla_yG_\sigma(y-H_{\omega_r}(\theta))
    =
    -\frac{1}{\sigma^2}
    (y-H_{\omega_r}(\theta))
    G_\sigma(y-H_{\omega_r}(\theta)),
\end{equation}
we obtain
\begin{equation}
\label{eq:sm-linearized-response-density}
    L_{\rho^\star}v(y)
    =
    \frac{1}{\sigma^2}
    \int_\Theta
    \int_{\Omega_\omega}
    \rho^\star(\theta)
    G_\sigma(y-H_{\omega_r}(\theta))
    \left\langle
        J_{\omega_r}(\theta)v(\theta),
        y-H_{\omega_r}(\theta)
    \right\rangle_Y
    \,d\pi(\omega_r)\,d\theta .
\end{equation}

We now rewrite the same object in a form adapted to cryo-EM observability. In the linearized cryo-EM model,
\begin{equation}
    H_{\omega_r}(\theta)=C P R_{\omega_r}\theta,
    \qquad
    Y=C P R_{\omega_r}\theta+\omega_n,
    \qquad
    \omega_n\sim\mathcal N(0,\sigma^2 I),
\end{equation}
with \(\theta\sim\rho^\star\), \(\omega_r\sim\pi\), and \(\omega_n\) independent of \((\theta,\omega_r)\). For a perturbation \(v\), define
\begin{equation}
\label{eq:sm-complete-data-response}
    A_v
    :=
    \frac{1}{\sigma^2}
    \left\langle
        C P R_{\omega_r}v(\theta),
        \omega_n
    \right\rangle_Y .
\end{equation}
The quantity \(A_v\) measures the complete-data linear response in the structural direction \(v\), before conditioning on the observed image \(Y=y\).

Since \(Y-H_{\omega_r}(\theta)=\omega_n\), Bayes' formula gives
\begin{equation}
\label{eq:sm-bayes-response}
    \mathbb E_{\theta,\omega_r}[A_v\mid Y=y]
    =
    \frac{N_v(y)}{q^\star(y)},
\end{equation}
where
\begin{equation}
\label{eq:sm-response-numerator}
    N_v(y)
    :=
    \frac{1}{\sigma^2}
    \int_\Theta
    \int_{\Omega_\omega}
    G_\sigma(y-H_{\omega_r}(\theta))
    \left\langle
        C P R_{\omega_r}v(\theta),
        y-H_{\omega_r}(\theta)
    \right\rangle_Y
    \,d\pi(\omega_r)\,d\rho^\star(\theta).
\end{equation}
Comparing this expression with~\eqref{eq:sm-linearized-response-density} yields the conditional representation
\begin{equation}
\label{eq:sm-conditional-response}
    L_{\rho^\star}v(y)
    =
    q^\star(y)
    \mathbb E_{\theta,\omega_{r}}[A_v\mid Y=y].
\end{equation}
Thus the observed image perturbation \(L_{\rho^\star}v\) is obtained by conditioning the complete-data response on the observed image. Consequently, marginalization over the latent conformation and acquisition variables can remove components of the structural perturbation from the observed image response.

\subsection{Origin of the linearized stability constant}
\label{sec:sm-linearized-stability-constant}

We now explain why the ratio between the image response and the latent Wasserstein velocity gives the natural local stability constant. Let \(D\) be an image-space discrepancy and suppose that, near \(q^\star=F_\sigma(\rho^\star)\), it admits the local quadratic expansion
\begin{equation}
\label{eq:sm-D-quadratic-expansion}
    D(q^\star+\epsilon h,q^\star)^2
    =
    \epsilon^2
    \|h\|_{D,q^\star}^2
    +
    o(\epsilon^2),
    \qquad
    \epsilon\to0,
\end{equation}
for signed perturbations \(h\) with zero total mass. The norm \(\|\cdot\|_{D,q^\star}\) is the tangent norm induced by the discrepancy \(D\) at \(q^\star\). For example, for KL this is the Fisher-type tangent norm, while for MMD it is the RKHS tangent norm.

Along the Wasserstein perturbation
\begin{equation}
    \rho_t=(\operatorname{Id}+tv)_\#\rho^\star,
\end{equation}
the forward map has the first-order expansion
\begin{equation}
    F(\rho_t)
    =
    q^\star
    +
    tL_{\rho^\star}v
    +
    o(t).
\end{equation}
Therefore, applying the quadratic expansion of \(D\) with \(h=L_{\rho^\star}v\), we obtain
\begin{equation}
    D(F(\rho_t),F(\rho^\star))^2
    =
    t^2
    \|L_{\rho^\star}v\|_{D,q^\star}^2
    +
    o(t^2).
\end{equation}
Equivalently,
\begin{equation}
    D(F(\rho_t),F(\rho^\star))
    =
    t
    \|L_{\rho^\star}v\|_{D,q^\star}
    +
    o(t).
\end{equation}
On the latent side, when \(v\) is the minimal-energy representative of the tangent direction, the Wasserstein expansion gives
\begin{equation}
    W_2(\rho_t,\rho^\star)
    =
    t
    \|v\|_{L^2(\rho^\star)}
    +
    o(t).
\end{equation}
Combining the image-space and latent-space expansions yields
\begin{equation}
    \lim_{t\to0}
    \frac{
        D(F_\sigma(\rho_t),F_\sigma(\rho^\star))
    }{
        W_2(\rho_t,\rho^\star)
    }
    =
    \frac{
        \|L_{\rho^\star}v\|_{D,q^\star}
    }{
        \|v\|_{L^2(\rho^\star)}
    }.
\end{equation}
Thus this ratio is the infinitesimal observable image response per unit latent Wasserstein displacement in the direction \(v\). It is therefore the local quantity that controls inverse stability.

This motivates the linearized local stability constant
\begin{equation}
\label{eq:sm-gamma-lin}
    \gamma_{\mathrm{lin}}^{(D)}
    :=
    \inf_{v\neq0}
    \frac{
        \|L_{\rho^\star}v\|_{D,q^\star}^2
    }{
        \|v\|_{L^2(\rho^\star)}^2
    }.
\end{equation}
The condition \(\gamma_{\mathrm{lin}}^{(D)}>0\) means that no nonzero Wasserstein tangent direction is invisible to first order. Equivalently, every unit latent perturbation produces an image-space response of size at least \(\sqrt{\gamma_{\mathrm{lin}}^{(D)}}\). If \(\gamma_{\mathrm{lin}}^{(D)}=0\), then there exist latent perturbations that are arbitrarily difficult to detect from images. In cryo-EM, this corresponds to local non-identifiability caused by projection, rotations, image noise, symmetries, CTF zeros, limited resolution, or missing angular information.

\subsection{KL and MMD tangent stability constants}
\label{sec:sm-kl-mmd-stability}

Let \(D\) be an image discrepancy and assume that it has a local quadratic expansion at \(q^\star=F_\sigma(\rho^\star)\):
\begin{equation}
    D(q^\star+\epsilon h,q^\star)^2
    =
    \epsilon^2
    \|h\|_{D,q^\star}^2
    +
    o(\epsilon^2),
    \qquad
    \epsilon\to0.
\end{equation}
The corresponding linearized stability constant is
\begin{equation}
    \gamma_{\mathrm{lin}}^{(D)}
    =
    \inf_{v\neq0}
    \frac{
        \|L_{\rho^\star}v\|_{D,q^\star}^2
    }{
        \|v\|_{L^2(\rho^\star)}^2
    }.
\end{equation}
The discrepancy \(D\) enters only through the tangent norm \(\|\cdot\|_{D,q^\star}\); the linearized marginal response \(L_{\rho^\star}v\) is common to all choices of discrepancy.

\paragraph{KL/Fisher}
For the KL/Fisher tangent geometry,
\begin{equation}
    \|h\|_{\mathrm{KL},q^\star}^2
    =
    \int
    \frac{h(y)^2}{q^\star(y)}
    \,dy.
\end{equation}
Using
\begin{equation}
    L_{\rho^\star}v(y)
    =
    q^\star(y)m_v(y),
    \qquad
    m_v(y):=\mathbb E[A_v\mid Y=y],
\end{equation}
we obtain
\begin{equation}
    \|L_{\rho^\star}v\|_{\mathrm{KL},q^\star}^2
    =
    \int q^\star(y)m_v(y)^2\,dy
    =
    \mathbb E_{Y\sim q^\star}
    \left[
        m_v(Y)^2
    \right].
\end{equation}
Hence
\begin{equation}
    \gamma_{\mathrm{lin}}^{(\mathrm{KL})}
    =
    \inf_{v\neq0}
    \frac{
        \mathbb E_{Y\sim q^\star}
        \left[
            \left(\mathbb E[A_v\mid Y]\right)^2
        \right]
    }{
        \|v\|_{L^2(\rho^\star)}^2
    }.
\end{equation}

On a finite-dimensional perturbation subspace \(\mathcal V_d=\operatorname{span}\{v_1,\ldots,v_d\}\), define
\begin{equation}
    m_i(y):=\mathbb E[A_{v_i}\mid Y=y].
\end{equation}
Then the KL/Fisher stability matrix is
\begin{equation}
    G_{\mathrm{KL}}[i,j]
    =
    \mathbb E_{Y\sim q^\star}
    \left[
        m_i(Y)m_j(Y)
    \right],
\end{equation}
with latent mass matrix
\begin{equation}
    M[i,j]
    =
    \langle v_i,v_j\rangle_{L^2(\rho^\star)}.
\end{equation}
The finite-dimensional stability constant is the smallest positive generalized eigenvalue of
\begin{equation}
    G_{\mathrm{KL}}c=\lambda M c.
\end{equation}

\paragraph{MMD}
For MMD with kernel \(k\), the squared MMD is exactly quadratic in signed perturbations:
\begin{equation}
    \mathrm{MMD}_k(q^\star+\epsilon h,q^\star)^2
    =
    \epsilon^2
    \int
    \int
    k(y,y')
    h(y)h(y')
    \,dy\,dy'.
\end{equation}
Therefore the MMD tangent norm is
\begin{equation}
    \|h\|_{\mathrm{MMD},q^\star}^2
    =
    \int
    \int
    k(y,y')
    h(y)h(y')
    \,dy\,dy'.
\end{equation}
Using again \(L_{\rho^\star}v(y)=q^\star(y)m_v(y)\), we get
\begin{equation}
    \|L_{\rho^\star}v\|_{\mathrm{MMD},q^\star}^2
    =
    \mathbb E_{Y,Y'\sim q^\star}
    \left[
        k(Y,Y')m_v(Y)m_v(Y')
    \right],
\end{equation}
where \(Y\) and \(Y'\) are independent samples from \(q^\star\). Thus
\begin{equation}
    \gamma_{\mathrm{lin}}^{(\mathrm{MMD})}
    =
    \inf_{v\neq0}
    \frac{
        \mathbb E_{Y,Y'\sim q^\star}
        \left[
            k(Y,Y')m_v(Y)m_v(Y')
        \right]
    }{
        \|v\|_{L^2(\rho^\star)}^2
    }.
\end{equation}
On a finite-dimensional perturbation subspace,
\begin{equation}
    G_{\mathrm{MMD}}[i,j]
    =
    \mathbb E_{Y,Y'\sim q^\star}
    \left[
        k(Y,Y')m_i(Y)m_j(Y')
    \right],
\end{equation}
with the same latent mass matrix \(M\). The corresponding stability estimate is again obtained from the smallest positive generalized eigenvalue
\begin{equation}
    G_{\mathrm{MMD}}c=\lambda M c.
\end{equation}
For the Gaussian kernel
\begin{equation}
    k_\tau(y,z)
    =
    \exp\left(
        -\frac{\|y-z\|^2}{2\tau^2}
    \right),
\end{equation}
the matrix becomes
\begin{equation}
    G_{\mathrm{MMD},\tau}[i,j]
    =
    \mathbb E_{Y,Y'\sim q^\star}
    \left[
        \exp\left(
            -\frac{\|Y-Y'\|^2}{2\tau^2}
        \right)
        m_i(Y)m_j(Y')
    \right].
\end{equation}
This shows explicitly how the kernel bandwidth weights correlations between observed directional responses.

\subsection{Physical observability and comparison with observed stability}
\label{sec:sm-physical-observability}

We now relate the observed stability to the physical complete-data observability operator. Recall that
\begin{equation}
    A_v
    =
    \frac{1}{\sigma^2}
    \left\langle
        C P R_{\omega_r}v(\theta),
        \omega_n
    \right\rangle_Y,
    \qquad
    \omega_n\sim\mathcal N(0,\sigma^2 I).
\end{equation}
Conditioning on \(\theta\) and \(\omega_r\), the vector
\begin{equation}
    u=C P R_{\omega_r}v(\theta)
\end{equation}
is fixed. Hence
\begin{equation}
    \mathbb E_{\omega_n}
    \left[
        \left\langle u,\omega_n\right\rangle_Y^2
    \right]
    =
    \sigma^2
    \|u\|_Y^2.
\end{equation}
Therefore
\begin{equation}
\label{eq:sm-complete-data-second-moment}
    \mathbb E_{\theta,\omega_r,\omega_n}
    \left[
        A_v^2
    \right]
    =
    \frac{1}{\sigma^2}
    \int_\Theta
    \int_{\Omega_\omega}
    \|C P R_{\omega_r}v(\theta)\|_Y^2
    \,d\pi(\omega_r)\,d\rho^\star(\theta).
\end{equation}
Define the physical observability operator averaged over the acquisition distribution \(\pi\) by
\begin{equation}
\label{eq:sm-Api}
    A_\pi
    :=
    \int_{\Omega_\omega}
    R_{\omega_r}^\ast
    P^\ast C^\ast C P
    R_{\omega_r}
    \,d\pi(\omega_r).
\end{equation}
Then, for any structural perturbation \(u\in\Theta\),
\begin{equation}
    \int_{\Omega_\omega}
    \|C P R_{\omega_r}u\|_Y^2
    \,d\pi(\omega_r)
    =
    \langle u,A_\pi u\rangle_\Theta.
\end{equation}
Consequently,
\begin{equation}
\label{eq:sm-Api-second-moment}
    \mathbb E_{\theta,\omega_r,\omega_n}
    \left[
        A_v^2
    \right]
    =
    \frac{1}{\sigma^2}
    \int_\Theta
    \left\langle
        v(\theta),
        A_\pi v(\theta)
    \right\rangle_\Theta
    \,d\rho^\star(\theta).
\end{equation}
This identity says that \(A_\pi\) measures the complete-data physical response energy.

The observed response is the conditional response
\begin{equation}
    m_v(y)
    =
    \mathbb E[A_v\mid Y=y].
\end{equation}
For the KL/Fisher tangent norm,
\begin{equation}
    \|L_{\rho^\star}v\|_{\mathrm{KL},q^\star}^2
    =
    \mathbb E_{Y\sim q^\star}
    \left[
        m_v(Y)^2
    \right].
\end{equation}
The complete-data and observed response energies are related by the conditional variance decomposition
\begin{equation}
\label{eq:sm-conditional-variance}
    \mathbb E_{\theta,\omega_r,\omega_n}
    \left[
        A_v^2
    \right]
    =
    \mathbb E_{Y\sim q^\star}
    \left[
        m_v(Y)^2
    \right]
    +
    \mathbb E_{Y\sim q^\star}
    \left[
        \operatorname{Var}(A_v\mid Y)
    \right].
\end{equation}
Thus marginalization over unobserved variables can only reduce the observed Fisher response energy.

For a general discrepancy \(D\), assume that the discrepancy tangent norm is \(L^2(q^\star)\)-bounded on the relevant response functions: there exists \(C_D>0\) such that
\begin{equation}
\label{eq:sm-D-domination}
    \|q^\star m\|_{D,q^\star}^2
    \le
    C_D
    \|m\|_{L^2(q^\star)}^2 .
\end{equation}
Since
\begin{equation}
    L_{\rho^\star}v=q^\star m_v,
\end{equation}
we get
\begin{equation}
    \|L_{\rho^\star}v\|_{D,q^\star}^2
    \le
    C_D
    \mathbb E_{Y\sim q^\star}
    \left[
        m_v(Y)^2
    \right]
    \le
    C_D
    \mathbb E_{\theta,\omega_r,\omega_n}
    \left[
        A_v^2
    \right].
\end{equation}
Combining with~\eqref{eq:sm-Api-second-moment} gives
\begin{equation}
\label{eq:sm-observed-physical-bound}
    \|L_{\rho^\star}v\|_{D,q^\star}^2
    \le
    \frac{C_D}{\sigma^2}
    \int_\Theta
    \left\langle
        v(\theta),
        A_\pi v(\theta)
    \right\rangle_\Theta
    \,d\rho^\star(\theta).
\end{equation}
Therefore, on a common finite-dimensional perturbation subspace,
\begin{equation}
\label{eq:sm-gamma-Api-bound}
    \gamma_{\mathrm{lin}}^{(D)}
    \le
    \frac{C_D}{\sigma^2}
    \lambda_{\min}(A_\pi).
\end{equation}
This is not an equivalence. It only says that poor physical visibility forces poor observed stability. In particular, a null direction of \(A_\pi\) forces a null direction for the observed stability matrix. The converse need not hold, because information can still be lost by conditioning, by symmetries, or by the choice of discrepancy.

\subsection{Derivation of the local oracle bound}
\label{sec:sm-local-oracle-bound}

We derive the local error decomposition used in Section~4.3 of the main text. Let \(\widehat{\mu}_n\) be the empirical distribution of the observed images, and let \(F_{\sigma,M}\) denote the Monte Carlo approximation of \(F_\sigma\). Let
\begin{equation}
    \mathcal P_K(\Theta)
    =
    \left\{
        \frac{1}{K}
        \sum_{i=1}^K
        \delta_{\theta_i}
        :
        \theta_i\in\Theta
    \right\}.
\end{equation}
Let
\begin{equation}
    \rho_K^\star
    \in
    \mathop{\mathrm{arg\,min}}_{\rho\in\mathcal P_K(\Theta)}
    W_2(\rho,\rho^\star)
\end{equation}
be a best \(K\)-particle approximation, and define
\begin{equation}
    b_K^{W_2}
    :=
    W_2(\rho_K^\star,\rho^\star),
    \qquad
    b_K^{\mathrm{img}}
    :=
    D(F_\sigma(\rho_K^\star),F_\sigma(\rho^\star)).
\end{equation}
Assume that local inverse stability holds on the relevant \(K\)-particle class:
\begin{equation}
\label{eq:sm-K-particle-stability}
    W_2(\rho,\rho_K^\star)
    \le
    \frac{1}{\sqrt{\gamma_K^{(D)}}}
    D(F_\sigma(\rho),F_\sigma(\rho_K^\star)),
    \qquad
    \rho\in\mathcal P_K(\Theta).
\end{equation}
Then
\begin{equation}
\begin{aligned}
    W_2(\widehat{\rho}_{n,K,M},\rho^\star)
    &\le
    W_2(\widehat{\rho}_{n,K,M},\rho_K^\star)
    +
    b_K^{W_2} \\
    &\le
    \frac{1}{\sqrt{\gamma_K^{(D)}}}
    D(F_\sigma(\widehat{\rho}_{n,K,M}),F_\sigma(\rho_K^\star))
    +
    b_K^{W_2}.
\end{aligned}
\end{equation}
It remains to bound the image-space discrepancy. Define
\begin{equation}
    \epsilon_{\mathrm{stat}}(n)
    :=
    D(\widehat{\mu}_n,F_\sigma(\rho^\star)),
\end{equation}
and
\begin{equation}
    \epsilon_{\mathrm{MC}}(M,K)
    :=
    \sup_{\rho\in\mathcal P_K(\Theta)}
    D(F_{\sigma,M}(\rho),F_\sigma(\rho)).
\end{equation}
Assume that \(\widehat{\rho}_{n,K,M}\) is an \(\epsilon_{\mathrm{opt}}\)-approximate minimizer in the distance scale of \(D\):
\begin{equation}
    D(F_{\sigma,M}(\widehat{\rho}_{n,K,M}),\widehat{\mu}_n)
    \le
    \inf_{\rho\in\mathcal P_K(\Theta)}
    D(F_{\sigma,M}(\rho),\widehat{\mu}_n)
    +
    \epsilon_{\mathrm{opt}}.
\end{equation}
Using \(\rho_K^\star\) as a competitor gives
\begin{equation}
    D(F_{\sigma,M}(\widehat{\rho}_{n,K,M}),\widehat{\mu}_n)
    \le
    D(F_{\sigma,M}(\rho_K^\star),\widehat{\mu}_n)
    +
    \epsilon_{\mathrm{opt}}.
\end{equation}
By the triangle inequality,
\begin{equation}
\begin{aligned}
    D(F_\sigma(\widehat{\rho}_{n,K,M}),F_\sigma(\rho_K^\star))
    &\le
    D(F_\sigma(\widehat{\rho}_{n,K,M}),F_{\sigma,M}(\widehat{\rho}_{n,K,M})) \\
    &\quad
    +
    D(F_{\sigma,M}(\widehat{\rho}_{n,K,M}),\widehat{\mu}_n) \\
    &\quad
    +
    D(\widehat{\mu}_n,F_{\sigma,M}(\rho_K^\star)) \\
    &\quad
    +
    D(F_{\sigma,M}(\rho_K^\star),F_\sigma(\rho_K^\star)).
\end{aligned}
\end{equation}
The first and last terms are controlled by \(\epsilon_{\mathrm{MC}}(M,K)\). The middle term is controlled by approximate optimality. Finally,
\begin{equation}
\begin{aligned}
    D(F_{\sigma,M}(\rho_K^\star),\widehat{\mu}_n)
    &\le
    D(F_{\sigma,M}(\rho_K^\star),F_\sigma(\rho_K^\star)) \\
    &\quad
    +
    D(F_\sigma(\rho_K^\star),F_\sigma(\rho^\star)) \\
    &\quad
    +
    D(F_\sigma(\rho^\star),\widehat{\mu}_n) \\
    &\le
    \epsilon_{\mathrm{MC}}(M,K)
    +
    b_K^{\mathrm{img}}
    +
    \epsilon_{\mathrm{stat}}(n).
\end{aligned}
\end{equation}
Combining the preceding inequalities, up to universal constants,
\begin{equation}
    D(F_\sigma(\widehat{\rho}_{n,K,M}),F_\sigma(\rho_K^\star))
    \lesssim
    \epsilon_{\mathrm{stat}}(n)
    +
    \epsilon_{\mathrm{MC}}(M,K)
    +
    b_K^{\mathrm{img}}
    +
    \epsilon_{\mathrm{opt}}.
\end{equation}
If the forward model or structural parametrization is misspecified, we add an image-space model error \(\epsilon_{\mathrm{model}}\). Therefore
\begin{equation}
    D(F_\sigma(\widehat{\rho}_{n,K,M}),F_\sigma(\rho_K^\star))
    \lesssim
    \epsilon_{\mathrm{stat}}(n)
    +
    \epsilon_{\mathrm{MC}}(M,K)
    +
    b_K^{\mathrm{img}}
    +
    \epsilon_{\mathrm{opt}}
    +
    \epsilon_{\mathrm{model}}.
\end{equation}
Substituting into the local inverse-stability estimate gives
\begin{equation}
\label{eq:sm-local-oracle-final}
    W_2(\widehat{\rho}_{n,K,M},\rho^\star)
    \lesssim
    \frac{1}{\sqrt{\gamma_K^{(D)}}}
    \left[
        \epsilon_{\mathrm{stat}}(n)
        +
        \epsilon_{\mathrm{MC}}(M,K)
        +
        b_K^{\mathrm{img}}
        +
        \epsilon_{\mathrm{opt}}
        +
        \epsilon_{\mathrm{model}}
    \right]
    +
    b_K^{W_2}.
\end{equation}
This is the local error decomposition used in the main text.

\subsection{Rates for the oracle terms}
\label{sec:sm-oracle-rates}

We now summarize the representative rates used in Section~4.3 of the main text. The purpose is to identify the scaling of each term in the local error decomposition under standard boundedness, Lipschitz, and effective dimension assumptions.

\paragraph{Statistical error.}
For bounded-kernel MMD, let
\begin{equation}
    \widehat{\mu}_n
    =
    \frac{1}{n}
    \sum_{i=1}^n
    \delta_{Y_i},
    \qquad
    Y_i \stackrel{\mathrm{i.i.d.}}{\sim} \mu^\star.
\end{equation}
Assume that \(k\) is a positive definite kernel with RKHS \(\mathcal H_k\) and
\begin{equation}
    \sup_y k(y,y)\le \kappa^2.
\end{equation}
With \(\varphi(y)=k(y,\cdot)\), the MMD can be written as
\begin{equation}
    \operatorname{MMD}_k(\widehat{\mu}_n,\mu^\star)
    =
    \left\|
        \frac{1}{n}
        \sum_{i=1}^n
        \left(
            \varphi(Y_i)
            -
            \mathbb E_{Y\sim\mu^\star}\varphi(Y)
        \right)
    \right\|_{\mathcal H_k}.
\end{equation}
The summands are independent, mean-zero, and bounded in \(\mathcal H_k\). Hence the Hilbert-space variance calculation gives
\begin{equation}
    \mathbb E
    \left[
        \operatorname{MMD}_k(\widehat{\mu}_n,\mu^\star)
    \right]
    \le
    \frac{\kappa}{\sqrt n}.
\end{equation}
A bounded-differences argument gives the corresponding high-probability bound. Thus
\begin{equation}
    \epsilon_{\mathrm{stat}}(n)
    =
    O_P(n^{-1/2})
\end{equation}
for bounded-kernel MMD; see Gretton et al.~\cite{Gretton2012} and McDiarmid~\cite{mcdiarmid1989method}.

\paragraph{Monte Carlo error.}
For fixed \(\rho\), the Monte Carlo forward map is an average of \(M\) independent random push-forwards. In bounded-kernel MMD, the same Hilbert-space variance calculation gives
\begin{equation}
    D_{\mathrm{MMD}}(F_{\sigma,M}(\rho),F_\sigma(\rho))
    =
    O_P(M^{-1/2})
\end{equation}
for fixed \(\rho\). To make the bound uniform over \(\mathcal P_K(\Theta)\), one covers the relevant \(K\)-particle class by an \(\eta\)-net. If this class has effective dimension \(dK\), then
\begin{equation}
    N(\eta,\mathcal P_K(\Theta))
    \lesssim
    \left(
        \frac{C}{\eta}
    \right)^{dK}.
\end{equation}
Applying the pointwise concentration bound on the net and extending by Lipschitz continuity gives
\begin{equation}
    \epsilon_{\mathrm{MC}}(M,K)
    =
    O_P
    \left(
        \sqrt{
            \frac{dK\log M}{M}
        }
    \right),
\end{equation}
up to constants and lower-order discretization terms. This is a standard concentration-plus-covering argument from empirical process theory; see van der Vaart and Wellner~\cite{vandervaartwellner1996weak}.

\paragraph{Particle approximation error.}
The best latent \(K\)-particle approximation error is
\begin{equation}
    b_K^{W_2}
    =
    \inf_{\rho\in\mathcal P_K(\Theta)}
    W_2(\rho,\rho^\star).
\end{equation}
If \(\rho^\star\) is effectively supported on a \(d\)-dimensional latent set and satisfies the standard regularity assumptions of quantization theory, then
\begin{equation}
    b_K^{W_2}
    =
    O(K^{-1/d});
\end{equation}
see Graf and Luschgy~\cite{Graf2000}. This is a best-approximation rate, not the sampling rate of an empirical measure. Related empirical Wasserstein rates are discussed in Fournier and Guillin~\cite{Fournier2015}.

The corresponding image-space particle bias is
\begin{equation}
    b_K^{\mathrm{img}}
    =
    D(F_\sigma(\rho_K^\star),F_\sigma(\rho^\star)).
\end{equation}
If \(F_\sigma\) is locally Lipschitz from latent \(W_2\) to the image discrepancy \(D\), namely
\begin{equation}
    D(F_\sigma(\rho),F_\sigma(\rho'))
    \le
    L_F W_2(\rho,\rho'),
\end{equation}
then
\begin{equation}
    b_K^{\mathrm{img}}
    \le
    L_F b_K^{W_2}
    =
    O(K^{-1/d}).
\end{equation}
In the cryo-EM model, the largest eigenvalue of the physical observability operator \(A_\pi\) on the relevant subspace controls the largest forward amplification of structural perturbations, and hence contributes to the local forward Lipschitz constant.

\subsection{Numerical details for the nanocluster experiment}
\label{sec:sm-nanocluster-forward-map}

The deterministic forward map used in Section~5.2 is defined as
\begin{equation}
    T:\mathbb R^2\to\mathbb R^{128\times128},
\end{equation}
which transforms a latent variable \(\theta\in\mathbb R^2\) into a two-dimensional image. The transformation is constructed through a symmetric arrangement of atomic positions.

First, initialize the first atomic position by halving the latent variable:
\begin{equation}
    a_1=\frac{\theta}{2}.
\end{equation}
Three additional atomic positions are generated by applying axis reflections:
\begin{equation}
    a_2=a_1\odot[1,-1],
    \qquad
    a_3=a_1\odot[-1,1],
    \qquad
    a_4=a_1\odot[-1,-1],
\end{equation}
where \(\odot\) denotes element-wise multiplication. These atomic positions are stacked into
\begin{equation}
    A=
    \left[
        a_1^\top,
        a_2^\top,
        a_3^\top,
        a_4^\top
    \right]^\top
    \in\mathbb R^{4\times2}.
\end{equation}

The rendering process maps the atomic structure into a two-dimensional image of size \(128\times128\). The pixel grid is a uniformly spaced grid \(g\in\mathbb R^{2\times128\times128}\) defined over the range \([-4,4]\), with grid points
\begin{equation}
    g_{ij}
    =
    \left(
        -4+\frac{8i}{128},
        -4+\frac{8j}{128}
    \right),
    \qquad
    i,j=0,\ldots,127.
\end{equation}
For each atomic position \(a_k\), we compute the squared displacement relative to each grid point:
\begin{equation}
    \Delta g_{ijk}
    =
    a_k-g_{ij},
\end{equation}
and the squared distance
\begin{equation}
    d_{ijk}^2
    =
    \sum_{\ell=1}^2
    (\Delta g_{ijk})_\ell^2.
\end{equation}
The atomic contributions are modeled with a Gaussian kernel:
\begin{equation}
    K_{ijk}
    =
    \exp\left(
        -\frac{d_{ijk}^2}{2\tau^2}
    \right).
\end{equation}
The final pixel intensity is obtained by summing the contributions of all atoms:
\begin{equation}
    I_{ij}
    =
    \sum_{k=1}^4
    K_{ijk}.
\end{equation}
This deterministic renderer is the map used to generate the nanocluster image observations in Section~5.2; additive image noise is then applied separately when specified in the experiment.

\subsection{Normal-mode visualization for the protein experiment}
\label{sec:sm-normal-modes}

The protein experiments use a finite-dimensional normal-mode parametrization of conformational variability. Starting from a reference structure, we compute low-frequency normal modes and represent a conformation by perturbing the reference coordinates along the dominant modes. The latent coordinates used in the particle method are the amplitudes of these normal-mode perturbations.

\begin{figure}
  \centering

  \begin{minipage}[b]{0.82\textwidth}
    \includegraphics[width=\textwidth]{Figures/mode1_move_compact.pdf}
  \end{minipage}

  \vspace{2mm}

  \begin{minipage}[b]{0.82\textwidth}
    \includegraphics[width=\textwidth]{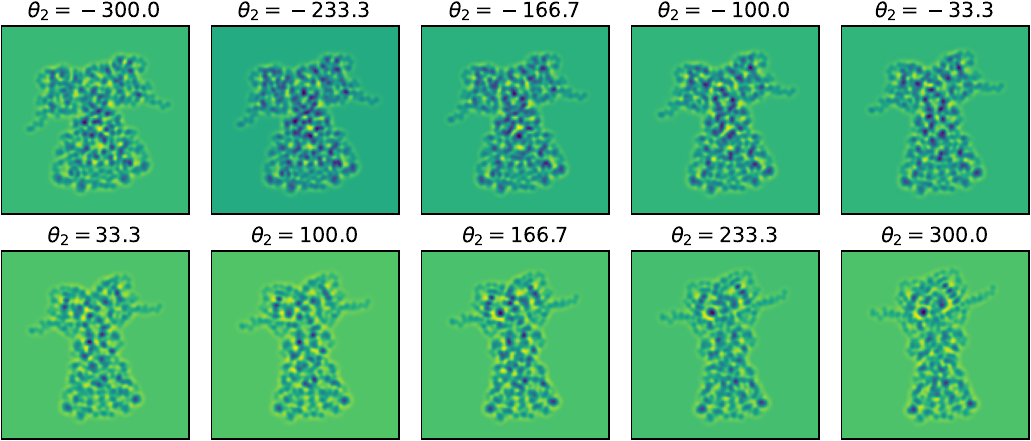}
  \end{minipage}

  \vspace{2mm}

  \begin{minipage}[b]{0.82\textwidth}
    \includegraphics[width=\textwidth]{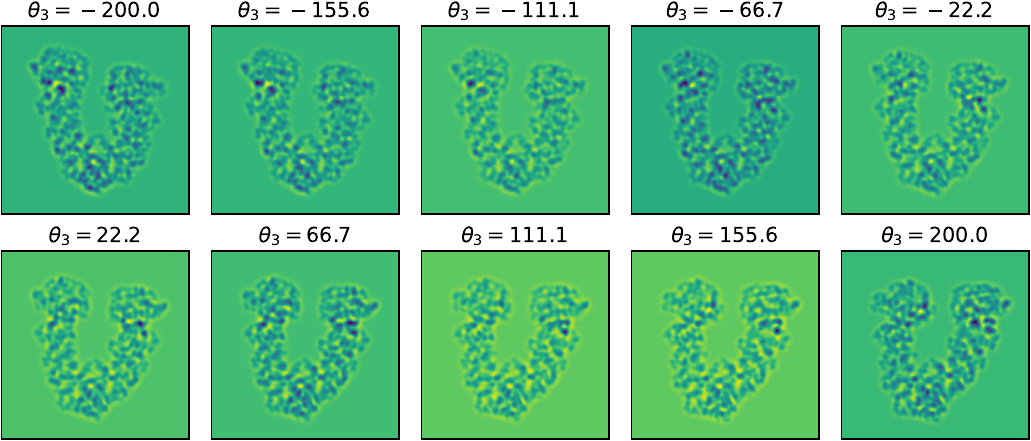}
  \end{minipage}

  \vspace{2mm}

  \begin{minipage}[b]{0.82\textwidth}
    \includegraphics[width=\textwidth]{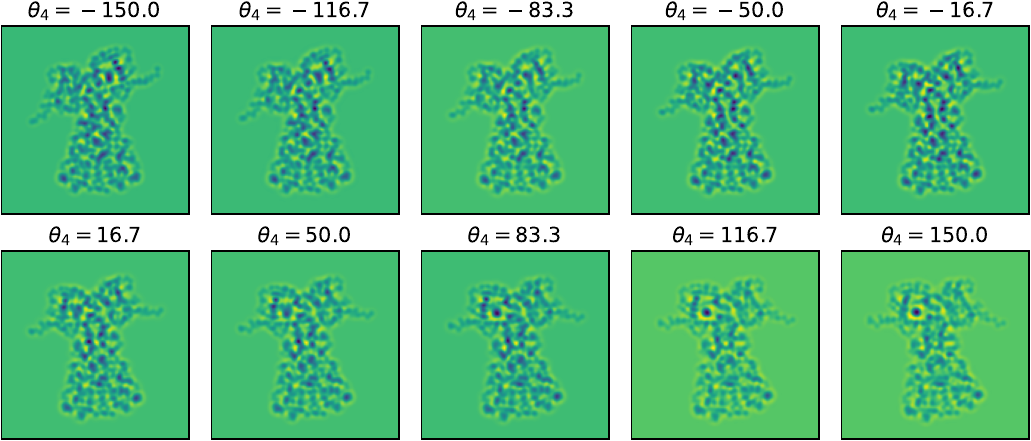}
  \end{minipage}

  \caption{
    Visualization of conformational variability along $4$ dominant normal modes. Each pair of rows corresponds to a distinct mode, ordered from top to bottom starting with Mode $1$ to Mode $4$.
  }
  \label{fig:mode_move}
\end{figure}

Figure~\ref{fig:mode_move} visualizes the conformational variability associated with the first four dominant modes used in the experiments. This visualization is intended only to illustrate the geometric meaning of the latent coordinates. The particle method is applied to the distribution of normal-mode amplitudes, while the cryo-EM forward model maps the corresponding perturbed structures to simulated projection images.

\FloatBarrier
\bibliography{references}
\bibliographystyle{abbrv}